\documentclass{article}
\usepackage[round]{natbib}
\usepackage{arxiv}
\usepackage{alias}
\usepackage[utf8]{inputenc}
\usepackage[T1]{fontenc}    
\usepackage[framemethod=TikZ]{mdframed}
\usepackage{caption}
\usepackage{subcaption}
\usepackage{xcolor}
\usepackage[colorlinks,citecolor=blue,urlcolor=blue,bookmarks=false,hypertexnames=true,linkcolor=blue]{hyperref}
\usepackage{nicefrac}       
\usepackage{microtype}      
\usepackage{amsfonts}
\usepackage{amsmath}
\usepackage{amssymb}
\usepackage{amsthm}
\usepackage{xcolor}
\usepackage{thm-restate}
\usepackage{booktabs}
\usepackage{tabu}
\usepackage{mathrsfs}
\usepackage{graphicx}
\usepackage{graphics}
\usepackage{wrapfig}
\usepackage{enumitem}
\usepackage{pgfplots}
\usepackage{newtxtext}
\usepackage{tikz}
\usetikzlibrary{arrows.meta,
                chains,
                positioning}
\usepackage{cleveref}

\declaretheorem[name=Conjecture,style=examplestyle]{conj}

\title{Fundamental tradeoffs between memorization and robustness in random features and neural tangent regimes}
\author{%
\name Elvis Dohmatob
\email{gmdopp@gmail.com}\\
\addr{Criteo AI Lab\\(*Now at Facebook)}\\
}
 
\begin{document}
\maketitle
\begin{abstract}
    This work studies the (non)robustness of two-layer neural networks in various high-dimensional linearized regimes. We establish fundamental trade-offs between memorization and robustness, as measured by the Sobolev-seminorm of the model w.r.t the data distribution, i.e the square root of the average squared $L_2$-norm of the gradients of the model w.r.t the its input. More precisely, if $n$ is the number of training examples, $d$ is the input dimension, and $k$ is the number of hidden neurons in a two-layer neural network, we prove for a large class of activation functions that, if the model memorizes even a fraction of the training, then its Sobolev-seminorm is lower-bounded by (i) $\sqrt{n}$ in case of infinite-width random features (RF) or neural tangent kernel (NTK) with $d \gtrsim n$; (ii) $\sqrt{n}$ in case of finite-width RF with proportionate scaling of $d$ and $k$; and (iii) $\sqrt{n/k}$ in case of finite-width NTK with proportionate scaling of $d$ and $k$. Moreover, all of these lower-bounds are tight: they are attained by the min-norm / least-squares interpolator (when $n$, $d$, and $k$ are in the appropriate interpolating regime). All our results hold as soon as data is log-concave isotropic, and there is label-noise, i.e the target variable is not a deterministic function of the data / features.
    We empirically validate our theoretical results with experiments. Accidentally, these experiments also reveal for the first time, (iv) a multiple-descent phenomenon in the robustness of the min-norm interpolator.
\end{abstract}

\section{Introduction}
\label{sec:intro}
Consider a random dataset $\mathcal D_n = \{(x_1,y_1),\ldots,(x_n,y_n)\}$ consisting of $n$ labeled iid datapoints from a distribution on $\mathbb R^d \times \{\pm 1\}$. It is now well-known (e.g see \cite{bubecknetsize,vershynin2020} and references therein) that a two-layer neural network (NN) $f_{W,v}:\mathbb R^d \to \mathbb R$ with appropriate choice of activation function $\sigma:\mathbb R \to \mathbb R$ and sufficiently many hidden neurons $k$ (the network's \emph{width}), defined for input $x \in \mathbb R^d$ by
\begin{eqnarray}
f_{W,v}(x) := \sum_{j=1}^k v_j\sigma(x^\top w_j),\text{ with }W=(w_1,\ldots,w_k) \in \mathbb R^{k \times d},\,v=(v_1,\ldots,v_k) \in \mathbb R^k,
\label{eq:tlnn}
\end{eqnarray}
can perfectly fit the dataset $\mathcal D_n$ in the sense that $y_i = f_{W,v}(x_i)$ for all $i \in [n]$.
For example, this can be done with $k \gtrsim n/d$ and the thresholding activation ~\cite{baum88} or ReLU ~\cite{bubecknetsize}. In the model \eqref{eq:tlnn}, each $w_j \in \mathbb R^d$ are the \emph{weights} or parameter vector of the $j$th in the hidden layer and $v_j$ is the corresponding output weight for that neuron. If the network is \emph{over-parametrized} in the sense that $k \ge n$, a very smooth / robust interpolation is realizable, in the sense that the input-to-output \emph{Lipschitz constant} is bounded. This is indeed achievable by appropriately tuning each neuron to only handle one datapoint. Robustness is important in many situations, e.g in machine learning applications where the goal is to learn a prediction function (aka model)
which will perform well on unseen data from the same distribution, and so it is reasonable to ask that the predictions of the model be stable w.r.t small perturbations in its input $x$, including \emph{adversarial} perturbations. Recent work ~\cite{lor} hints that over-parametrization might not just be sufficient, but also \emph{necessary} for robustness.

In this work, we study the robustness of (in)finite NNs in the random features (RF) \cite{rf,rf2} and tangent kernel (NTK) regimes \cite{jacot18}. We establish quantitative trade-offs between memorization and robustness in these regimes, as a function of complexity parameters $n$, $d$, and $k$. We also observe for the first time, a \emph{multiple-descent} phenomenon ~\cite{belkin18,Loog20} in the robustness of models in these regimes.

 \paragraph{Notation.} We will use the notation $a_n \gtrsim b_n$ (also written $a_n = \Omega(b_n)$ or equivalently, $b_n = \mathcal O(n)$) to mean that $a_n \ge c b_n$ for some $c>0$ and for sufficiently large $n$, while $a_n \asymp b_n$ means $a_n \gtrsim b_n \gtrsim a_n$. We will use $\widetilde{\Omega}(...)$ to mean $\Omega(...)$ modulo log-factors.
The notation $o(1)$ will be used  to denote a quantity which goes to zero with $n$.
 Probabilistic versions of these notations are written with a subscript $\mathbb P$, for example $\mathcal O_{\mathbb P}(...)$, $o_{\mathbb P}(...)$, etc. The acronym \emph{a.s} means \emph{almost-surely}, \emph{a.e} means \emph{almost-everywhere}, \emph{w.p} means \emph{with probability}, and \emph{w.h.p} means \emph{with high probability}. The $L_p$-norm of a finite-dimensional vector $w$ is denoted $\|w\|_p$. We will write $\|w\|$ to mean $\|w\|_2$.

\subsection{Problem setup}
\label{subsec:setup}
\paragraph{Generic dataset.}
Suppose the distribution $P$ of the dataset $\mathcal D_n$ is supported on $\sphere \times \mathbb R$, where $\sphere := \{x \in \mathbb R^d \mid \|x\| = 1\}$ is the the unit-sphere in $\mathbb R^d$, and the marginal distribution of each $x_i$ is $\tau_d$, the uniform distribution on $\sphere$. Given a function $f:\sphere \to \mathbb R$ (e.g a neural network), its training error is defined by $\widehat{\varepsilon}_n(f):=(1/n)\sum_{i=1}^n(f(x_i)-y_i)^2$ and its generalization error is $\varepsilon_{\mathrm{test}}(f) := \mathbb E_{(x,t) \sim P}[(f(x)-t)^2]$.


\begin{restatable}[Bayes-optimal error]{df}{}
Let $\varepsilon^\star_{\mathrm{test}} \ge 0$ denote the Bayes-optimal error for the problem, that is
$\varepsilon_{\mathrm{test}}^\star := \inf_f \varepsilon_{\mathrm{test}}(f)$,
where the infimum is taken over all measurable functions $f:\sphere \to \mathbb R$.
\label{df:bayesopt}
\end{restatable}
For concreteness, consider the linear data-generating process with
\begin{eqnarray}
\label{eq:noisylinear}
y_i = w_0^\top x_i + z_i,\text{ for }i \in [n]
\end{eqnarray}
where $w_0 \in \mathbb R^d$ with $\|w_0\| \le 1$ and $z_1,\ldots,z_n$ is an iid sequence of label-noise from $\mathcal N(0,\zeta^2)$, independent of the $x_i$'s.
A simple calculation reveals then reveals that the Bayes-optimal error for the prediction problem is $\varepsilon^\star_{\mathrm{test}} = \zeta^2$.
Thus, the variance $\zeta^2$ of the label noise completely controls the difficulty of the learning problem. 
To avoid being corner cases, we shall assume the following condition non-degeneracy condition.
\begin{restatable}[Labels are not a deterministic function of inputs]{cond}{}
All through this manuscript, we will assume that
$\varepsilon^\star_{\mathrm{test}} \ge \varepsilon_0$,
for some absolute constant $\varepsilon_0 \in (0, 1/2]$.
\end{restatable}
A dataset $\mathcal D_n$ verifying the above condition will be referred to as a \emph{generic dataset}. In ~\cite{lor}, the authors considered the noise-only scenario where the $y_i$'s are uniformly distributed in $\{\pm 1\}$ and are completely independent of the $x_i$'s, which in our notations, corresponds to taking $w_0 = 0$ and $\zeta = $1.
\label{ex:linearmodel}
For later use, let $X:=(x_1,\ldots,x_n) \in \mathbb R^{n \times d}$ be the corresponding \emph{design matrix} and let $y=(y_1,\ldots,y_n) \in \mathbb R^n$ be the corresponding sequence of training targets / labels.

\begin{restatable}[Memorization]{df}{}
Given $\varepsilon \ge 0$, $f$ is said to $\varepsilon$-memorize the dataset $\mathcal D_n$ if $\widehat{\varepsilon}_n(f) \le \varepsilon$; by convention, if $\varepsilon \le \varepsilon^\star_{\mathrm{test}}/2$ (or any other absolute fraction), we simply say $f$ memorizes $\mathcal D_n$.
\label{df:memo}
\end{restatable}
Thus, memorization essentially refers to a model which minimizes the training error way beyond the (Bayes) optimal test-error. It turns out that price of doing this is a degradation in robustness, as measured a sense that will become clear in a bit. Moreover, and as one would expect, this price grows with the sample size.

\subsection{Prior works}
\paragraph{Is over-parametrization necessary for robustness ?}
Recently, it has been conjectured in \cite{lor} that
over-parametrization is not just \emph{sufficient} for robustness, but also \emph{necessary}.
More precisely, suppose the activation function $\sigma$ is $1$-Lipschitz and adopting the notation of \cite{lor}, let $\mathcal F_{d,k}(\sigma) := \{f_{W,v} \mid W \in \mathbb R^{k \times d},\,v \in \mathbb R^k\}$
be the set of all two-layer neural networks of width $k$, input dimension $d$, and activation function $\sigma$.

\begin{conj}[~\cite{lor}]
It holds with high probability over the dataset $\mathcal D_n$ that any $f \in \mathcal F_{d,k}(\sigma)$ which memorizes $\mathcal D_n$ must satisfy
$\Lip(f) \ge \widetilde{\Omega}(\sqrt{n/k})$. Thus, in order for $\mathcal F_{d,k}(\sigma)$ to contain a neural network which smoothly interpolates $\mathcal D_n$, it must be over-parametrized, i.e $k \gtrsim n$ hidden neurons are required.
\label{conj:bubeck}
\end{conj}
Recall that the \emph{Lipschitz constant} of a function $f:\sphere \to \mathbb R$ is defined by
\begin{eqnarray}
\Lip(f) := \sup_{x,x' \in \sphere,\;x' \ne x}\dfrac{|f(x)-f(x')|}{\|x-x'\|},
\end{eqnarray}
and measures the \textbf{maximum} absolute change in the output of $f$ as a fraction of the change in its input.
\paragraph{Progress on Conjecture \ref{conj:bubeck}} A number of particular cases of Conjecture \ref{conj:bubeck} were proven in ~\cite{lor}. Most notably, the conjecture was proved in the following regimes
\begin{itemize}
    \item[--]\emph{Low-dimensional under-complete regime where $n \ge d \gtrsim k$.} In this regime, a weaker form of the conjecture was proved with $n$ replaced by $k$ in the lower-bound. More precisely, it was proved in  Theorem 4 of the aforementioned paper that in this case, $\Lip(f) \ge \widetilde{\Omega}(\sqrt{d/k})$ w.p $1-e^{-\Omega(d)}$. The condition $n \ge d$ is crucial for the arguments in that theorem to hold.
\item[--]\emph{Lower-bounding a proxy for $\Lip(f)$.}
The authors also proved a weaker form of the conjecture, in which $\Lip(f)$ of the neural network $f=f_{W,v} \in \mathcal F_{k,d}(\sigma)$, is replaced with an upper-bound $\eta(f)$ defined by $\eta(f) := \sum_{j=1}^k |v_j|\|w_j\|_2$,
which is well-known to be a reasonable measure of complexity for neural networks ~\cite{bartlett98}. \cite{bubeck2018} then proved that
\begin{itemize}
\item With positive probability, any $f \in \mathcal F_{k,d}(\sigma)$ which memorizes generic data must verify $\eta(f) \ge \Omega(\sqrt{n/k})$. We note that such a result does not say anything useful about {Conjecture 1} itself, since $\eta(f)$ is only an upper-bound for $\Lip(f)$, the object the conjecture is ultimately about. 
 \end{itemize}

\item[--]\emph{Converse of the conjecture.} Upper-bounds for the Lipschitz-constant were established in that paper (see {\bfseries Conjecture 2} therein), under different regimes.

\item[--]\emph{The case of bounded network parameters.} Very recently, ~\cite{husain21} studied Conjecture \ref{conj:bubeck} in the restrictive scenario where the parameters of the network are constrained to be bounded. 
\end{itemize}

\subsection{A new measure of robustness: Sobolev-seminorm}
\paragraph{Limitations of Lipschitz constants to study (non)robustness.}
Although a small Lipschitz constant for a model $f$ immediately implies robustness in the sense that small changes in the input $x$ can only cause small changes in the output $f(x)$ (by norm duality), a large Lipschitz constant is uninformative.
Indeed, one can imagine an otherwise very smooth $f$, the norm of whose input-to-output gradient explodes on a subset of the sphere of arbitrarily small measure w.r.t the true distribution of the data.
However, such a model could be perfectly robust (for example if the function is constant outside this "bad" set), in any practical sense.
Thus unlike \cite{lor} which studies the Lipschitz constants, we propose to instead study the models \emph{Sobolev-seminorm}. To simplify (with abuse of language), we study the average norm of the gradient rather than the maximum (i.e the worst-case). 

\paragraph{Sobolev-seminorm as a measure of robustness.}
Let $f:\mathbb R^d \to \mathbb R$ be a function which is continuously-differentiable
in the usual sense, almost-everywhere (a.e). The spherical gradient of $f$ is the map $\nabla_{\sphere} f:\sphere \to T\sphere$ defined for each $x \in \sphere$ by
\begin{eqnarray}
\nabla_{\sphere} f (x) := P_{T_x\sphere}(\nabla f(x)) = P_{x^\perp}(\nabla f (x)) = (I_{d}-xx^\top) \nabla f(x),
\end{eqnarray}
where $\nabla f:\mathbb R^d \to \mathbb R^d$ is the usual euclidean gradient of $f$.
Here $T_x\sphere$ is the tangent space of $\sphere$ at the point $x$; $P_{x^\perp} \in \mathbb R^{d \times d}$ is the projector onto the orthogonal complement of $T_x\sphere$; and $T\sphere := \{(x,z) \mid x \in \sphere,\;z \in T_x\sphere\}$ is the tangent bundle.
\begin{restatable}{df}{}
 Define the Sobolev-seminorm of $f$ w.r.t the uniform $\tau_d$ on $\sphere$, denoted $\mathfrak S(f)$, by
 \begin{eqnarray}
 \mathfrak{S}(f) := \|\nabla_{\sphere} f\|_{L^2(\tau_d)} = \left(\int_{\sphere}\|\nabla_{\sphere} f(x)\|^2\dif\tau_d(x)\right)^{1/2}.
 \end{eqnarray}
\end{restatable}
Thus, $\mathfrak{S}(f)^2$ is the average squared $L_2$-norm of the input-to-output gradient of $f$. From the above definition, is clear that
\begin{mdframed}
$
\Lip(f)
\ge \mathfrak{S}(f)
$.
Thus, lower-bounds on $\mathfrak{S}(f)$ immediately translate to lower-bounds on $\mathfrak{S}(f)$ and upper-bounds on $\Lip(f)$ translate to upper-bounds on $\mathfrak{S}(f)$.
\end{mdframed}
\begin{restatable}[The case of non-differentiable functions]{rmk}{}
Our restriction to differentiable $f:\mathbb R^d \to \mathbb R$ is only artificial. In case $f$ is non-differentiable, we may replace pointwise gradient-norm $\|\nabla_{\sphere} f(x)\|_2$ in the above definitions with the \emph{strong slope}~\cite{degiorgi80,hoffmanlsc, nonlinearhoffman} $|\nabla^-f|(x)$, defined by
$
|\nabla^-f|(x) := \limsup_{x' \to x}\dfrac{(f(x)-f(x'))_+}{\|x-x'\|}.
$
In particular, if $f$ is differentiable at $x$, then $|\nabla^- f|(x) = \|\nabla_{\sphere} f(x)\|$.
\end{restatable}

\subsection{Summary of main contributions}
\label{sec:contribsum}
We consider two-layer neural networks \eqref{eq:tlnn} in the random features (RF) \cite{rf,rf2} and neural tangent kernel (NTK) \cite{jacot18} regimes, in both the finite-width and infinite-width settings and establish a number of theorems which highlight a fundamental tradeoff between memorization and robustness.

\paragraph{Tradeoffs between memorization and robustness.}
We establish explicit tradeoffs between memorization and robustness in the following form, valid for a large class of models including but not limited to models obtained via ridge(less) regression (the so-called \emph{representer subspace}\footnote{The concept of "representer subspace" is formally defined in \eqref{eq:kernspan}. As we shall see, different regimes (finite / infinite-width RF, finite / infinite-width NTK, etc.) of neural networks will give rise to different kernels and different representer subspaces.}),
\begin{eqnarray}
\underbrace{\mathfrak{S}(f)}_{\text{Nonrobustness}} \ge \underbrace{(\varepsilon^\star_{\mathrm{test}}-\widehat{\varepsilon}_n(f))}_{\mathrm{Memorization}}\cdot \underbrace{\widetilde{\Omega}(\sqrt{n'})}_{\text{"Sample size"}}
\label{eq:template}
\end{eqnarray}
where $n'=n$ for infinite-width RF / NTK and finite-width RF, and $n' = n/k$ for finite-width NTK. Numerous experiments confirm our theoretical results. Moreover, the bounds \eqref{eq:template} are tight: they are attained by the min-norm interpolator with certain choices of activation function $\sigma$ (including the ReLU), and setting scaling regimes for $n$, $d$, and $k$.

To give a sense of these results, let us restrict to models that memorize the dataset $\mathcal D_n$.
Recall from Definition \ref{df:memo} that a model $f$ memorizes $\mathcal D_n$ if its training error $\widehat{\varepsilon}_n(f)$ is a constant short of the Bayes-optimal test error $\varepsilon^\star_{\mathrm{test}}$. For such models, we may breakdown \eqref{eq:template} as follows.
 \begin{itemize}
     \item[--] \emph{{Infinite-width RF and NTK.}}
     In the high-dimensional setting $d>n$, we consider infinite-width neural networks in the RF or NTK regimes and prove that for a variety of activation functions including the ReLU, and prove in Theorem \ref{thm:infiniteRFNTKLaw} that with probability tending to $1$, any model in the induced "representer subspace" which memorizes dataset $\mathcal D_n$ must verify $\mathfrak{S}(f) \ge \Omar(\sqrt{n})$ (ignoring log-factors). Moreover, we establish in Theorem \ref{thm:infiniteRFNTKLowerbound} the tightness of this lower-bound: it is attained by the min-norm / least squares model with ReLU activation function.
     \item[--] \emph{{Finite-width RF.}} Consider finite-width neural networks \eqref{eq:tlnn} in the RF, where the parameter vector $w_j$ of each neuron is sampled iid from the uniform-distribution on the unit-sphere $\sphere$, and only the output weights $v \in \mathbb R^k$ learned. For $d \asymp k$, we prove in Theorem \ref{thm:finiteRFLaw} that for a large class of activation functions including the ReLU, that with probability tending to $1$, any model in the induced "representer space" which memorizes dataset $\mathcal D_n$ must verify $\mathfrak S(f) \ge \widetilde{\Omega}(\sqrt{n})$ (ignoring log-factors). Moreover, we show in Theorem \ref{thm:law}: it is attained by the min-norm / least squares interpolator with ReLU activation function.
     \item[--] \emph{{Given hidden weights.}}
     Fo any choice of hidden weights matrix $W=(w_1,\ldots,w_k) \in \mathbb R^{k \times d}$, we prove in Theorem \ref{thm:frozenlaw} that w.p tending to $1$: any two-layer model in the induced "representer subspace" which memorizes $\mathcal D_n$ satisfies $\mathfrak{S}(f) \ge \widetilde{\Omega}(\sqrt{1/\alpha_\sigma(W)}\sqrt{nd/k})$, where $\alpha_\sigma(W)$ is a kind of \emph{condition number} of $W$ w.r.t the activation function $\sigma$. For example, for the identity activation function, we have $\alpha_{\sigma_{\mathrm{id}}}(W) = \condnumb(W)^2$,
     where $\condnumb(W)$ is the usual / linear-algebraic condition number of $W$.
     \item[--] \emph{{Finite-width NTK.}}
     Consider finite-width two-layer neural network \eqref{eq:tlnn} in the NTK regime, where the parameter vector $w_j$ of each neuron is as previously. In the setting where $d \asymp k$, we prove in Theorem \ref{thm:finiteNTKLaw} that with probability tending to $1$, any model in the induced "representer subspace" which memorizes the dataset $\mathcal D_n$ must verify $\mathfrak{S}(f) \ge \widetilde{\Omega}(\sqrt{n/k})$.
\end{itemize}
These results, stated more formally in the following sections, are empirically confirmed by experiments in section \ref{sec:exp}.
\paragraph{Multiple-descent in robustness.}
We empirically observe \emph{multiple-descent} \cite{doubledescent,Loog20,ascoli20,Mei2019,adlam20} in robustness of two-layer NNs in the above linearized regimes. Refer to Figure \ref{fig:multiple}. To the best of our knowledge, this is the first time such a phenomenon has been observed. We speculate that the multiple-descent phenomenon occurs for a variety of statistical functionals (here robustness) of NNs, other than their generalization error, which is currently under intensive research in the theoretical machine learning community.

\section{Preliminaries}
\label{sec:theorems}

\subsection{Warmup: ordinary linear models}
Consider the linear data generating process in Example \ref{ex:linearmodel}, with label noise variance $\zeta^2 > 0$. Note that the Bayes-optimal error for the problem is $\varepsilon^\star_{\mathrm{test}} = \zeta^2$. We start with a result on linear models, which already illustrates that the price of memorization is robustness.

\begin{restatable}[Law of robustness for ordinary linear models]{thm}{linear}
For sufficiently large $n$ and $d$, the following holds w.p $1-n^{-\Omega(1)}$ over $\mathcal D_n$: every linear model $g_w$ which $\varepsilon$-memorizes $\mathcal D_n$ verifies $\Lip(g_w) \ge \mathfrak S(g_w) \gtrsim (\varepsilon^\star_{\mathrm{test}}-\varepsilon)\sqrt{n}$.
In particular, for the high-dimensional regime $d>n$, the min-norm / least squares interpolator $\widehat{g}=g_{\widehat{w}}$, defined by setting $\widehat{w} = X^\top (XX^\top)^{-1}y \in \mathbb R^d$, satisfies $\widetilde{\Omega}(\varepsilon^\star_{\mathrm{test}}\sqrt{n}) \le \mathfrak{S}(\widehat{g}) \le \mathcal O(\varepsilon^\star_{\mathrm{test}}\sqrt{n})$ w.p $1-n^{-\Omega(1)}$.
\label{thm:linear}
\end{restatable}
\begin{restatable}{rmk}{}
We note that the second part of the above result was established in ~\cite{lor}, at least for noise-only data where $w_0=0$.
\end{restatable}


The proof of Theorem \ref{thm:linear} (provided in the appendix) is based on standard Rademacher complexity-based generalization bounds for squared loss.
The theorem highlights a clear tradeoff between the memorization error $\varepsilon^\star_{\mathrm{test}}-\widehat{\varepsilon}_n(g_w)$ of a linear model $g_w:x \mapsto x^\top w$, and its robustness as measured by its Sobolev-seminorm $\mathfrak S(g_w)$. In the sequel, we will obtain results of this sort, for linearized neural nets like random features and neural tangent models, in both finite and infinite-width regimes.
\subsection{Kernelization}
The rest of the manuscript will be concerned with the complicated case of neural networks in various linearized regimes. We will employ the language and toolbox of kernel methods in order to give a unified treatment.
\label{sec:prelim}

\paragraph{Reproducing Kernel Hilbert Spaces (RKHS).}
Consider a continuous positive-definite definite kernel $K:\sphere \times \sphere \to \mathbb R$, where positive-definiteness means that $\sum_{i=1}^N\sum_{\ell=1}^N b_i b_\ell K(x'_i,x'_\ell) \ge 0$ for every finite sequence $x'_1,\ldots,x'_N \in \sphere$ and every $c_1,\ldots,c_N \in \mathbb R$.
Let $\mathcal H_K$ be the Reproducing Kernel Hilbert Space (RKHS) induced by $K$. Note that $\mathcal H_K \subseteq L^2(\tau_d)$ since $K$ is a Mercer kernel\footnote{By continuity of $K$ and compactness of the unit-sphere $\sphere$}. Let $K(X,X) \in \mathbb R^{n \times n}$ be the kernel gram matrix with entries $K(x_i,x_j)$, where $X=(x_1,\ldots,x_n) \in \mathbb R^{n \times d}$ is the design matrix associated for the generic dataset $\mathcal D_n := \{(x_1,y_1),\ldots,(x_n,y_n)\}$ in \eqref{eq:noisylinear}.

\paragraph{The "representer subspace".}
We will denote by $\lspan_K(X) \subseteq \mathcal H_K$ the subspace of functions in $\mathcal H_K$, formed by linear combinations of the  $n$ functions $x \mapsto K(x_i,x)$, that is
\begin{eqnarray}
\begin{split}
\lspan_K(X) &:= \lspan(\{K(x_1,\cdot),\ldots,K(x_n,\cdot)\}) = \{f_c:=\sum_{i=1}^nc_iK(x_i,\cdot)\mid c \in \mathbb R^n\}.
\end{split}
\label{eq:kernspan}
\end{eqnarray}
As we shall see, different regimes (finite / infinite-width RF, finite / infinite-width NTK, etc.) of neural networks will give rise to different kernels and different representer subspaces.

It is a classical result that the RKHS norm of any $f_c \in \lspan_K(X)$ is given by the simple formula
\begin{eqnarray} \|f_c\|_{\mathcal H_K} = \sqrt{c^\top K(X,X) c}.
\end{eqnarray}
Also, from the so-called \emph{Generalized Representer Theorem (GRT)} \cite{scholkopf2001generalized} (also see appendix for a statement), $\lspan_K(X)$ contains all models which can be constructed by doing certain kinds of penalized kernel regression in $\mathcal H_K$.

\begin{restatable}[\cite{scholkopf2001generalized} Generalized representer theorem for the sphere]{prop}{}
If $g:\mathbb R_+ \to \mathbb R$ is a strictly increasing function and $\ell_n:(\sphere \times \mathbb R)^n \to \mathbb R$ is an arbitrary "cost function", then every minimizer of the functional
$$
\mathcal H_K \to \mathbb R,\, f \mapsto \ell_n((x_1,y_1,f(x_1)),\ldots,(x_n,y_n,f(x_n)) + R(\|f\|_{\mathcal H_K})
$$
is an element of $\lspan_K(X)$.
\label{prop:representer}
\end{restatable}
Some notable functions that lie in $\lspan_K(X)$ include
\begin{itemize}
    \item The least-squares model $\widehat{f}_n$,
    defined by $\widehat{f}_n(x)=\widehat{c}_n^\top K(X,x)$  with $\widehat{c}_n := K(X,X)^{-1}y$ (provided the kernel gram matrix $K(X,X)$ is invertible). This is precisely the element of $\lspan_K(X)$ with minimal RKHS norm, and corresponds to taking $\ell_n((x_1,y_1,f(x_1)),\ldots,(x_n,y_n,f(x_n)) := \sum_{i=1}^n (f(x_i)-y_i)^2$ and $R \equiv 0$.
    \item Any function in the so-called \emph{version space} $\{f \in \mathcal H_K \mid f(x_i) = y_i\,\forall i \in [n]\}$. This corresponds to the same choise of $\ell_n$ and $g$ as above. Note that the least-squares estimator above (when it exists) is itself an element of the version space.
    \item Any ridge interpolator $\widehat{f}_{n,\lambda}$, namely any function of the form $\widehat{f}_{n,\lambda}(x) = \widehat{c}_{n,\lambda}^\top K(X,x)$ for all $x \in \sphere$, where $\widehat{c}_{n,\lambda} := (K(X,X)+n\lambda I_n)^{-1}y \in \mathbb R^n$, with $\lambda \ge 0$. This corresponds to taking the cost function $\ell_n$ as in the previous example, and $R(\|f\|_{\mathcal H_K}) = \lambda \|f\|_{\mathcal H_K}^2$.
    \item etc.
\end{itemize}

We will conveniently exploit this universal property of $\lspan_K(X)$ to give a unified treatment for robustness in dot-product kernels, finite / infinite-width NNs in RF and NTK regimes, etc. (see Table \ref{tab:kernels}), by reducing to questions about the extreme eigenvalues of certain random matrices (including those of the kernel gram matrix $K(X,X)$).


\begin{table}[h!]
  \begin{center}
       {\tabulinesep=1.2mm
    \begin{tabu}{|c|c|}
    \hline
       Model class & Equivalent kernel on unit-sphere\\
      \hline
        General dot-product RKHS & $K(x,x'):=\phi(x^\top x')$, $t:=x^\top x'$, $\phi \in \mathcal C^0([-1,1] \to \mathbb R)$\\
        \hline
        Laplace RKHS & $K_{\mathrm{Lap}}(x,x') = \phi_{\mathrm{Lap}}(t):=e^{-\sqrt{2-2t}/s}$, $s>0$\\
        \hline
        Gaussian RKHS & $K_{\mathrm{Gauss}}(x,x') = \phi_{\mathrm{Gauss}}(t):=e^{-(2-2t)/s^2}$\\
      \hline
        Infinite-width RF & $K^\infty_\rf(x,x')=\mathbb E_w[\sigma(x^\top w)\sigma(w^\top x')]$, with $w \sim \tau_d$\\
\hline
        Infinite-width NTK & $K^\infty_\ntk(x,x')=(x^\top x')\mathbb E_w[\sigma'(x^\top w)\sigma'(w^\top x')]$ \\
\hline
        General feature-based & $K_\Phi(x,x')=\langle \Phi(x),\Phi(x')\rangle $, with $\Phi:\sphere \to \mathcal H_0$ \\
        \hline
        Finite-width RF & $K_\rf(x,x')=\Phi_\rf(x)^\top\Phi_\rf(x')$\\
    \hline
        Finite-width NTK & $K_\ntk(x,x')=\Phi_\ntk(x)^\top \Phi_\ntk(x')$ \\
\hline
    \end{tabu}
    }
  \end{center}
    \caption{Kernel reformulation of different model classes. This allows us to give a unified treatment of otherwise very disparate situations, by considering the "representer" subspace of induced by the appropriate kernel function $K$.}
\label{tab:kernels}
\end{table}

\section{Law of robustness for general dot-product kernels}
Consider the case of a dot-product kernel $K:\sphere \times \sphere \to \mathbb R$ given by $K(x,x') \equiv \phi(x^\top x')$ for some continuous $\phi:[-1,1] \to \mathbb R$. Further, we impose the following regularity condition
\begin{restatable}{cond}{}
$\phi$ is thrice continuously-differentiable at $0$ and verifies $\phi'(0) \ne 0$.
  \label{cond:awfulkappa}
\end{restatable}
For example, this is the case for exponential-type kernels like the Gaussian kernel on the unit-sphere $\sphere$, which is known to be the infinite-width version of the Fourrier random features \cite{rf,rf2}; infinite-width RF and NTK kernels corresponding to the ReLU activation function \cite{inductivebietti}; etc. 

Let $\varepsilon^\star_{\mathrm{test}}$ be the Bayes-optimal error for the underlying squared-loss regression problem (see Definition \ref{df:bayesopt}) and let $\varepsilon$ be any error threshold in the interval $[0,\varepsilon^\star_{\mathrm{test}})$.
The following is our first main result.
  \begin{restatable}[Law of robustness for dot-product kernels]{thm}{lawdotprodkernel}
    Under Condition \ref{cond:awfulkappa}, in the limit $n,d \to \infty$ such that $n/d \le \gamma_1 < 1$, it holds w.p tending to $1$ that: every $f \in \lspan_K(X)$ which $\varepsilon$-memorizes $\mathcal D_n$ satisfies $\mathfrak{S}(f) \ge \widetilde{\Omega}((\varepsilon^\star_{\mathrm{test}}-\varepsilon)\sqrt{n})$.
In particular, if the kernel gram matrix $K(X,X)$ is nonsingular, then the min-norm interpolator $\widehat{f}_n(x) := K(X,x)^\top K(X,X)^{-1} y$ satisfies 
$\mathfrak{S}(\widehat{f}_n) \ge \widetilde{\Omega}(\varepsilon^\star_{\mathrm{test}} \sqrt{n})$.
\label{thm:lawdotprodkernel}
\end{restatable}
Like all our other results, the proof is deferred to the appendix. It uses tools from probability theory like the spherical \emph{Poincar\'e} inequality~\cite{ledoux,Gozlan2015,Villani} and random matrix theory (RMT) ~\cite{rmt,elkaroui2010}.
    
\subsection{Law of robustness for infinite-width random features  and neural tangent kernel}
Now, consider an infinite-width (i.e having $k=\infty$ hidden neurons) neural network in the RF or NTK regime. The learning problem is reduced to RKHS regression with kernels $K^\infty_{\rf/\ntk}:(\sphere)^2 \to \mathbb R$,
\begin{eqnarray}
\begin{split}
K_\ntk^\infty(x,x') &:= (x^\top x')\mathbb E_{w}[\sigma'(x^\top w)\sigma(w^\top x')],\quad\quad\quad\text{ (1st layer-only kernel)},\\
K_\rf^\infty(x,x') &:= \mathbb E_{w}[\sigma(x^\top w)\sigma(w^\top x')], \quad\quad\quad\quad\text{ (2nd layer-only kernel)}
\end{split}
\label{eq:rfntkkernelfunc}
\end{eqnarray}
for $w \sim \tau_d$ (see \cite{inductivebietti}, e.g). Different choices for the activation function $\sigma$ give rise to different kernels. For example, in the case of the ReLU activation, the corresponding 
kernels are given by
\begin{eqnarray}
K_{\rf/\ntk}^\infty(x,x') = \begin{cases}
(x^\top x')\phi_0(x^\top x'),&\mbox{ if }\text{ NTK (i.e 1st  layer-only kernel)},\\\phi_1(x^\top x'),&\mbox{ if }\text{ RF (i.e 1st layer-only kernel)},
\label{eq:arccos}
\end{cases}
\end{eqnarray}
where $\phi_0(t) = \pi^{-1}\arccos(-t)$ and 
$\phi_1(t) = \pi^{-1}(t\arccos(-t)+\sqrt{1-t^2})$
are the arc-cosine kernels of order 0 and 1 respectively \cite{inductivebietti}. In this particular case, both $K_\rf^\infty$ and $K_\ntk^\infty$ are dot-product kernels.

\begin{restatable}{cond}{}
$K_{\rf/\ntk}^\infty$ is a dot-product kernel by means of a continuous function $\phi_{\rf/\ntk}^\infty:[-1,1] \to \mathbb R$ which is thrice continuously-differentiable at $0$ w.p $(\phi_{\rf/\ntk}^\infty)'(0) > 0$.
\label{cond:nonlin}
\end{restatable}
For example, the absolute-value activation function fails to satisfy this condition. On the other hand, the ReLU, tanh, and the gaussian error-function (erf) satisfy the condition. Table 1 of \cite{louart2018} provides explicit formula for $\Phi^\infty_{\rf/\ntk}$ for a variety of activation functions, including: ReLU, absolute-value, sign, sin, cos, gaussian erf, etc.

When the kernel gram matrix $K_{\rf/\ntk}^\infty(X,X) \in \mathbb R^{n \times n}$ is invertible (which happens for example in the high-dimensional regime $d > n$ with ReLU activation function), let $\widehat{f}_{\rf/\ntk}^\infty$ be the min-norm interpolator defined by (see ~\cite{aroraexact,justinterpolate,Hastie2019})
\begin{eqnarray}
\widehat{f}_{\rf/\ntk}^\infty(x) := K_{\rf/\ntk}^\infty(X,x)^\top K_{\rf/\ntk}^\infty(X,X)^{-1}y,\,x \in \sphere.
\label{eq:infiniteRFNTKminnorm}
\end{eqnarray}



The following theorem is an important corollary to Theorem \ref{thm:lawdotprodkernel},  and establishes a quantitative tradeoff between memorization and robustness for two-layer neural networks \eqref{eq:tlnn} in the infinite-width RF and NTK regimes. The proof of the theorem (given in the appendix) makes use of the spherical Poincar\'e inequality \cite{ledoux,Villani,villaniTopics,Gozlan2015}, together with the classical generalization theory for kernel methods \cite{boucheron05}.
\begin{restatable}[Law of robustness for infinite-width RF / NTK]{thm}{infiniteRFNTKLaw}
Assume Condition \ref{cond:nonlin}. In the limit $n,d \to \infty$ such that $n/d \to \gamma_1$ for some $\gamma_1 \in [0, 1)$, the following holds w.p $1-n^{-\Omega(1)}$ over $\mathcal D_n$: every $f \in \lspan_{K^\infty_{\rf/\ntk}}(X)$ which $\varepsilon$-memorizes $\mathcal D_n$ satisfies
$
\mathfrak{S}(f) \ge \Omar((\varepsilon^\star_{\mathrm{test}}-\varepsilon)\sqrt{n})$.
In particular, the min-norm interpolator $\widehat{f}_{\rf/\ntk}^\infty$ satisfies 
$\mathfrak{S}(\widehat{f}_{\rf/\ntk}^\infty) \ge \Omar(\varepsilon^\star_{\mathrm{test}} \sqrt{n})$ w.p $1-o(1)$.
\label{thm:infiniteRFNTKLaw}
\end{restatable}
In light of the above theorem, in the high-dimensional $d>n$ regime, it might thus be dangerous to do min-norm / unpenalized interpolation with infinite-width NNs as advocated in ~\cite{justinterpolate}. Regularization should be used to select a good tradeoff between fit and robustness. This is also empirically confirmed in section \ref{sec:exp}.

\begin{proof}[Proof of Theorem \ref{thm:infiniteRFNTKLaw}]
Follows directly from Theorem \ref{thm:lawdotprodkernel} with the kernel $K$ taken to be the dot-product kernel $K^\infty_{\rf/\ntk}:(x,x') \mapsto \phi^\infty_{\rf/\ntk}(x^\top x')$.
\end{proof}


\paragraph{A matching upper-bound: the min-norm interpolator.}
Still in the high-dimensional setting where both $n$ and $d$ are large with $d \ge n$, we now establish the tightness of the lower-bounds in Theorem \ref{thm:infiniteRFNTKLaw} for RF approximation. To this end, we will prove that in the case of the ReLU activation function, the reverse bound is satisfied by the min-norm  estimator $\widehat{f}_\rf$!
First, we must ensure that the kernel gram matrix $K^\infty_\rf(X,X)$ is invertible w.h.p, when the activation function $\sigma$ is the ReLU. The following lemma establishes that its eigenvalues are contained in a finite-closed interval not containing zero, with probability tending to $1$, a crucial ingredient for our upper-bound.
\begin{restatable}[Invertibility of RF kernel gram matrix]{lm}{invertibleK}
For the ReLU activation function and for sufficiently large $n$ and $d$ with $n/d \le \gamma_1<1$, it holds w.p $1-d^{-1+o(1)}$ that the eigenvalues $\lambda_1 \ge \ldots \ge \lambda_n$ of the RF kernel gram matrix $K_\rf^\infty(X,X)$ satisfy $c \le \lambda_n \le \ldots \le  \lambda_1 \le C$,
for constants $c,C>0$ which only depend on $\gamma_1$. In particular, $K_\rf^\infty(X,X)$ is invertible w.p $1-d^{-1+o(1)}$.
\label{lm:invertibleK}
\end{restatable}
\vspace{-.4cm}
Equipped with this lemma, following result establishes tightness of the lower-bound in Theorem \ref{thm:infiniteRFNTKLaw}.
\begin{restatable}[Tightness of lower-bound in Theorem \ref{thm:infiniteRFNTKLaw}]{thm}{infiniteRFNTKLowerbound}
For sufficiently large $n$ and $d$ such that $n/d \le \gamma_1 < 1$, it holds w.p $1-n^{-\Omega(1)}$ over $\mathcal D_n$ that the RF min-norm interpolator $\widehat{f}_\rf^\infty$ defined in \eqref{eq:infiniteRFNTKminnorm} with ReLU activation function verifies
$\mathfrak{S}(\widehat{f}_\rf^\infty) \le \Lip(\widehat{f}_{\rf}^\infty) \le \mathcal O(\varepsilon^\star_{\mathrm{test}}\sqrt{n})$.
\label{thm:infiniteRFNTKLowerbound}
\end{restatable}
The proof of the theorem is given in the appendix.

\section{Laws of robustness for feature-based kernels}
Let $\mathcal H_0$ is a separable Hilbert space. For concreteness, take $\mathcal H_0 = \mathbb R^m$. A continuous embedding mapping $\Phi:\sphere \to \mathbb R^m$ induces a kernel $K_\Phi:\sphere \times \sphere \to \mathbb R$ defined by
\begin{eqnarray}
K_\Phi(x,x') := \langle \Phi(x),\Phi(x')\rangle_{\mathcal H_0},
\label{eq:featurebased}
\end{eqnarray}
which in turn induces an RKHS $\mathcal H_{K_\Phi}$.
The embedding map $\Phi$ may also be referred to as a \emph{dictionary}, with \emph{atoms} $\Phi_j:\sphere \to \mathbb R$ given by $\Phi_j(x) := (\Phi(x))_j$.

\begin{restatable}[Condition number of embedding $\Phi$]{df}{}
The condition number of the embedding map $\Phi:\sphere \to \mathbb R^m$, denoted $\alpha_\Phi$, is defined by
\begin{eqnarray}
\alpha_\Phi = \dfrac{\|\Phi\|_{L^2(\tau_d)}^2}{\lambda_{\min}(C_\Phi)}.
\label{eq:alpharatio}
\end{eqnarray}
\end{restatable}

We have the following theorem, a law of robustness for general feature-based models.
\begin{restatable}{thm}{dico}
The following holds w.p $1-n^{-\Omega(1)}$ over the generic dataset $\mathcal D_n$: every $f \in \lspan_{K_\Phi}(X)$ which $\varepsilon$-memorizes the generic dataset $\mathcal D_n$ satisfies
$\mathfrak{S}(f) \ge \Omar((\varepsilon^\star_{\mathrm{test}}-\varepsilon)\sqrt{\dfrac{n}{\alpha_\Phi}})$.

In particular, if the gram matrix $K_\Phi(X,X):=\Phi(X)\Phi(X)^\top$ is invertible (which necessarily implies $n \le d$), then w.p $1-n^{-\Omega(1)}$  over $\mathcal D_n$ it holds that the least-squares model $\widehat{f}_\Phi(x):=K_\Phi(X,x)^\top 
K_\Phi(X,X)^{-1}y$ satisfies $\mathfrak{S}(\widehat{f}_\Phi) \ge \Omar((\varepsilon^\star_{\mathrm{test}}-\varepsilon)\sqrt{\dfrac{n}{\alpha_\Phi}})$.
\label{thm:dico}
\end{restatable}

\subsection{Ordinary linear models (again)}
A remarkable property of the bound in \ref{thm:dico} is that its is completely free of the design matrix $X$. To illustrate the potential benefit of this, reconsider the ordinary linear model from section \ref{subsec:linearkernel}. This is equivalent to taking $m=d$ and $\Phi(x) = \Phi_{\mathrm{id}}(x) := x$ for all $x \sim \sphere$. One easily computes $\mathbb E_{x \sim \tau_d}[\|\Phi_{\mathrm{id}}(x)\|^2]=\mathbb E[\|x\|^2] = 1$ and $C_{\Phi_{\mathrm{id}}}=\cov_{x \sim \tau_d}(\sqrt{d} x) = I_d$, so that $\lambda_{\max}(C_{\Phi_{\mathrm{id}}}) = 1$. We deduce that
\begin{eqnarray}
\alpha_{\Phi_{\mathrm{id}}}=1 \le \dfrac{\lambda_{\max}(XX^\top)}{\lambda_{\min}(XX^\top)} = \mathrm{cond}(X)^2,
\end{eqnarray}
where $\mathrm{cond}(X)$ is the \emph{condition number of} the design matrix $X$.
We thus have the following improved version of Corollary \ref{cor:linearlaw}.
\begin{restatable}[Law of robustness for linear model (improved bound)]{cor}{}
For every $\varepsilon \in [0,\varepsilon^\star_{\mathrm{test}})$, the following holds with probability $1-n^{-\Omega(1)}$ over the generic dataset $\mathcal D_n$: every $f \in \lspan_{K_{\mathrm{id}}}(X)$ which $\varepsilon$-memorizes $\mathcal D_n$ satisfies
$ \mathfrak{S}(f) \ge \Omar((\varepsilon^\star_{\mathrm{test}}-\varepsilon)\sqrt{n})$.
In particular, if $n/d \le \gamma_1 < 1$, then for the min-norm  interpolator $\widehat{f}_n(x):=x^\top X^\top(XX^\top)^{-1}y$, it holds w.p $1-n^{-\Omega(1)}$ over $\mathcal D_n$ that $\mathfrak{S}(\widehat{f}_n) \ge \Omar(\varepsilon^\star_{\mathrm{test}}\sqrt{n})$.
\label{cor:Phila}
\end{restatable}
Just as in the case of Theorem \ref{thm:linear}, the above result is tight because a $\sqrt{n}$ upper-bound for the Lipschitz constant of the min-norm  interpolator was obtained in \cite{lor}.

\subsection{Finite-width networks with prescribed hidden weights}
Let $f_{W,v} \in \mathcal F_{d,k}(\sigma)$ be two-layer neural network on the unit-sphere $\sphere$, with $k$ hidden neurons, activation function $\sigma: \mathbb R \to \mathbb R$, as defined in \eqref{eq:tlnn}. Fix the hidden weights matrix $W \in \mathbb R^{k \times d}$ (for example, consider a random matrix or a pretrained matrix), and consider the subset $\mathcal F_W(\sigma) \subseteq \mathcal F_{k,d}(\sigma)$ of two-layer neural networks \eqref{eq:tlnn} with hidden weights matrix fixed at $W$. The embedding function
\begin{eqnarray}
\Phi_W:\sphere \to \mathbb R^k,\,\Phi_W(x) = \dfrac{1}{\sqrt{k}}\sigma(Wx) := \dfrac{1}{\sqrt{k}}(\sigma(x^\top w_1),\ldots,\sigma(x^\top w_k))
\label{eq:frozenPhiW}
\end{eqnarray}
induces a kernel
$K_W(x,x') = \Phi_W(x)^\top \Phi_W(x') = \dfrac{1}{k}\sum_{j=1}^k\sigma(x^\top w_j)\sigma(w_j^\top x')$. This is an instance of \eqref{eq:featurebased} with $\mathcal H_0 = \mathbb R^k$ and $\Phi$ given by \eqref{eq:frozenPhiW}.
Note that both $\Phi_W$ and $K_W$ depend on frozen value of $W$.
Later in this section, we will consider the scenario where the hidden weights matrix $W$ is random.
Let $C_\sigma(W) \in \mathbb R^{k \times k}$ be the covariance matrix of $\sqrt{d}\sigma(Wx) \in \mathbb R^{k}$ for $x \sim \tau_d$. The following result, one of our main contributions, can be used used to evaluate the robustness of NNs with given hidden weights, e.g trained neural networks.
\begin{restatable}[Law of robustness with given hidden weights]{thm}{frozenlaw}
The following holds with probability $1-n^{-\Omega(1)}$ over the generic dataset $\mathcal D_n$: for every $W \in \mathbb R^{k \times d}$, any $f \in \lspan_{\Phi_W}(X)$ which $\varepsilon$-memorizes $\mathcal D_n$ satisfies
$\mathfrak{S}(f) \ge (\varepsilon^\star_{\mathrm{test}}-\varepsilon)\dfrac{\sqrt{\lambda_{\min}(C_\sigma(W))}}{\|W\|_F}\Omar(\sqrt{nd})
\ge (\varepsilon^\star_{\mathrm{test}}-\varepsilon)\dfrac{\sqrt{\lambda_{\min}(C_\sigma(W))}}{\|W\|_{op}}\Omar(\sqrt{\dfrac{nd}{k}})$.

\label{thm:frozenlaw}
\end{restatable}
 Note that the quantities $\lambda_{\min}(C_\sigma(W))$, $\|W\|_{op}$, and $\|W\|_F$, appearing in the theorem can be computed on unlabeled data. The ratio $\kappa_\sigma(W):=\sqrt{\lambda_{\min}(C_\sigma(W))}/\|W\|_{op}$ 
 can be seen as a kind of (inverse) \emph{condition number} for $W$. Indeed, in the special case where activation function $\sigma$ is the identity, one easily computes $\kappa_{\sigma_{\mathrm{id}}}(W) = \sqrt{\lambda_{\min}(WW^\top)}/\sqrt{\lambda_{\max}(WW^\top)}=\condnumb(W)^{-1}$, where $\condnumb(W)$ is the usual / linear-algebraic condition number of $W$. In the case of random features models where $W$ is frozen at its random value at initialization, random matrix theory (RMT) can used to bound $\kappa_\sigma(W)$ away from zero.

\begin{proof}[Proof of Theorem \ref{thm:frozenlaw}]
The result is Corollary Theorem \ref{thm:dico}. We need to compute the following quantities
\begin{itemize}
    \item $\|\sigma \circ W\|_{L^2(\tau_d)}^2 = \mathbb E_{x \sim \tau_d}[\|\sigma(Wx)\|^2]$, and
    \item $\lambda_{\min}(C_\sigma(W))$, where $C_\sigma(W) \in \mathbb R^{k \times k}$ is the covariance matrix of the random vector $\sqrt{d}\Phi_W(x) = \sqrt{d}\sigma(Wx)$ for $x \sim \tau_d$.
\end{itemize}
Since the activation function $\sigma$ is $1$-Lipschitzness, one easily upper-bound the first quantity as
\begin{eqnarray*}
\begin{split}
\|\sigma \circ W\|_{L^2(\tau_d)}^2 = \mathbb E_{x \sim \tau_d}[\|\sigma(Wx)\|^2] &\le \mathbb E_{x \sim \tau_d}[\|Wx\|^2] = \trace(WW^\top\cov_{x \sim \tau_d}(x)) \\
&=\frac{\|W\|_F^2}{d}\le \frac{k\|W\|_{op}^2}{d}.
\end{split}
\end{eqnarray*}

The result then follows directly from Theorem \ref{thm:dico}.
\end{proof}
\section{Law of robustness for finite-width / empirical RF}
    Consider the finite-width RF regime ~\cite{rf,rf2,bach17,inductivebietti,telgarskytransport}, where the rows of the hidden weights matrix $W=(w_1,\ldots,w_k) \in \mathbb R^{k \times d}$
    are chosen iid according to $\tau_d$ (the uniform distribution on the unit-sphere $\sphere$) or equivalently, from $\mathcal N(0,(1/d)I_d)$, and only the output weights vector $v \in \mathbb R^k$ is optimized. With this choice of $W$, let $\Phi_\rf:\sphere \to \mathbb R^k$ be the feature map resulting from \eqref{eq:frozenPhiW} with corresponding kernel $K_\rf(x,x') = \Phi_\rf(x)^\top \Phi_\rf(x')$, which can be seen  as an empirical version of the kernel given in \eqref{eq:rfntkkernelfunc}. However, $K_\rf$ is not a dot-product kernel; this leads to technical difficulties.

\begin{restatable}[Curvature constants of activation function]{df}{}
For $z\sim \mathcal N(0,1)$, define scalars
\begin{eqnarray}
\beta_0(\sigma) := \mathbb E_z[\sigma(z)]^2 \ge 0,\, \beta_1(\sigma) := \mathbb E_z[z\sigma(z)]^2 \ge 0,\, \beta_\star(\sigma) := \mathbb E_z[\sigma(z)^2]-\beta_0(\sigma) - \beta_1(\sigma) \in \mathbb R.
\label{eq:betas}
\end{eqnarray}
\label{df:beta}
\end{restatable}
\vspace{-.5cm}
These quantities, which measure the degree of nonlinearity and curvature of the activation function $\sigma$, appear naturally in our analysis of the eigenvalues of the random matrix $C_\sigma(W):=\cov_{x \sim \tau_d}(\sqrt{d}\sigma(Wx))$, an essential step in our analysis of the Sobolev-seminorm of functions in representer subspace $\lspan_{K_\rf}(X)$. They have also appeared in the analysis of the generalization error of neural networks in RF and NTK regimes ~\cite{Mei2019,montanari2020,federica2020} and also in the analysis of the \emph{multiple-descent phenomenon} ~\cite{adlam20,ascoli20}.

Observe that one may write $C_\sigma(W) = \widetilde{C}_\sigma(W)-\mu_\sigma(W)\mu_\sigma(W)^\top$, where $\mu_\sigma(W) := \mathbb E_{x \sim \tau_d}[\sigma(Wx)]$ and the entries of $\widehat{C}_{j,\ell}(W) \in \mathbb R^{k \times k}$ are given by
\begin{eqnarray}
\widetilde{C}_\sigma(W))_{j,\ell} := \mathbb E_{x \sim \tau_d}[\sqrt{d}\sigma(x^\top w_j)\sqrt{d}\sigma(x^\top w_\ell)].
\label{eq:Ctilde}
\end{eqnarray}
We will need the following technical condition.
\begin{restatable}{cond}{relulike}
\label{cond:relulike}
(1) There exists a continuous function $\phi_\sigma:[-1,1] \to \mathbb R$ which is thrice continuously-differentiable at $0$ such that $\widetilde{C}_\sigma(W)_{j,\ell}=\phi_\sigma(w_j^\top w_\ell)$ for all $j,\ell \in [k]$ and \textbf{at least one} of the following two-conditions holds:
\begin{itemize}
\item [(2A)] $\beta_\star(\sigma) > 0$.
\item [(2B)] $\beta_\star(\sigma) \ge 0$, $\beta_1(\sigma)>0$ and $k/d \le \gamma_1$ for some absolute constant $\gamma_1 \in [0,1)$.
\end{itemize}
\end{restatable}
The above condition is satisfied when the underlying activation function $\sigma$ is the ReLU or absolute-value, or the gaussian error-function. Part (1) implies $C_\sigma(W)=\widetilde{C}_\sigma(W)-\phi_\sigma(0)1_k1_k^\top$ is a dot-product kernel matrix. Part (2A) was introduced in \cite{Mei2019,pennington17,montanari2020,Hastie2019} in the analysis of the generalization error for finite-width neural networks in various linearized regimes (RF, NTK, etc.). 
\begin{restatable}[Law of robustness for finite-width RF]{thm}{finiteRFLaw}
Assume $k \asymp d$ and Condition \ref{cond:relulike}. Then it holds w.p $1-(n \land d)^{-\Omega(1)}$ over $W$ and the generic dataset $\mathcal D_n$ that every $f \in \lspan_{K_{\rf}}(X)$ which $\varepsilon$-memorizes $\mathcal D_n$ verifies $\mathfrak{S}(f) \ge \Omar((\varepsilon^\star_{\mathrm{test}}-\varepsilon)\sqrt{n})$.
\label{thm:finiteRFLaw}
\end{restatable}
Note that the above lower-bound matches the infinite-width bound established in Theorem \ref{thm:infiniteRFNTKLaw}. Intuitively, this was to be expected as using fewer than infinitely many hidden neurons can only make the resulting model less smooth / robust (it is easier to smoothly draw when given more options).

\subsection{Matching upper-bound for min-norm interpolator}
We now establish a matching $\sqrt{n}$ an upper-bound which proves that the lower-bound in Theorem \ref{thm:finiteRFLaw} is tight: it is achieved by the min-norm interpolator. We consider the following so-called proportionately scaling regime, where $n$, $d$, and $k$ are allowed to simultaneously go to infinity at the same rate, i.e according to
\begin{restatable}[Proportionate scaling]{cond}{}
$n,d,k\to \infty$ in such a way that
\begin{eqnarray}
n/d \to \gamma_1 \in (0,\infty),\, k/d \to \gamma_2 \in (0,\infty),\, n/k \to \gamma = \gamma_1/\gamma_2 \in (0,\infty).
\label{eq:proportionate}
\end{eqnarray}
\end{restatable}
For a ridge penalty parameter $\lambda \ge 0$, consider the ridged random features interpolator
\begin{eqnarray}
\widehat{f}_{\rf,\lambda}(x):=K_\rf(X,x)^\top (K_\rf(X,X)+ (k\lambda/d)I_k)^{-1} y.
\label{eq:finiterfridge}
\end{eqnarray}

In \cite{Mei2019}, a fine analysis was done and explicit analytic formulae for the test error, the training error, and the norm of the optimal output weights vector $\widehat{v}_{\rf,\lambda}$ were obtained. Most importantly, it was shown that the training error is close to zero for $\lambda$ close to zero; the norm of $\widehat{v}_{\rf,\lambda}$ increases interpolation threshold ($k=n$) where it diverges to infinity; then beyond this threshold, it converges to a constant as $\gamma \to \infty$. This behavior was proposed as an explanation of the origins of the \emph{double-descent} phenomenon. and then decreases.

\paragraph{The upper-bound.}
For stating and proving the upper-bound we promised, will need the following technical restriction from ~\cite{Mei2019}.
\begin{restatable}[\cite{Mei2019}]{cond}{}
The activation function $\sigma: \mathbb R \to \mathbb R$ is weakly differentiable and satisfies the growth condition $|\sigma(t)|,|\sigma'(t)|\le c_0 e^{c_1 |t|}$.
for some $c_0,c_1 \in [0,\infty)$. Recall the definition of the coefficients $\beta_0(\sigma) \ge 0$, $\beta_1(\sigma)\ge 0$, and $\beta_\star(\sigma) \in \mathbb R$ from \eqref{eq:betas}. Assume that $\beta_\star(\sigma) > 0$ and define the coefficient $\theta = \theta(\sigma) := \sqrt{\dfrac{\beta_1(\sigma)}{\beta_\star(\sigma)}}$.
\label{cond:purelynonlin}
\end{restatable}
For example, the ReLU and the tanh activation functions satisfy the above condition.
This condition was introduced in  \cite{pennington17,Mei2019,Hastie2019} to help compute the traces of random matrices involving the nonlinear gram matrix $\Phi_\rf(X)^\top \Phi_\rf(X) \in \mathbb R^{n \times n}$, which eventually yield analytic formula for train error (MSE), test error, and squared norm of output weights $\widehat{v}_{\rf,\lambda}$. This is akin to the use of so-called \emph{Gaussian Equivalence Conjecture} in the analysis of shallow neural networks, whereby random nonlinear features can be replaced by noisy linear ones (noise), to obtain an equivalent model which has the same training error, test error, etc. asymptotics.

\begin{restatable}[Upper-bound for nonrobustness in finite-width RF regime]{thm}{law}
\label{thm:law}
For a large class of activation functions including the ReLU, tanh, gaussian error-function (erf), and the absolute-value, we have the following. In the limit when $n,d,k \to \infty$ in the sense of \eqref{eq:proportionate} and fixed ridge parameter $\lambda \ge 0$, it holds w.p tending to $1$ that $\mathfrak{S}(\widehat{f}_{\rf,\lambda}) \asymp \varepsilon_{\mathrm{test}}^\star \sqrt{n}$ if $(\gamma,\lambda) \ne (1,0)$ and $\mathfrak{S}(\widehat{f}_{\rf,\lambda})/(\varepsilon_{\mathrm{test}}^\star \sqrt{n}) \to \infty$ otherwise.
\end{restatable}
 
\section{Law of robustness for finite-width / empirical NTK}
    Now consider the finite-width NTK regime where the rows of the hidden weights matrix $W \in \mathbb R^{k \times d}$ are drawn iid from $\tau_d$, producing a feature map $\Phi_\ntk:\sphere \to \mathbb R^{kd}$ given by
    \begin{eqnarray}
    \Phi_{\ntk}(x) := \frac{1}{\sqrt{k}}\sigma'(Wx) \otimes x = \frac{1}{\sqrt{k}}(\sigma'(x^\top w_1) x,\ldots,\sigma'(x^\top w_k)x)),
    \label{eq:Phintk}
    \end{eqnarray}
    with associated kernel $K_\ntk(x,x') := \Phi_\ntk(x)^\top \Phi_\ntk(x')$, an empirical version of the infinite-width NTK kernel $K_\ntk^\infty$ given in \eqref{eq:rfntkkernelfunc}. The generalization properties for regression with this kernel have studied extensively in the literature (see \cite{Mei2019,adlam20} for example).
    Our contribution here focuses on robustness, and its interplay with memorization.
We will need the function condition on the activation function $\sigma$.
\begin{restatable}{cond}{}
Condition \ref{cond:relulike} holds with the activation function $\sigma$ replaced by its first derivative $\sigma'$. 
\label{cond:signlike}
 \end{restatable}
This condition ensures that the smallest eigenvalue of the covariance matrix of $C_{\Phi_\ntk} \in \mathbb R^{kd \times kd}$ of $\Phi_\ntk(x)$ for $x \sim \tau_d$, is lower-bounded i.e lower-bounded, by $\Omega(1/k)$ w.h.p. On the other hand, a simple calculation reveals that $\|\Phi_\ntk\|_{L^2(\tau_d)} = \mathcal O(1)$. Thus, the ratio $\alpha_{\Phi_\ntk}$ defined in \eqref{eq:alpharatio} is $\mathcal O(k)$ w.h.p. This gives the following corollary to Theorem \ref{thm:dico} (proved in the appendix).
\begin{restatable}[Law of robustness for finite-width NTK]{thm}{finiteNTKLaw}
Assume Condition \ref{cond:signlike} holds. For sufficiently large $n$, $d$, and $k$ such that $d \asymp k  \lesssim O(n)$, the following holds w.p $1-d^{-\Omega(1)}$ over $W$ and the generic dataset $\mathcal D_n$:  every $f \in \lspan_{\Phi_\ntk}(X)$ which $\varepsilon$-memorizes $\mathcal D_n$ satisfies
$\mathfrak{S}(f) \ge \Omar((\varepsilon^\star_{\mathrm{test}}-\varepsilon)\sqrt{\dfrac{n}{k}})$.
\label{thm:finiteNTKLaw}
\end{restatable}

\begin{restatable}[Implications for Conjecture \ref{conj:bubeck}]{rmk}{}
Since it is believed that neural networks trained via gradient-descent (GD) behave like NTK approximations ~\cite{jacot18}, the above theorem suggests that Conjecture \ref{conj:bubeck} might be true for models trained via GD.
\end{restatable}

\subsection{Consequences for min-norm and ridged finite-width NTK models}
In \cite{montanari2020}, the following scaling for finite-width NTK was considered 
\begin{eqnarray}
n,d,k \to \infty \text{ such that }
kd \gtrsim n(\log d)^C,\text{ and }
n \gtrsim d \gtrsim k \gtrsim d^\delta,
\label{eq:ntkscaling}
\end{eqnarray}
for constants $C,\delta > 0$.
For any $\lambda \ge 0$, consider the ridged interpolator $\widehat{f}_{\ntk,\lambda}$ defined by
\begin{eqnarray}
\widehat{f}_{\ntk,\lambda}(x):= K_{\ntk}(X,x)^\top (K_\ntk(X,X)+\lambda I_n)^{-1} y.
\label{eq:finitentkridge}
\end{eqnarray}
Note that $\widehat{f}_{\ntk} = \widehat{f}_{\ntk,0}$ is the min-norm interpolator, on the event that the kernel gram matrix $K_\ntk(X,X)$ is invertible. Under the scaling limit \eqref{eq:ntkscaling}, it was established in ~\cite{montanari2020} that 
\begin{eqnarray}
\lambda_{\min}(K_\ntk(X,X)) \ge \mathrm{var}_{z \sim \mathcal N(0,1)}(\sigma'(z))-o_{\mathbb P}(1),
\label{eq:lambdaminZZT}
\end{eqnarray}
which immediately implies $\rank(K_\ntk(X,X)) = n$ w.p tending to $1$, and so the min-norm / least squares interpolator  corresponding to $\lambda=0$ in \eqref{eq:finitentkridge}, perfectly memorizes generic the generic dataset $\mathcal D_n$. In the following result, we establish lower-bound on the robustness this interpolator.



\begin{restatable}[Nonrobust memorization in finite-width NTK]{thm}{finiteNTKminormLaw}
For a large class of activation functions including the ReLU, tanh, and the absolute-value, in the scaling \eqref{eq:ntkscaling}, it holds w.p tending to $1$ that
$
\mathfrak{S}(\widehat{f}_{\ntk}) \ge \Omega(\varepsilon_{\mathrm{test}}^\star \sqrt{\dfrac{n}{k}})$.
\label{thm:finiteNTKminormLaw}
\end{restatable}
\section{Experiments}
\label{sec:exp}

\subsection{Experimental setup}
\label{subsec:expsetup}
\paragraph{Experiment 1: Finite-width NTK (only first-layer kernel).}
For this experiment, we fix the input dimension $d = 50$ and the width of the neural network
to $k=40$.
The number of samples $n$ sweeps the range of integers from $20$ through $3020$, in increments of $100$.
We consider a variety of activation functions: ReLU, tanh, absolute-value, and the gaussian error-function (erf). For each value of $n$, we sample $10$ generic datasets with $n$ samples on the unit-sphere in $\sphere$, more precisely, random data points drawn iid from $\tau_d$
and given labels according to \eqref{eq:noisylinear}, with the $w_0 \in \sphere$ and noise level $\zeta$ sweeping from $0$ through $1$ in steps of $0.2$. For each such dataset $\mathcal D_n$, we also sample $15$ iid realizations of $k$ rows of hidden weights matrix $W$, iid from  $\tau_d$. Finally, for each $\lambda \in \{0, 10^{-5}, 10^{-4}, 10^{-3}\}$, we do ridge-regression to get an instance $\widehat{f}_{\ntk,\lambda}$ of the model \eqref{eq:finitentkridge}. 



\paragraph{Experiment 2: Finite-width RF (NTK with only second-layer kernel).}
The experimental setting is as in Experiment 1, except that now: $d=300$, $k$ sweeps from $100$ through $1000$ in steps of $50$, while $n$ sweeps the random of integers from $200$ through $1000$ in steps of $100$. For each such dataset and each value of $\lambda$ as in Experiment 1, we do ridge-regression to get $\widehat{f}_{\rf}$ as in
\eqref{eq:finiterfridge}. 


\paragraph{Experiment 3: RF and NTK with infinite-width.}
We run a similar experiment as in Experiment 1 and 2, but with $d=500$ and $n$ sweeps from $100$ through $1000$ in steps of $100$,  $k = \infty$, and $\lambda=0$. For each dataset, we compute the min-norm interpolator in RF and NTK regimes $\widehat{f}_{\rf/\ntk}^\infty$ via \eqref{eq:infiniteRFNTKminnorm}.

\paragraph{Metrics.} For each fitted model $\widehat{f}$ in each experiment, we estimate its Sobolev-seminorm (our measure of robustness) $\mathfrak{S}(\widehat{f})$ by drawing $500$ random points iid from $\tau_d$ (the uniform distribution on the unit-sphere $\sphere$), and computing the square-root of the average value of $\|\nabla_\sphere \widehat{f}(x)\|^2$ over these $500$ points. We also compute the squared test error on this test dataset.

\begin{figure}[!htb]
    \centering
    \includegraphics[width=1\linewidth]{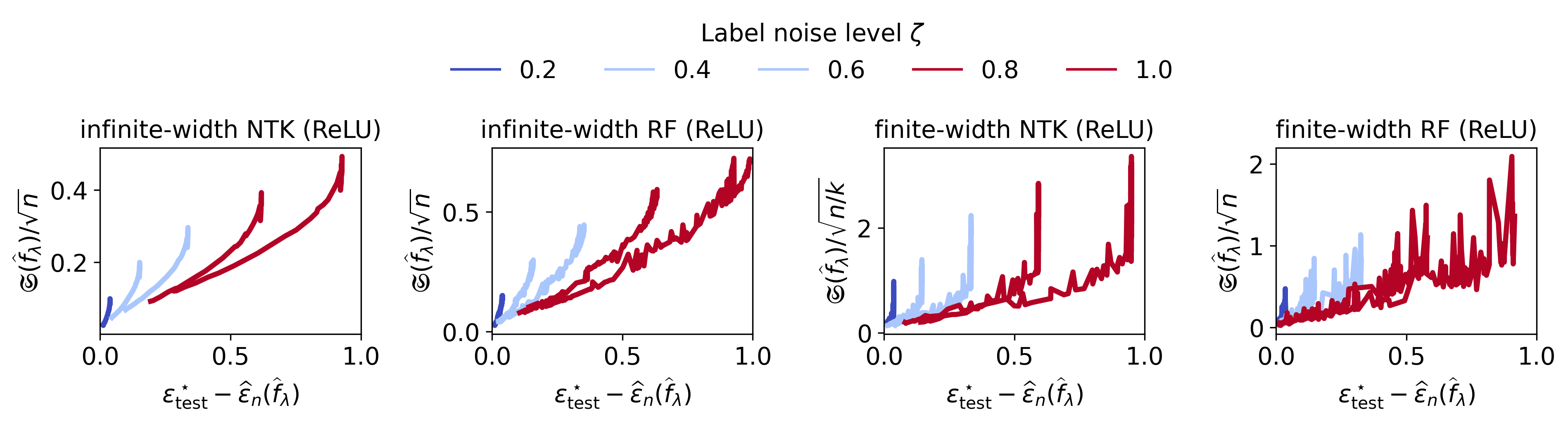}
    \caption{(Experiments 1 -- 3) Empirical verification of our proposed laws of robustness at different noise levels in the problem.
    Results shown are for the ReLU activation function (results for other activation functions are in the appendix).
    Each color corresponds to strength $\zeta$ of the label noise. Notice the (super)linear trend between the x-axis and the y-axis of the figures, in conformity with the predictions of our theorems.}
    \label{fig:alllaws}
\end{figure}

\subsection{Results of the experiments}
\paragraph{Confirmation of the robustness laws.}
As predicted by Theorems \ref{thm:infiniteRFNTKLaw}, \ref{thm:finiteNTKLaw}, and \ref{thm:finiteRFLaw}, in Figure \ref{fig:alllaws} we observe a clear linear relationship between the Sobolev-seminorm  $\mathfrak{S}(\widehat{f})$ of the models $\widehat{f}$ (see Experiments 1, 2, and 3 of section \ref{subsec:expsetup} for details) and $(\varepsilon^\star-\widehat{\varepsilon}_n(\widehat{f}))\sqrt{n}$ (for infinite-width RF / NTK and finite-width RF) and $(\varepsilon^\star-\widehat{\varepsilon}_n(\widehat{f}))\sqrt{n/k}$, for finite-width NTK; a quantitative tradeoff between memorization and robustness.
\begin{figure}[!htb]
    \centering
    \begin{subfigure}{\textwidth}
    \centering
    \includegraphics[width=1\linewidth]{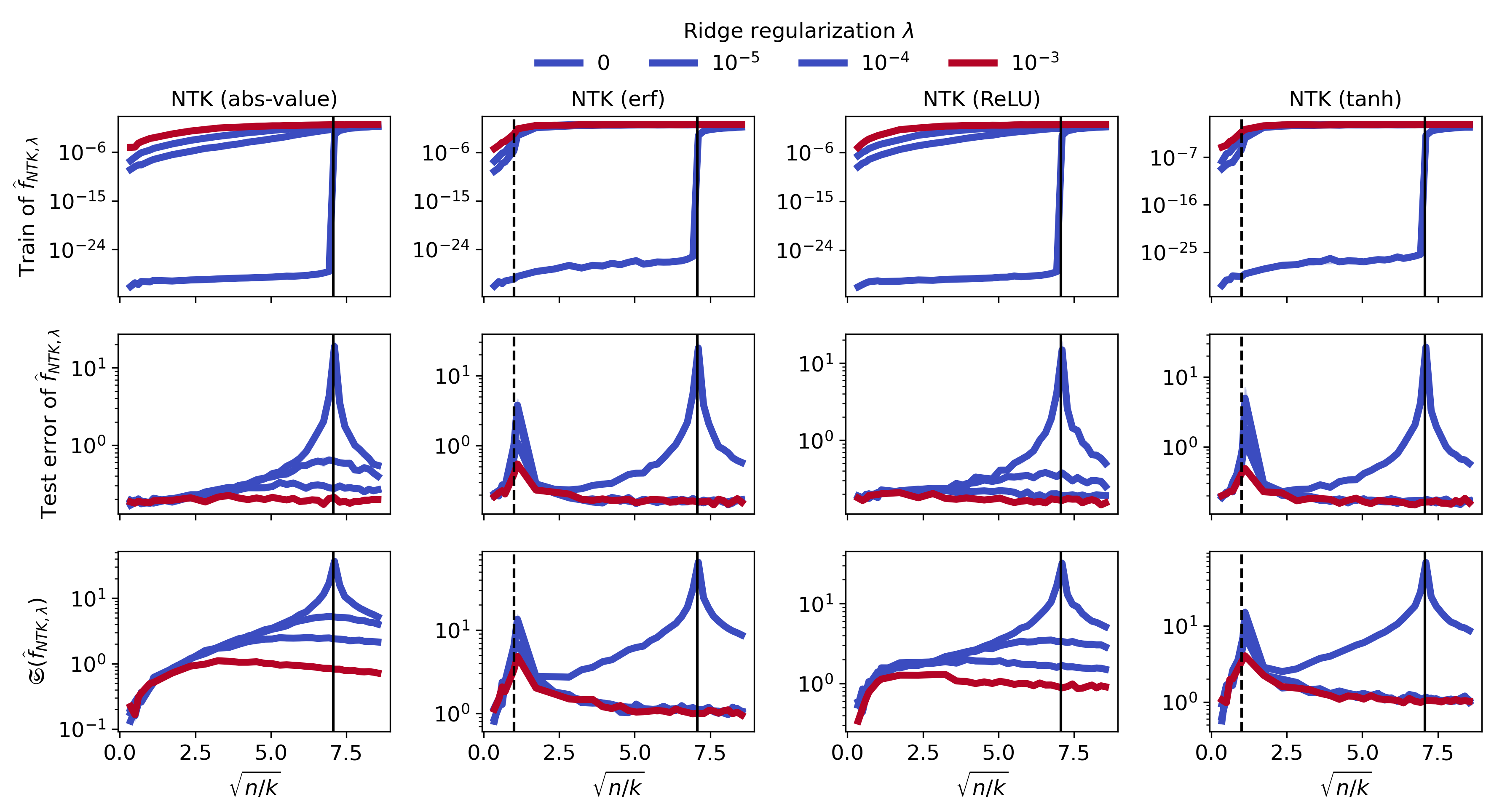}
    \caption{(Experiment 1) Finite-width NTK with $d=50$, $k=40$, and $n \in \{100,200,300,\ldots,3000\}$. The vertical lines correspond to interpolation thresholds at $n=k$ and $n=kd$ \cite{adlam20,ascoli20}.}
    \end{subfigure}
    \hfill
    \begin{subfigure}{\textwidth}
    \centering
    \includegraphics[width=1\linewidth]{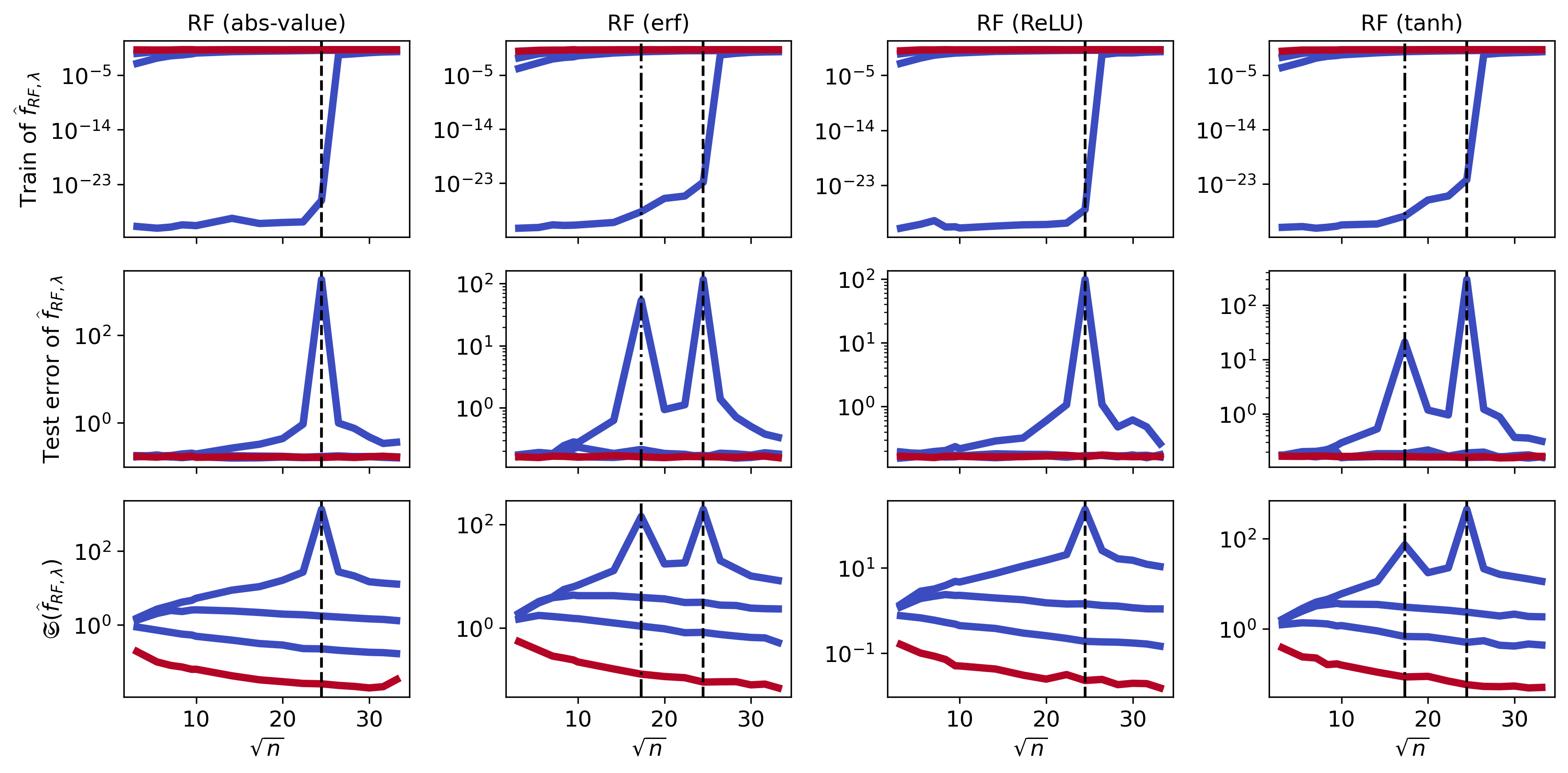}
    \caption{(Experiment 2) RF regime with $d=300$ and $k=600$ and $n \in \{100,200,300,1200\}$. The vertical lines correspond to interpolation thresholds at $n=d$ and $n=k$ \cite{adlam20,ascoli20}.}
    \end{subfigure}
    \caption{Multiple-descent in robustness. The first row of each plot corresponds to training error, second row corresponds to test / generalization error, while the third row corresponds to Sobolev-seminorm of the model (our measure of nonrobustness). Columns are different choices of activation function $\sigma$.
    The data is generated according to \eqref{eq:noisylinear} with label noise level is fixed at $\zeta = 0.2$. The colors correspond to different values of the ridge parameter $\lambda$. Observe how the test error and the Sobolev-seminorm of each model follow the same multiple-descent pattern.}
    \label{fig:multiple}
\end{figure}
\begin{figure}[!htb]
    \centering
    \begin{subfigure}{\textwidth}
    \centering
    \includegraphics[width=1\linewidth]{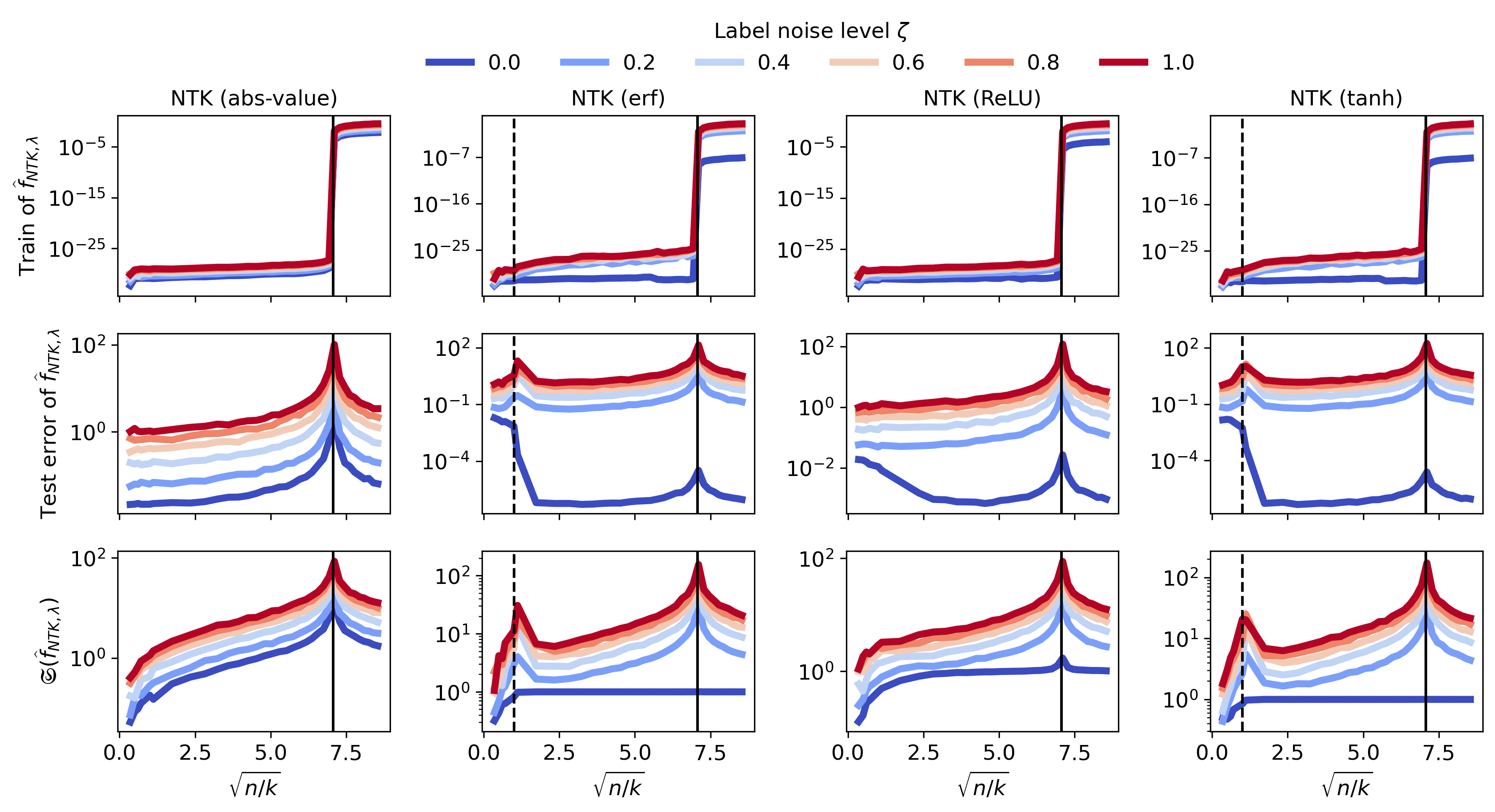}
    \caption{(Experiment 1) Finite-width NTK with $d=50$, $k=40$, and $n \in \{100,200,300,\ldots,3000\}$. The vertical lines correspond to interpolation thresholds at $n=k$ and $n=kd$, as predicted in \cite{adlam20,ascoli20}.}
    \end{subfigure}
    \hfill
    \begin{subfigure}{\textwidth}
    \centering
    \includegraphics[width=1\linewidth]{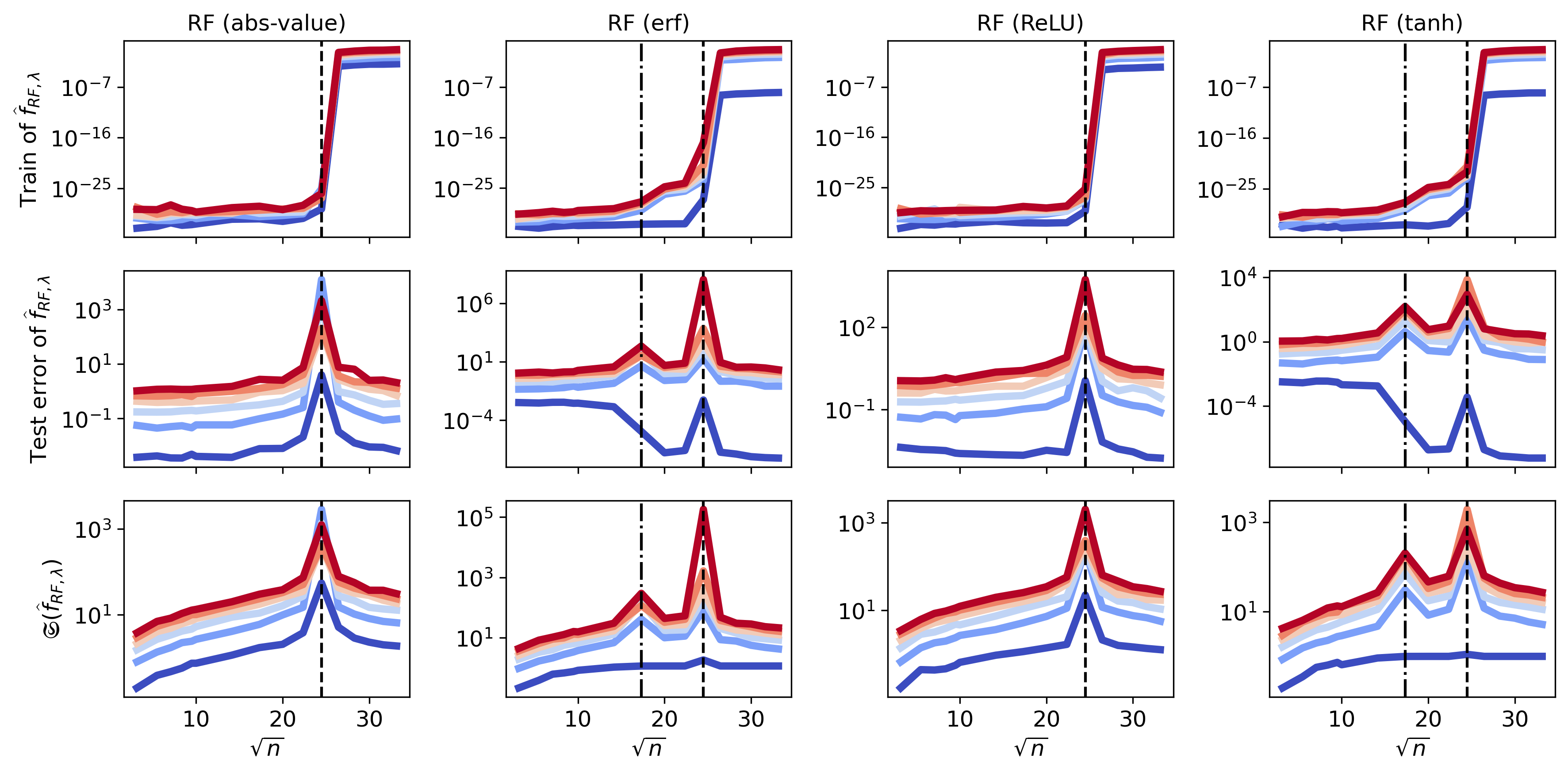}
    \caption{(Experiment 2) RF regime with $d=300$ and $k=600$ and $n \in \{100,200,300,1200\}$. The vertical lines correspond to interpolation thresholds at $n=k$ and $n=d$, predicted in \cite{adlam20,ascoli20}.}
    \end{subfigure}
    \caption{Multiple-descent in robustness in min-norm interpolator. All plots correspond to no regularization ($\lambda=0$). The first row of each plot corresponds to training error, second row corresponds to test / generalization error, while the third row corresponds to Sobolev-seminorm of the model (our measure of nonrobustness). Columns are different choices of activation function $\sigma$.
    The data is generated according to \eqref{eq:noisylinear} with different values the level $\zeta$ of label-noise.}
    \label{fig:multiple2}
\end{figure}

\paragraph{Multiple-descent (MD) behavior in robustness for finite-width regimes.}
In Figure \ref{fig:multiple}(a), we plot the Sobolev-seminorm $\mathfrak{S}(\widehat{f}_{\ntk})$ of the min-norm interpolator $\widehat{f}_{\rf}$ versus $\sqrt{n/k}$ (Experiment 1). We observe a multiple-descent phenomenon (MD) whereby $\mathfrak{S}(\widehat{f}_{\ntk})/\sqrt{n/k}$ becomes unbounded for all the activation functions, at the point $kd=n$ in the phase diagram. Interestingly, this singularity point (i.e for which $kd=n$) in phase-space corresponds to the so-called nonlinear interpolation threshold which has been recently identified in \cite{ascoli20,adlam20}.
For the tanh and erf activation functions, we see a second singularity at the point $n=k$. This corresponds to the so-called \emph{linear} interpolation threshold \cite{ascoli20,adlam20}.
Notice how the test error and the Sobolev-semnorm $\mathfrak{S}(\widehat{f})$ of the model follow similar multiple-descent patterns. Also notice the attenuation effect of regularization
between the interpolation thresholds, we observe a linear trend between $\mathfrak{S}(\widehat{f}_{\ntk})$ and $\sqrt{n/k}$ as predicted by Theorem \ref{thm:finiteNTKLaw}.

In the case of RF, in Figure \ref{fig:multiple}(b) we plot the Sobolev-seminorm interpolator $\widehat{f}_{\rf}$, for different activation functions and label noise levels.
We observe a singularity point inthe the phase space along $n=d$ for all the activation functions, and another one at $n=k$ for the tanh and erf activation functions.
Outside the interpolation thresholds, we confirm the linear law predicted by Theorem \ref{thm:finiteRFLaw} and Theorem \ref{thm:law}. 

Importantly, we observe in Figure \ref{fig:multiple} that in both the finite-width RF and finite-width NTK experiments, the generalization error and nonrobustness curves have the same multiple-descent pattern. This is the first time MD is exhibited in a statistical functional (here, robustness) other generalization error. Finally, we observe that the Sobolev-seminorm $\mathfrak{S}(\widehat{f})$ is reduced with increasing ridge regularization level $\lambda$. This also kills the multiple-descent.

\paragraph{The effect of the noise level.} In Figure \ref{fig:multiple2}, we plot multiple-descent curves (again for robustness and test / generalization error) as a function of the amount of label noise in the data distribution (see \ref{eq:noisylinear}). As would be expected, we see that MD is amplified with increasing label noise level ($\zeta$).

\subsection{Partial explanation of multiple-descent in robustness (in case of finite-width RF)}
Refer to Figure \ref{fig:multiple}.
As with multiple-descent (MD) in generalization error ~\cite{doubledescent,Loog20,adlam20,ascoli20}, MD in robustness we observe here is probably due to bad conditioning of the kernel gram matrix $K_M(X,X)$ (for $M \in \{\rf,\ntk\}$) close to the interpolation thresholds. We observe that the Sobolev-seminorm $\mathfrak{S}(\widehat{f})$ is reduced with increasing ridge regularization level $\lambda$. This also kills the multiple-descent.

\paragraph{Explaining MD for finite-width RF.}
In the case of RF, Theorem \ref{thm:law} rigorously predicts
the singularity observed in Figure \ref{fig:multiple} at $n=k$ (i.e $\gamma = 1$) for the min-norm interpolator (corresponding to $\lambda=0$).
Indeed, the prove of Theorem \ref{thm:law} reveals that if $\widehat{v}_{\rf,\lambda} \in \mathbb R^k$ is the output weights vector of the RF interpolator with ridge penalty $\lambda=\mathcal O(1)$, then the following holds w.p $1-d^{-\Omega(1)}$ over the random hidden weights matrix $W \in \mathbb R^{k \times d}$
\begin{eqnarray}
\mathfrak{S}(\widehat{f}_{\rf,\lambda}) \asymp \|\widehat{v}_{\rf,\lambda}\| = \begin{cases}
\omega_{\mathbb P}(\sqrt{d})=\omega_{\mathbb P}(\sqrt{n}),&\mbox{ if }(\gamma,\lambda)=(1,0),\\
\Theta_{\mathbb P}(\sqrt{d})=\Omega_{\mathbb P}(\sqrt{n}),&\mbox{ else.}
\end{cases}
\end{eqnarray}
Thus, there is a singularity at $(\gamma,\lambda)=(1,0)$, i.e for the ridge-less interpolator at $n = k$, where in $\mathfrak{S}(\widehat{f}_{\rf,\lambda})/\sqrt{n} = \omega_{\mathbb P}(1) \to \infty$ in the limit $n,d,k \to \infty$ according to \eqref{eq:proportionate}. Moreover, the above formula reveals that any multiple-descent behavior in $\|\widehat{v}_{\rf,\lambda}\|$, produces the exactly the same multiple-descent behavior in $\mathfrak{S}(\widehat{f}_{\rf,\lambda})$, asymptotically in the sample size $n$.

\paragraph{Still missing the full picture.}
Providing rigorous explanations for the other singularities in Figures \ref{fig:multiple} and \ref{fig:multiple2}, namely at $n=d$ for RF, and at $n=k$ and $n=kd$ for NTK (finite-width and infinite-width) is left for future work.

\section{Conclusion}
In this work, we have derived precise laws for robustness of neural networks in both the (in)finite-width random features (RF) and (in)finite-width neural tangent kernel (NTK) regimes. Our results show a clear tradeoff between memorization and  robustness, as measured by the Sobolev-seminorm of the model, a new measure of (non)robustness we propose, for the min-norm interpolators, ridged interpolators, or generalizations thereof (in fact, any model in the so-called "representer subpace" of the data \eqref{eq:kernspan}). Empirical results confirm our theoretical findings. We also accidentally observe a new phenomenon in the finite-width regimes: multiple-descent in robustness, for which we provide a theoretical explanation in the case of finite-width RF.




\paragraph{Limitations and future directions.}
\begin{itemize}
\item[(1)] \emph{Going beyong log-concavity.}
For technical reasons, our work only considers log-concave isotropic data. A step towards removing this assumption would be to consider the student-teacher paradigm with block structure like in ~\cite{moredascoli}.
\item[(2)]\emph{Analysis of fully-trained neural networks.}
We have provided a complete picture of the fundamental tradeoffs between robustness of linearized neural networks (RF and NTK). An analysis of fully-trained neural networks would be a big next step. Exploring Theorem \ref{thm:frozenlaw} could be a starting point for this.
\end{itemize}

\paragraph{Acknowledgement.} Thanks to Alberto Bietti (NYU) for stimulating discussions around random features and NTK. Thanks to Mike Gartrell and Lorenzo Croissant (Criteo), and also to Alberto Bietti for proof-reading various versions of this manuscript and making useful suggestions. Finally, thanks to Criteo (my previous employer) where this work was carried out.





\bibliographystyle{plainnat}
\bibliography{literature.bib}

\begin{thebibliography}{41}
\providecommand{\natexlab}[1]{#1}
\providecommand{\url}[1]{\texttt{#1}}
\expandafter\ifx\csname urlstyle\endcsname\relax
  \providecommand{\doi}[1]{doi: #1}\else
  \providecommand{\doi}{doi: \begingroup \urlstyle{rm}\Url}\fi

\bibitem[Adlam and Pennington(2020)]{adlam20}
Ben Adlam and Jeffrey Pennington.
\newblock The neural tangent kernel in high dimensions: Triple descent and a
  multi-scale theory of generalization.
\newblock In Hal~Daumé III and Aarti Singh, editors, \emph{Proceedings of the
  37th International Conference on Machine Learning}, volume 119 of
  \emph{Proceedings of Machine Learning Research}, pages 74--84. PMLR, 13--18
  Jul 2020.

\bibitem[Arora et~al.(2019)Arora, Du, Hu, Li, Salakhutdinov, and
  Wang]{aroraexact}
Sanjeev Arora, Simon~S. Du, Wei Hu, Zhiyuan Li, Ruslan Salakhutdinov, and
  Ruosong Wang.
\newblock On exact computation with an infinitely wide neural net.
\newblock In Hanna~M. Wallach, Hugo Larochelle, Alina Beygelzimer, Florence
  d'Alch{\'{e}}{-}Buc, Emily~B. Fox, and Roman Garnett, editors, \emph{Advances
  in Neural Information Neural Information Processing Systems}, pages
  8139--8148, 2019.

\bibitem[Azé and Corvellec(2017)]{nonlinearhoffman}
Dominique Azé and Jean-Noël Corvellec.
\newblock Nonlinear error bounds via a change of function.
\newblock \emph{Journal of Optimization Theory and Applications}, 172, 2017.

\bibitem[Bach(2017)]{bach17}
Francis~R. Bach.
\newblock On the equivalence between kernel quadrature rules and random feature
  expansions.
\newblock \emph{J. Mach. Learn. Res.}, 18:\penalty0 21:1--21:38, 2017.

\bibitem[{Bartlett}(1998)]{bartlett98}
P.~L. {Bartlett}.
\newblock The sample complexity of pattern classification with neural networks:
  the size of the weights is more important than the size of the network.
\newblock \emph{IEEE Transactions on Information Theory}, 44\penalty0
  (2):\penalty0 525--536, 1998.

\bibitem[Baum(1988)]{baum88}
Eric~B Baum.
\newblock On the capabilities of multilayer perceptrons.
\newblock \emph{Journal of Complexity}, 4\penalty0 (3):\penalty0 193--215,
  1988.

\bibitem[Belkin et~al.(2018)Belkin, Ma, and Mandal]{belkin18}
Mikhail Belkin, Siyuan Ma, and Soumik Mandal.
\newblock To understand deep learning we need to understand kernel learning.
\newblock In Jennifer Dy and Andreas Krause, editors, \emph{Proceedings of the
  35th International Conference on Machine Learning}, volume~80 of
  \emph{Proceedings of Machine Learning Research}, pages 541--549. PMLR, 10--15
  Jul 2018.

\bibitem[Belkin et~al.(2019)Belkin, Hsu, Ma, and Mandal]{doubledescent}
Mikhail Belkin, Daniel Hsu, Siyuan Ma, and Soumik Mandal.
\newblock Reconciling modern machine-learning practice and the classical
  bias{\textendash}variance trade-off.
\newblock \emph{Proceedings of the National Academy of Sciences}, 116\penalty0
  (32):\penalty0 15849--15854, 2019.
\newblock ISSN 0027-8424.

\bibitem[Bietti and Mairal(2019{\natexlab{a}})]{biettiinvariance}
Alberto Bietti and Julien Mairal.
\newblock Group invariance, stability to deformations, and complexity of deep
  convolutional representations.
\newblock \emph{Journal of Machine Learning Research}, 20\penalty0
  (25):\penalty0 1--49, 2019{\natexlab{a}}.

\bibitem[Bietti and Mairal(2019{\natexlab{b}})]{inductivebietti}
Alberto Bietti and Julien Mairal.
\newblock On the inductive bias of neural tangent kernels.
\newblock In \emph{Advances in Neural Information Processing Systems}, pages
  12873--12884, 2019{\natexlab{b}}.

\bibitem[Boucheron et~al.(2013)Boucheron, Lugosi, and Massart]{boucheron2013}
S.~Boucheron, G.~Lugosi, and P.~Massart.
\newblock \emph{Concentration Inequalities: A Nonasymptotic Theory of
  Independence}.
\newblock OUP Oxford, 2013.
\newblock ISBN 9780199535255.

\bibitem[{Boucheron, St\'ephane} et~al.(2005){Boucheron, St\'ephane},
  {Bousquet, Olivier}, and {Lugosi, G\'abor}]{boucheron05}
{Boucheron, St\'ephane}, {Bousquet, Olivier}, and {Lugosi, G\'abor}.
\newblock Theory of classification: a survey of some recent advances.
\newblock \emph{ESAIM: PS}, 9:\penalty0 323--375, 2005.
\newblock \doi{10.1051/ps:2005018}.

\bibitem[Bubeck et~al.(2018)Bubeck, Price, and Razenshteyn]{bubeck2018}
S{\'{e}}bastien Bubeck, Eric Price, and Ilya~P. Razenshteyn.
\newblock Adversarial examples from computational constraints.
\newblock \emph{CoRR}, abs/1805.10204, 2018.

\bibitem[Bubeck et~al.(2020a)Bubeck, Eldan, Lee, and Mikulincer]{bubecknetsize}
S{\'{e}}bastien Bubeck, Ronen Eldan, Yin~Tat Lee, and Dan Mikulincer.
\newblock Network size and size of the weights in memorization with two-layers
  neural networks.
\newblock In \emph{Advances in Neural Information Processing Systems}, 2020a.

\bibitem[{Bubeck} et~al.(2020b){Bubeck}, {Li}, and {Nagaraj}]{lor}
S{\'e}bastien {Bubeck}, Yuanzhi {Li}, and Dheeraj {Nagaraj}.
\newblock {A law of robustness for two-layers neural networks}.
\newblock \emph{arXiv e-prints}, art. arXiv:2009.14444, September 2020b.

\bibitem[Buchweitz(2016)]{Erez2016}
Erez Buchweitz.
\newblock Concentration of functions beyond lévy's inequality, 2016.

\bibitem[Corvellec and Motreanu(2007)]{hoffmanlsc}
Jean-No{\"e}l Corvellec and Viorica~V. Motreanu.
\newblock Nonlinear error bounds for lower semicontinuous functions on metric
  spaces.
\newblock \emph{Mathematical Programming}, 114\penalty0 (2):\penalty0 291,
  2007.

\bibitem[d'Ascoli et~al.(2021)d'Ascoli, Gabri{\'{e}}, Sagun, and
  Biroli]{moredascoli}
St{\'{e}}phane d'Ascoli, Marylou Gabri{\'{e}}, Levent Sagun, and Giulio Biroli.
\newblock More data or more parameters? investigating the effect of data
  structure on generalization.
\newblock abs/2103.05524, 2021.

\bibitem[{De Giorgi} et~al.(1980){De Giorgi}, Marino, and Tosques]{degiorgi80}
Ennio {De Giorgi}, Antonio Marino, and Mario Tosques.
\newblock Problemi di evoluzione in spazi metrici e curve di massima pendenza.
\newblock \emph{Atti della Accademia Nazionale dei Lincei. Classe di Scienze
  Fisiche, Matematiche e Naturali. Rendiconti}, 68\penalty0 (3):\penalty0
  180--187, 1980.

\bibitem[d\textquotesingle Ascoli et~al.(2020)d\textquotesingle Ascoli, Sagun,
  and Biroli]{ascoli20}
St\'{e}phane d\textquotesingle Ascoli, Levent Sagun, and Giulio Biroli.
\newblock Triple descent and the two kinds of overfitting: where \&amp; why do
  they appear?
\newblock In H.~Larochelle, M.~Ranzato, R.~Hadsell, M.~F. Balcan, and H.~Lin,
  editors, \emph{Advances in Neural Information Processing Systems}, volume~33,
  pages 3058--3069. Curran Associates, Inc., 2020.

\bibitem[El~Karoui(2010)]{elkaroui2010}
Noureddine El~Karoui.
\newblock The spectrum of kernel random matrices.
\newblock \emph{Ann. Statist.}, 2010.

\bibitem[Gerace et~al.(2020)Gerace, Loureiro, Krzakala, Mezard, and
  Zdeborova]{federica2020}
Federica Gerace, Bruno Loureiro, Florent Krzakala, Marc Mezard, and Lenka
  Zdeborova.
\newblock Generalisation error in learning with random features and the hidden
  manifold model.
\newblock In \emph{Proceedings of the 37th International Conference on Machine
  Learning}, volume 119 of \emph{Proceedings of Machine Learning Research},
  pages 3452--3462. PMLR, 13--18 Jul 2020.

\bibitem[Gozlan et~al.(2015)Gozlan, Roberto, and Samson]{Gozlan2015}
Nathael Gozlan, Cyril Roberto, and Paul-Marie Samson.
\newblock From dimension free concentration to the poincar{\'e} inequality.
\newblock \emph{Calculus of Variations and Partial Differential Equations},
  52\penalty0 (3):\penalty0 899--925, Mar 2015.

\bibitem[Hastie et~al.(2019)Hastie, Montanari, Rosset, and
  Tibshirani]{Hastie2019}
Trevor Hastie, Andrea Montanari, Saharon Rosset, and Ryan~J. Tibshirani.
\newblock Surprises in high-dimensional ridgeless least squares interpolation.
\newblock \emph{arXiv}, page 1903.08560v4, 2019.

\bibitem[{Husain} and {Balle}(2021)]{husain21}
Hisham {Husain} and Borja {Balle}.
\newblock {A Law of Robustness for Weight-bounded Neural Networks}.
\newblock \emph{arXiv e-prints}, art. arXiv:2102.08093, 2021.

\bibitem[Jacot et~al.(2018)Jacot, Gabriel, and Hongler]{jacot18}
Arthur Jacot, Franck Gabriel, and Clement Hongler.
\newblock Neural tangent kernel: Convergence and generalization in neural
  networks.
\newblock In \emph{Advances in Neural Information Processing Systems 31}. 2018.

\bibitem[Ji et~al.(2020)Ji, Telgarsky, and Xian]{telgarskytransport}
Ziwei Ji, Matus Telgarsky, and Ruicheng Xian.
\newblock Neural tangent kernels, transportation mappings, and universal
  approximation.
\newblock In \emph{8th International Conference on Learning Representations,
  {ICLR} 2020, Addis Ababa, Ethiopia, April 26-30, 2020}. OpenReview.net, 2020.

\bibitem[Ledoux(1999)]{ledoux}
Michel Ledoux.
\newblock Concentration of measure and logarithmic sobolev inequalities.
\newblock \emph{S\'eminaire de probabilit\'es de Strasbourg}, 33:\penalty0
  120--216, 1999.

\bibitem[Liang and Rakhlin(2020)]{justinterpolate}
Tengyuan Liang and Alexander Rakhlin.
\newblock Just interpolate: Kernel "ridgeless" regression can generalize.
\newblock \emph{ANNALS OF STATISTICS}, pages 1329--1347, 2020.

\bibitem[Loog et~al.(2020)Loog, Viering, Mey, Krijthe, and Tax]{Loog20}
Marco Loog, Tom Viering, Alexander Mey, Jesse~H. Krijthe, and David M.~J. Tax.
\newblock A brief prehistory of double descent.
\newblock \emph{Proceedings of the National Academy of Sciences}, 117\penalty0
  (20):\penalty0 10625--10626, 2020.
\newblock ISSN 0027-8424.

\bibitem[Louart et~al.(2018)Louart, Liao, and Couillet]{louart2018}
Cosme Louart, Zhenyu Liao, and Romain Couillet.
\newblock {A random matrix approach to neural networks}.
\newblock \emph{The Annals of Applied Probability}, 28\penalty0 (2):\penalty0
  1190 -- 1248, 2018.

\bibitem[{Mei} and {Montanari}(2019)]{Mei2019}
Song {Mei} and Andrea {Montanari}.
\newblock {The generalization error of random features regression: Precise
  asymptotics and double descent curve}.
\newblock \emph{arXiv e-prints}, art. arXiv:1908.05355, August 2019.

\bibitem[Montanari and Zhong(2020)]{montanari2020}
Andrea Montanari and Yiqiao Zhong.
\newblock The interpolation phase transition in neural networks: Memorization
  and generalization under lazy training.
\newblock \emph{CoRR}, abs/2007.12826, 2020.

\bibitem[Pennington and Worah(2017)]{pennington17}
Jeffrey Pennington and Pratik Worah.
\newblock Nonlinear random matrix theory for deep learning.
\newblock In \emph{Advances in Neural Information Processing Systems}, pages
  2637--2646, 2017.

\bibitem[Rahimi and Recht(2008)]{rf}
Ali Rahimi and Benjamin Recht.
\newblock Uniform approximation of functions with random bases.
\newblock 2008.

\bibitem[Rahimi and Recht(2009)]{rf2}
Ali Rahimi and Benjamin Recht.
\newblock Weighted sums of random kitchen sinks: Replacing minimization with
  randomization in learning.
\newblock 2009.

\bibitem[Sch\"{o}lkopf et~al.(2001)Sch\"{o}lkopf, Herbrich, and
  Smola]{scholkopf2001generalized}
Bernhard Sch\"{o}lkopf, Ralf Herbrich, and Alex~J. Smola.
\newblock A generalized representer theorem.
\newblock In \emph{COLT '01/EuroCOLT '01: Proceedings of the 14th Annual
  Conference on Computational Learning Theory and and 5th European Conference
  on Computational Learning Theory}, pages 416--426, London, UK, 2001.
  Springer-Verlag.
\newblock ISBN 3-540-42343-5.

\bibitem[Vershynin(2012)]{rmt}
Roman Vershynin.
\newblock \emph{Introduction to the non-asymptotic analysis of random
  matrices}, page 210–268.
\newblock Cambridge University Press, 2012.
\newblock \doi{10.1017/CBO9780511794308.006}.

\bibitem[Vershynin(2020)]{vershynin2020}
Roman Vershynin.
\newblock Memory capacity of neural networks with threshold and rectified
  linear unit activations.
\newblock \emph{SIAM Journal on Mathematics of Data Science}, 2\penalty0
  (4):\penalty0 1004--1033, 2020.

\bibitem[Villani(2003)]{villaniTopics}
C{\'e}dric Villani.
\newblock \emph{Topics in Optimal Transportation}.
\newblock American Mathematical Society, 2003.

\bibitem[Villani(2008)]{Villani}
C\'{e}dric Villani.
\newblock \emph{{Optimal Transport: Old and New}}.
\newblock Grundlehren der mathematischen Wissenschaften. Springer, 2009
  edition, September 2008.
\newblock ISBN 3540710493.

\end{thebibliography}
\appendix

\section{More experimental results}
In this section, we present additional empirical results to complement the results presented in section \ref{sec:exp} of the manuscript.

\begin{figure}[!htb]
    \centering
    \includegraphics[width=.75\linewidth]{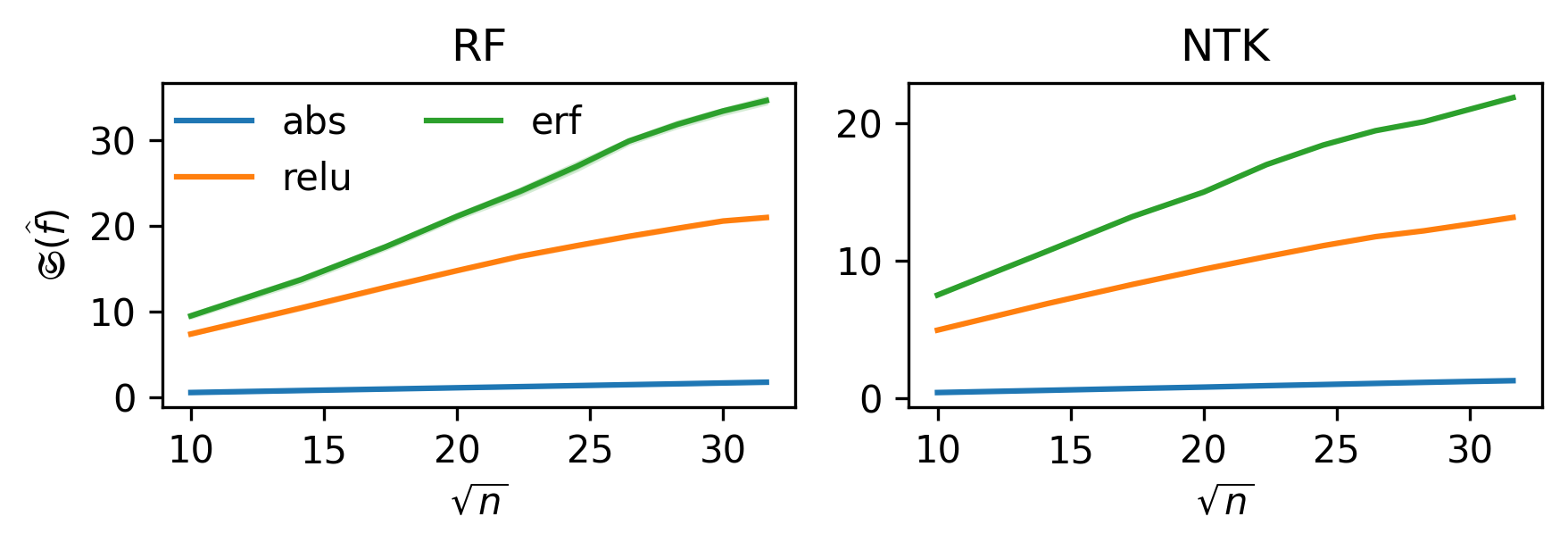}
\caption{(Experiment 3) Sobolev-seminorm $\mathfrak{S}(\widehat{f})$ of min-norm interpolator for infinite-width RF and NTK regime \eqref{eq:infiniteRFNTKminnorm}. Notice the linear relation between $\mathfrak{S}(\widehat{f})$ and $\sqrt{n}$ predicted by Theorems \ref{thm:infiniteRFNTKLaw} and  \ref{thm:infiniteRFNTKLowerbound}. For the absolute-value activation function, we see that the the curve is flat. This does not contradict Theorem \ref{thm:infiniteRFNTKLaw} because this activation function fails to satisfy the "$\phi'(0) \ne 0$ condition of the theorem. Indeed, thanks to \cite[Table 1]{louart2018}, we know that $\phi_\ntk(t) = (2/\pi)t^2 + \mathcal O(t^4)$ and $\phi_\rf(t) = 2/\pi+(1/\pi)t^2 + \mathcal O(t^4)$ for the absolute-value activation function.}
\label{fig:inf}
\end{figure}
\begin{figure}[!htb]
\begin{subfigure}{\textwidth}
    \centering
    \includegraphics[width=\linewidth]{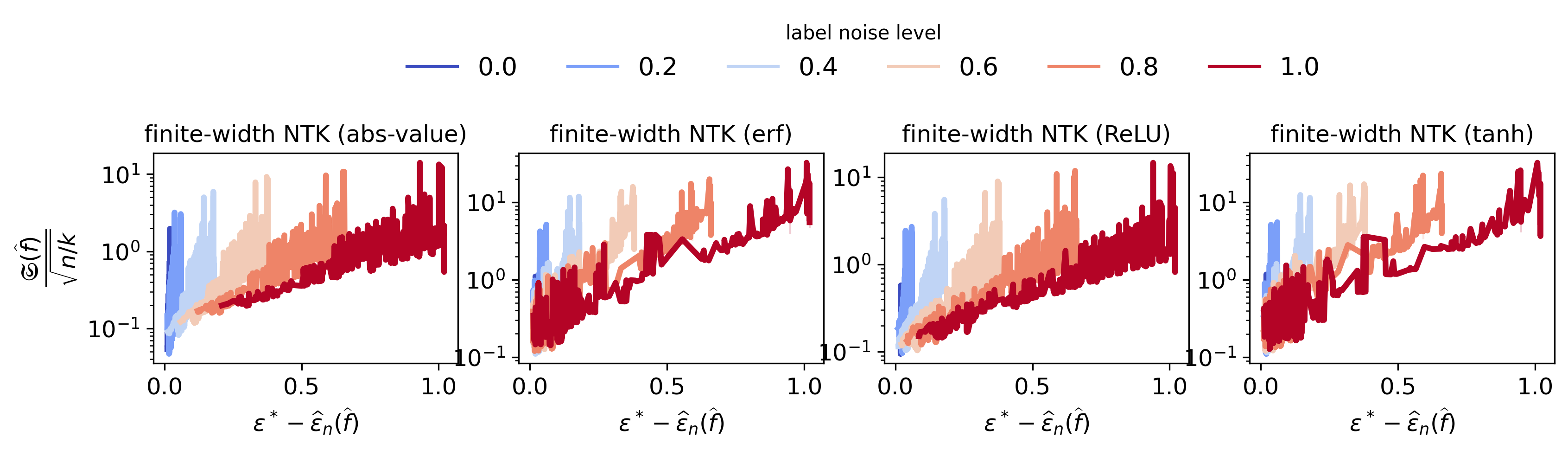}
    \caption{(Experiment 1) Finite-width NTK.}
\end{subfigure}
\hfill
\begin{subfigure}{\textwidth}
    \centering
    \includegraphics[width=\linewidth]{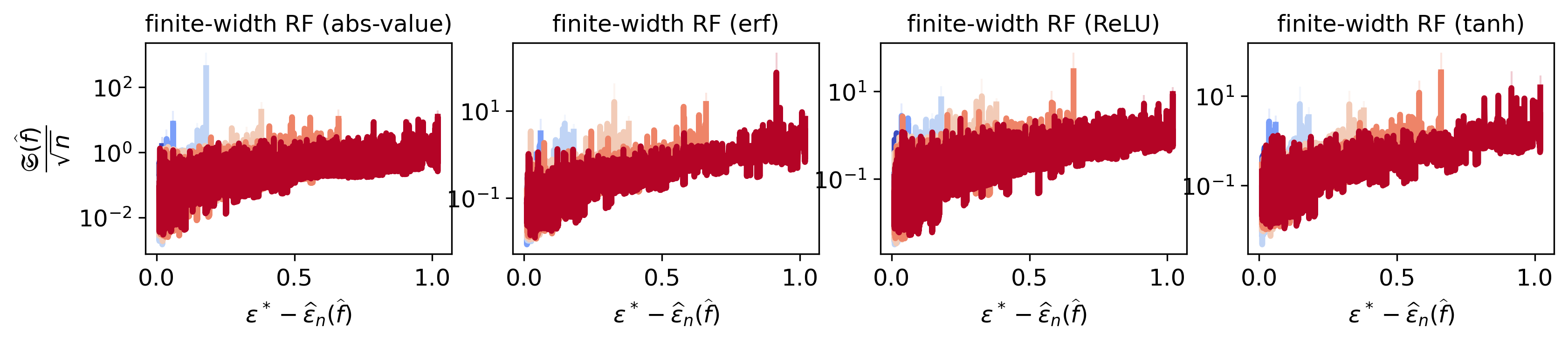}
    \caption{(Experiment 2) Finite-width RF.}
\end{subfigure}

\hfill

\begin{subfigure}{\textwidth}
    \centering
    \includegraphics[width=.49\linewidth]{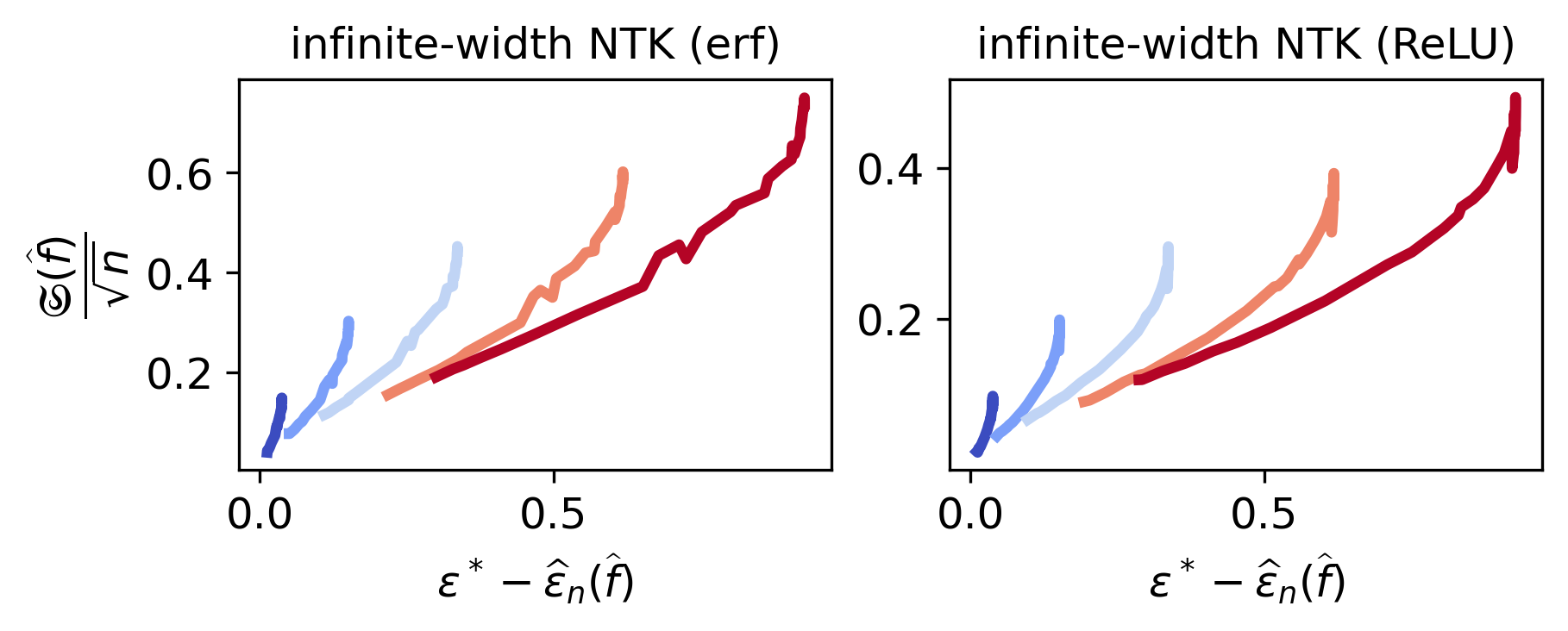}
    \includegraphics[width=.49\linewidth]{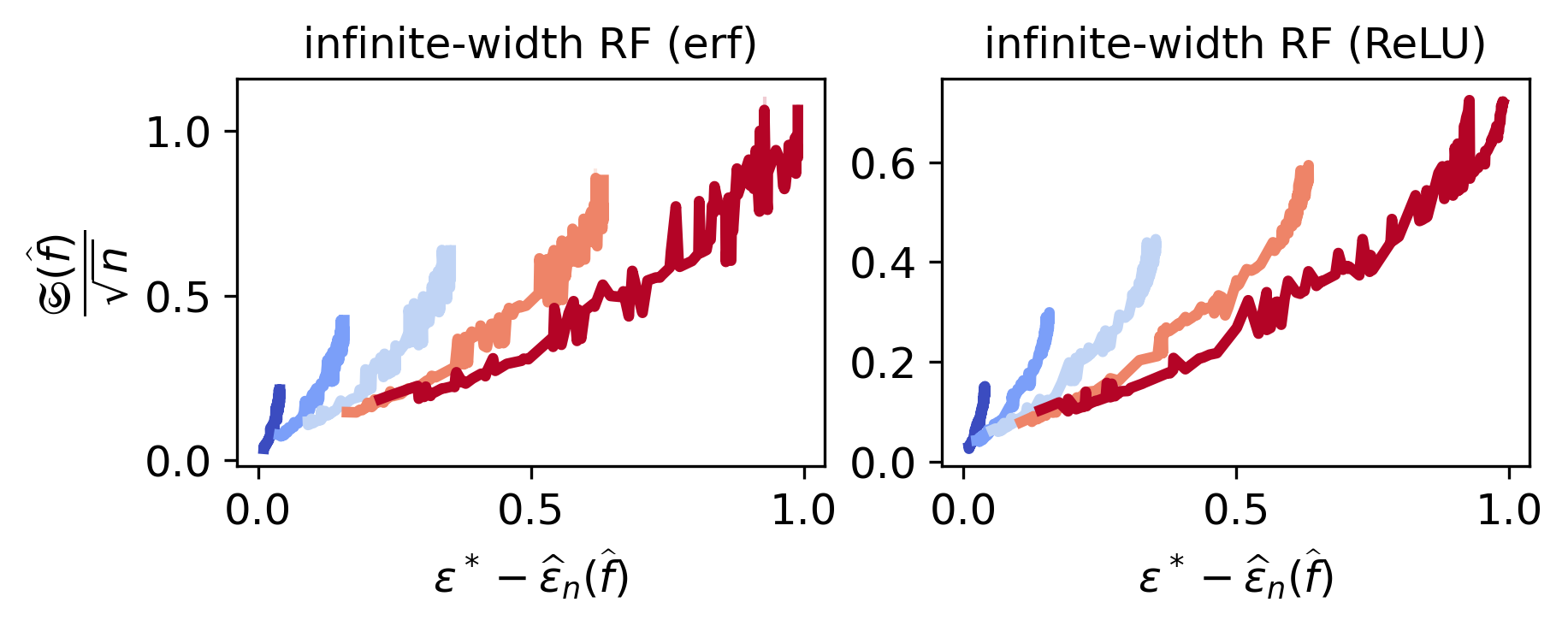}
    \caption{(Experiment 3) Infinite-width RF and NTK. Notice the linear trend slope $\Omega(1)$, for ReLU and gaussian error-function (erf) activation functions, in accordance with Theorem \ref{thm:infiniteRFNTKLaw}. This figure is complemented by Fig. \ref{fig:inf} with more activation functions.}
\end{subfigure}

\hfill

    \caption{Detailed version of Figure \ref{fig:alllaws}.}
    \label{fig:alllaws_detailed}
\end{figure}

\clearpage

\section{Law of robustness for kernel function classes over the sphere}


\subsection{Proof of Theorem \ref{thm:linear}}
\linear*
\begin{proof}[Proof of Theorem \ref{thm:linear}]
By direction computation, we have
\begin{eqnarray}
\begin{split}
\mathfrak{S}(g_w)^2 &:= \mathbb E_{x \sim \tau_d}[\|\nabla g_w(x)\|^2 - (x^\top \nabla g_w(x))^2]\\
&=  \mathbb E_{x \sim \tau_d}[\|w\|^2 - (x^\top w)^2] = \|w\|^2(1-1/d),
\end{split}
\label{eq:soblinear}
\end{eqnarray}
and so $\mathfrak{S}(g_w) \approx \|w\|=\Lip(g_w)$ for large $d$ (high dimensions). Thus, the analysis of the robustness of the linear model $g_w$ is reduced to the analysis of how the norm of $w$ varies with overfitting.

 \emph{-- First part (lower-bound).}
 Fix any $r \ge 0$, and let $C_d(r) := \{w \in \mathbb R^d \mid \mathfrak S(g_w) \le r\}$ and $B_d(r') := \{w \in \mathbb R^d \mid \|w\| \le r'\}$ the closed ball of radius $r' := r\sqrt{1-1/d}$ in $\mathbb R^d$. Thanks to the computation \ref{eq:soblinear}, it is clear that $C_d(r) \subseteq B_d(r')$. Let $\mathfrak R_n(C_d(r))$ be the Rademacher complexity of $C_d(r)$ w.r.t the sample $x_1,\ldots,x_n$. We deduce that $\mathfrak R_n(C_d(r)) \le \mathfrak R_n(B_d(r')) = r'/\sqrt{n} \lesssim r/\sqrt{n}$. Invoking standard results on $L_p$-loss generalization bounds for bounded function classes (see ~\cite{boucheron05}, for example\footnote{Because the noise is sub-Gaussian, we use a standard truncation argument to argue as if the squared loss as bounded.}), we obtain: w.p $1-\delta$, it holds for all $w \in \mathcal C_d(r)$ that
 $$
 \varepsilon^\star_{\mathrm{test}} \le \varepsilon(g_w) \le \widehat{\varepsilon}_n(g_w) + \dfrac{r}{\sqrt{n}} + r\sqrt{\dfrac{\log(2/\delta)}{n}}.
 $$
 The first part of the result then follows by taking $\delta = n^{-c}$, for any constant $c>0$, and the rearranging (while ignoring factors which are logarithmic in $n$).

 \emph{-- Second part (tightness).} Note that if $n < d$, then $XX^\top$ is invertible w.p $1$. Also, by construction, the min-norm interpolator has zero training error, i.e $\widehat{\varepsilon}_n(\widehat{g})=0$. It follows from the first part that $\mathfrak S(\widehat{g}) \ge \widetilde{\Omega}(\varepsilon_{\mathrm{tes}}^\star\sqrt{n})$ w.p $1-n^{-\Omega(1)}$. We now show that $\|\widehat{w}\| \le \mathcal O(\varepsilon^\star_{\mathrm{test}}\sqrt{n})$ w.p $1-n^{-\Omega(1)}$. Indeed, by standard random matrix theory (RMT) ~\cite{rmt}, w.p $1-e^{-\Omega(d)}$ over $X$, all the eigenvalues of the gram matrix $XX^\top$ are contained in in interval $[c_1,c_2]$, for absolute constant $c_1,c_2>0$. Let $z=(z_1,\ldots,z_n)$ be the iid $\zeta^2$-subGaussian noise vector of the dataset, so that $y_i = w_0^\top x_i + z_i$ for all $i \in [n]$. We deduce that w.p $1-e^{-\Omega(n)}$
 $$
 \|\widehat{w}\|^2 = y^\top (XX^\top)^{-1}y = z^\top (XX^\top)^{-1}z + \underbrace{w_0^\top X^\top(XX^\top)^{-1}Xw_0}_{\le \|w_0\|^2 \le 1}-2w_0^\top X^\top(XX^\top)^{-1} z,
 $$
 By standard concentration standard concentration for the sub-Gaussian random vector $z$ combined with previous remark on the eigenvalues of $XX^\top$, the first and last terms in the above display are w.p $1-e^{-\Omega(n)}$ at most $\mathcal O(n\zeta^2)$, from which the second part of result follows.
\end{proof}

\subsection{RKHS norm of a memorizer}
We now extend Theorem \ref{thm:linear} to general kernel function classes. For $r \ge 0$, let $B_K=B_K(r) := \{f \in \mathcal H_K \mid \|f\|_{\mathcal H_K} \le r\}$ be the ball of radius $R$ in $\mathcal H_K$. Also, let $T_K:L^2(\tau_d) \to L^2(\tau_d)$ be the induced integral operator defined for every $f \in L^2(\tau_d)$ by
\begin{eqnarray}
T_K f: \sphere \to \mathbb R,\,(T_Kf)(x) = \int_{\sphere}K(x,x')f(x')d\tau_d(x').
\end{eqnarray}
This is a compact positive operator and thus has countably many eigenvalues $\lambda_1 \ge \lambda_2 \ge \ldots$, all of which are nonnegative.

We start with the following auxiliary lemma which establishes that w.h.p, and function  $f \in \mathcal H_K$ which memorizes even a fraction of the generic dataset $\mathcal D_n$ must have RKHS norm at least $\sqrt{n}$. 
\begin{restatable}{lm}{}
 It holds w.p $1-n^{-\Omega(1)}$ (independent of $\varepsilon$) over the generic dataset $\mathcal D_n$ that: every $f \in \mathcal H_K$ which $\varepsilon$-memorizes $\mathcal D_n$ satisfies
\begin{eqnarray}
\|f\|_{\mathcal H_K} \ge \widetilde{\Omega}(\frac{(\varepsilon^\star_{\mathrm{test}}-\varepsilon)}{\sqrt{\trace(T_K)}}\sqrt{n}).
\end{eqnarray}
\label{lm:gen}
\end{restatable}
\begin{restatable}{rmk}{}
  We make the following important notes above the above theorem.
\begin{itemize}
\item In the above lower-bound, thanks to concentration arguments the trace term $\trace(T_K)$ can be replaced by a sample version $(1/n)\sum_{i=1}^n K(x_i,x_i)$.
\item \cite[Theorem 1]{belkin18} establishes a lower-bound of the form $\|f\|_{\mathcal H_K} \ge Ae^{Bn^{1/d}}$, for absolute constants $A,B>0$. For fixed $d=\mathcal O(1)$, this bound is better than the $\widetilde{\Omega}(\sqrt{n})$ bound above, but becomes unspecial when $d$ goes to infinity, say at the same rate as $n$. Indeed, for such $d$, the bound in \cite{belkin18} predicts $\|f\|_{\mathcal H_K} \ge Ae^{B}$, a lower-bound which is $\mathcal O(1)$,  while our bound in Lemma \ref{lm:gen} ensures $\|f\|_{\mathcal H_K} \gtrsim \sqrt{n} \to \infty$.
\end{itemize}
\end{restatable}
\begin{proof}[Proof of Lemma \ref{lm:gen}]
  First note that one has $\sup_{x \in \sphere} K(x,x) < \infty$ since $\sphere$ is compact and $K:\sphere \times \sphere \to \mathbb R$ is continuous by hypothesis.
  Now, for any $r \ge 0$, the Rademacher complexity of the RKHS ball $B_K(r)$ is upper-bounded by $r\sqrt{\mathbb E_{x \sim \tau_d}K(x,x)]/n} = r\sqrt{\trace(T_K)/n}$.
  By classical theory of generalization theory for $L_p$-losses (e. see ~\cite{boucheron05}), it holds w.p $1-\delta$ over the dataset $\mathcal D_n$ that
$$
\varepsilon^\star_{\mathrm{test}} \le \epsilon(f) \le \widehat{\varepsilon}_n(f) + C\frac{r}{\sqrt{n}}(\sqrt{\trace(T_K)}+ \sqrt{2\log(1/\delta)}),\,\forall f \in B_K(r).
$$
The claim then follows with $\delta=n^{-c}$ for an absolute constant $c>0$, and then simplifying to get $r \ge \Omega((\varepsilon^\star_{\mathrm{test}}-\varepsilon)\sqrt{n/\log n}) \ge \widetilde{\Omega}(\varepsilon^\star_{\mathrm{test}}\sqrt{n})$ w.p $1-n^{-c}$.
\end{proof}

{\bfseries Not game over yet.} Our concern is robustness in the ambient space $\sphere$ in which the data lives. A priori, there is no direct implication between large norm does not imply large Sobolev-seminorm (though the converse is true\footnote{Indeed, $\mathfrak{S}(f) \le \Lip(f) \le \|f\|_{\mathcal H_K}$, where the last inequality is classical  (see \cite{inductivebietti}, for example).}). That is, a priori, we cannot directly salvage a lower-bounds for the nonrobustness of a memorizer $f$ by exploiting the lower-bound on its RKHS norm given by Lemma \ref{lm:gen}. For this we need to exploit the geometric structure of the specific kernel $K$. This will allow us convert the lower-bound on $\|f\|_{\mathcal H_K}$ into lower-bounds on nonrobustness $\mathfrak{S}(f)$. The rest of the manuscript is more or less dedicated to this.


\subsection{Quantitative tradeoff between memorization and robustness}
\begin{restatable}{df}{}
  Let $C_K(X) \in \mathbb R^{n \times n}$ be the covariance matrix of the random vector $(\sqrt{d}K(x,x_1),\ldots,\sqrt{d}K(x,x_n)) \in \mathbb R^n$ for $x \sim \tau_d$ independent of the $x_i$'s. Also define the following ``condition number'' of the design matrix $X$ relative to the kernel $K$ by
  \begin{eqnarray}
  \alpha_K(X) := \dfrac{\lambda_{\max}(K(X,X))}{\lambda_{\min}(C_K(X))} \frac{1}{n}\sum_{i=1}K(x_i,x_i).
  \end{eqnarray}
\end{restatable}
Note that $\alpha_K(X)$ is a random variable, since it dependences on the design matrix $X$. Recall the definition of memorization in Definition \ref{df:memo}. The following generic result will be the main stepping stone for most of the results in the remainder of this section and the next. As before, let $\varepsilon^\star_{\mathrm{test}}$ be the Bayes-optimal error for the problem and let $\varepsilon$ be any error threshold in the interval $[0,\varepsilon^\star_{\mathrm{test}})$.
\begin{restatable}[Law of robustness for the "representer" subspace]{thm}{}
 The following holds w.p $1-n^{-\Omega(1)}$ over the generic dataset $\mathcal D_n$: every $f \in \lspan_K(X)$ which $\varepsilon$-memorizes $\mathcal D_n$ satisfies
\begin{eqnarray}
\mathfrak{S}(f) \ge \widetilde{\Omega}((\varepsilon^\star_{\mathrm{test}}-\varepsilon)\sqrt{\frac{n}{\alpha_K(X)}}).
\end{eqnarray}
In particular, if $K(X,X)$ is invertible (which necessarily implies we are in a high-dimensional regime $d \ge n$), then for min-norm interpolator $\widehat{f}_n \in \lspan_K(X)$, it holds that $\widehat{\varepsilon}_n(\widehat{f}_n)=0$ almost-surely and $\mathfrak{S}(\widehat{f}_n) \ge \widetilde{\Omega}(\varepsilon^\star_{\mathrm{test}} \sqrt{\dfrac{n}{\alpha_K(X)}})$ w.p $1-n^{-\Omega(1)}$.
\label{thm:generalkernelbound}
\end{restatable}
\begin{proof}
For any $c \in \mathbb R^n$, let $f_c:\sphere \to \mathbb R$ be the function defined by $f_c(x) := \sum_{i=1}^n c_iK(x_i, x)=c^\top K(X,x)$. The Poincar\'e inequality for the probability space $(\sphere,\tau_d)$ gives
  \begin{eqnarray*}
    \begin{split}
  \mathfrak{S}(f_c)^2  &\ge d \cdot \var_{x \sim \tau_d} (f_c(x)) = d\cdot \var_{x \sim \tau_d} (c^\top K(X,x))\\
  &= c^\top \cov(\sqrt{d} K(X,x)) c =:c^\top C_K(X)c.
    \end{split}
  \end{eqnarray*}
  From the well-known identity $\|f_c\|_{\mathcal H_K}^2 = c^\top K(X,X) c$, one deduces
  \begin{eqnarray}
    \begin{split}
  \mathfrak{S}(f_c)^2 \ge c^\top C_K(X) c \ge \|c\|^2 \lambda_{\min}(C_K(X)) &\ge \|f_c\|_{\mathcal H_K}^2 \dfrac{\lambda_{\min}(C_K(X))}{\lambda_{\max}(K(X,X))}\\
  &= \|f_c\|_{\mathcal H_K}^2\frac{\trace(T_K)}{\alpha_K(X)}.
  \label{eq:rabbithole}
    \end{split}
  \end{eqnarray}
  Invoking Lemma \ref{lm:gen} with $\delta = n^{-\Omega(1)}$ ensures $\|f_c\|^2_{\mathcal H_K} \ge \widetilde{\Omega}((\varepsilon^\star_{\mathrm{test}}-\widehat{\varepsilon}_n(f_c))\sqrt{n})$. The first part of the result follows upon combining with \eqref{eq:rabbithole}.
  
The second part is a direct consequence of the first part and the fact that $\widehat{f}_n(x_i) = y_i$ for all $i \in [n]$ and so $\widehat{\varepsilon}_n(\widehat{f}_n) = 0$.
\end{proof}

\subsection{Example: Ordinary linear models}
\label{subsec:linearkernel}
As an example, consider the RKHS on the unit-sphere $\sphere$, induced by the trivial kernel $K_{\mathrm{id}}(x,x') = x^\top x'$. One immediately computes $K_{\mathrm{id}}(X,X) = C_{K_{\mathrm{id}}}(X) = XX^\top$ and the representer subspace is $\lspan_{K_{\mathrm{id}}}(X) = \lspan(X) := \{x \mapsto \sum_{i=1}^n c_i x_i^\top x \mid c \in \mathbb R^n\}$. Also, one computes $\trace(T_K) = \mathbb E_{x \sim \tau_d}[K_{\mathrm{id}}(x,x)] = \mathbb E_x\|x\|^2 = 1$ and
$$
\alpha_{K_{\mathrm{id}}}(X) := \dfrac{\lambda_{\max}(K_{\mathrm{id}}(X,X))\trace(T_K)}{\lambda_{\min}(C_{K_{\mathrm{id}}}(X))} = \dfrac{\lambda_{\max}(XX^\top)}{\lambda_{\min}(XX^\top)}
=:\mathrm{cond}(X)^2,
$$
where $\mathrm{cond}(X)$ is the \emph{condition number} (the one from classical linear algebra) of the design matrix. Theorem \ref{thm:generalkernelbound} then predicts that w.p $1-o(1)$ over the generic dataset $\mathcal D_n$, every $f \in \lspan_{K_{\mathrm{id}}}(X)$ which $\varepsilon$-memorizes $\mathcal D_n$ must veify
\begin{eqnarray}
\mathfrak{S}(f) \ge \frac{(\varepsilon^\star_{\mathrm{test}}-\varepsilon)}{\mathrm{cond}(X)}\Omar(\sqrt{n}). 
\end{eqnarray}
From standard random matrix theory (RMT) \cite{rmt}, we know that if $n \asymp d$ are sufficiently large with $n/d \le \gamma_1 < 1$, then $\mathrm{cond}(X) = \Theta(1)$ w.p $1-e^{-\Omega(d)}$. Putting things together we obtain the following corollary to Theorem \ref{thm:generalkernelbound}.
\begin{restatable}{cor}{}
  If large $n \asymp d$ such that $n/d \le \gamma_1 < 1$, then it holds w.p $1-n^{-\Omega(1)}$ that over the dataset $\mathcal D_n$: every linear model $f \in \lspan_{K_{\mathrm{id}}}(X)$ which $\varepsilon$-memorizes $\mathcal D_n$ satisfies
  $\mathfrak{S}(f) \ge \Omar((\varepsilon^\star_{\mathrm{test}}-\varepsilon)\sqrt{n})$.
In particular the min-norm interpolator $\widehat{f}_n(x):= x^\top X(XX^\top)^{-1} y$ verifies 
$\mathfrak{S}(\widehat{f}_n) \ge \Omar(\varepsilon^\star_{\mathrm{test}} \sqrt{n})$ w.p $1-n^{-\Omega(1)}$.
  \label{cor:linearlaw}
\end{restatable}

\paragraph{Gaussian kernel.} Consider the case of a Laplace type kernel on $\sphere$ defined by $K_\beta(x,x') := \exp(-s^{-1}\|x-x'\|^\beta)$, with smoothness parameter $\beta > 0$ and bandwidth parameter $s>0$. This is of course a dot-product kernel with $\phi(t) = e^{-(2(1-t))^{\beta/2}/s}$. 

In the next subsection, we will extend this corollary to RKHS function classes corresponding to infinite-width neural networks for certain activation functions.



\section{Laws of robustness for dot-product kernels}
Suppose the kernel $K$ is a dot-product function in the following sense
\begin{restatable}[Kernels of dot-product type]{df}{}
\label{df:dotprodtype}
A kernel $K:\sphere \times \sphere \to \mathbb R$ is said to be a radial or dot-product kernel if there exists a continuous function $\phi:[-1,1] \to \mathbb R$ such that
\begin{eqnarray}
  K(x,x')= \phi(x^\top x'),\,\forall x,x' \in \sphere.
  \label{eq:dotfunction}
\end{eqnarray}
\end{restatable}
Examples of dot-product kernels are abundant in machine learning. To name a few, let us mention:
\begin{itemize}
  \item The gaussian kernel $K_{\mathrm{Gauss}}(x,x') := \phi_{\mathrm{Gauss}}(x^\top x')$, where $\phi_{\mathrm{Gauss}}(t) = e^{-(2-2t)/s^2}$ ( for some bandwidth parameter $s>0$). This kernel is known to the kernel corresponding to the infinite-width random Fourier features networks ~\cite{rf}.
  \item The Laplace kernel $K_{\mathrm{Lap}}(x,x') := \phi_{\mathrm{Lap}}(x^\top x')$, with $\phi_{\mathrm{Lap}}(t) = e^{-\sqrt{2-2t}/s}$. 
  \item General exponential-type kernels given by $K_s(x,x') := \phi_{\mathrm{Exp}(s)}(t)$, for which $\phi_{\mathrm{Exp}(s)}(t) = e^{-(2-2t)^{s/2}/s}$. Note that the gaussian and Laplace kernels correspond respectively to $s=2$ and $s=1$.
  \item Polynomial kernels $K_{c,p}(x,x') := (c + x^\top x')^p$, where $p > 0$ is the degree (allowed to be fractional!) and $c \ge 0$ is an offset parameter. These are indeed dot-product kernels with $\phi(t) = (c + t)^p$. The kernel considered in section \ref{subsec:linearkernel} is a linear kernel with $c=0$ and $p=1$.
\end{itemize}
In the case of dot-product kernels, many things simplify. For example, for the associated kernel integral operator $T_K$, one has
\begin{eqnarray}
  \|T_K\|_{op} \le \trace(T_K) = \mathbb E_{x \sim \tau_d}[K(x,x)] = \phi(1).
\end{eqnarray}


\begin{restatable}{thm}{}
 Suppose Condition \ref{cond:awfulkappa} holds and $n,d \to \infty$ such that $n/d \to \gamma_1 \in [0, 1)$. Then
    \begin{eqnarray}
    \begin{split}
      \lambda_{\min}(C_K(X)) &\ge 
      \Omega_{\mathbb P}(1),\,
      \lambda_{\max}(K(X,X)) \le
      \mathcal O_{\mathbb P}(1),\, \alpha_K(X) \le \mathcal O_{\mathbb P}(1).
      \end{split}
    \end{eqnarray}
\label{thm:kernelCmin}
\end{restatable}
Theorem \ref{thm:kernelCmin} provides us with a lower-bound on the condition number $\alpha_K(X)$ using very macroscopic information abound the dot-product function $\phi$. To proof this theorem we will need the following auxiliary result (proved in section XXX) which is important in its own right, and is therefore stated as a theorem.

\begin{restatable}{thm}{}
Let $V$ be an $m \times d$ random matrix with iid rows sampled from $\tau_d$.
Consider the random mapping mapping $\Phi:\sphere \to \mathbb R^m$ defined by $\Phi(x) = \varphi(Vx)$, where $\varphi:[-1,1] \to \mathbb R$ is a continuous function which is thrice continuously-differentiable at $0$. For $x \sim \tau_d$, let $C_{\Phi} \in \mathbb R^{m \times m}$ be the covariance matrix of the random vector $\sqrt{d}\Phi(x) \in \mathbb R^m$.  In the limit $m,d \to \infty$ such that $m/d \le \gamma_1 < 1$, it holds that $\lambda_{\min}(C_\Phi) \to \varphi'(0)^2(1-\sqrt{\gamma_1})^2$ almost-surely.
\label{thm:randVrestated}
\end{restatable}

\begin{proof}[Proof of Theorem \ref{thm:kernelCmin}]
From Theorem \ref{thm:randVrestated}, with $m=n$, $V = X$, and $\varphi = \phi$ (the dot-product function of the kernel $K$), we have $\lambda_{\min}(C_K(X)) = \phi'(0)^2(1-\sqrt{\gamma_1})^2-o_{\mathbb P}(1) \ge \Omega_{\mathbb P}(1)$. It remains to upper-bound $\lambda_{\max}(K(X,X))$. For this, it suffices to apply ~\cite[Theorem 2.1]{elkaroui2010} to obtain that $\|K(X,X)-K(X,X)^{\mathrm{lin}}\|_{op} = o_{\mathbb P}(1)$, where the matrix $K(X,X)^{\mathrm{lin}} \in \mathbb R^{n \times n}$ has entries 
$$
K(X,X)^{\mathrm{lin}}_{i,\ell} := \phi(0) + \frac{\phi''(0)}{2n} + \phi'(0)x_i^\top x_\ell + v\delta_{i,\ell},
$$
with $v := \phi(1)-\phi(0)-\phi'(0)$ and $\delta_{i,\ell} = 1$ if $i = \ell$ and $\delta_{i,\ell}=0$ otherwise. Noting that the finite-rank (here rank 1) perturbations do not affect the limiting spectral distribution of a random matrix, we deduce that \begin{eqnarray*}
\begin{split}
\lambda_{\max}(K(X,X)) &= \|K(X,X)^{\mathrm{lin}}\|_{op} + o_{\mathbb P}(1) = \|\phi'(0)XX^\top+vI_n\|_{op}+ o_{\mathbb P}(1)\\
&\le |\phi'(0)|(1-\sqrt{\gamma_1})^2 + |v|+o_{\mathbb P}(1),\text{ by Bin-Yai (\mytodo{add ref})}\\
&= \mathcal O_{\mathbb P}(1).
\end{split}
\end{eqnarray*}
Finally, we deduce that $\alpha_K(X) := \dfrac{\lambda_{\max}(K(X,X))\phi(1)}{\lambda_{\min}(C_K(X))} \le \dfrac{\mathcal O_{\mathbb P}(1)}{\Omega_{\mathbb P}(1)} = \mathcal O_{\mathbb P}(1)$.
\end{proof}
The following result which extends Corollary \ref{cor:linearlaw}, is a Corollary to \ref{thm:generalkernelbound}. We state it as a theorem because it is important in its own right, and will in tern give laws of robustness for kernels induced by certain infinite-width neural networks, for example (section \ref{sec:infinitewidth}).
\lawdotprodkernel*

\begin{proof}
Follows Theorems \ref{thm:generalkernelbound} and \ref{thm:kernelCmin}.
\end{proof}

\paragraph{Example: Exponential-type kernels.} As an example, consider an exponential-type kernel $K_\beta(x,x') := e^{-\|x-x'\|^\beta/s}$, where $\beta > 0$ is a "smoothness" parameter and $s$ is a bandwidth parameter. As discussed in the paragraph just after \eqref{eq:dotfunction}, such is a dot-product kernel with dot-product function $\phi_\beta(t) = e^{-(2-2t)^{\beta/2}/s}$ which is infinitely continuosly differentiable with $\phi_\beta'(0) = \dfrac{\beta 2^{\beta/2-1}e^{-2^{\beta/2}/s}}{s} \ne 0$. Thus, such kernels satisfy Condition \ref{cond:awfulkappa}, and we deduce the following Corollary to Theorem \ref{thm:lawdotprodkernel}.

\begin{restatable}[Law of robustness for exponential-type kernels]{cor}{}
In the limit $n,d \to \infty$ such that $n/d \le \gamma_1 < 1$ the following holds in probability: every $f \in \lspan_{K_\beta}(X)$ which $\varepsilon$-memorizes $\mathcal D_n$ satisfies $\mathfrak{S}(f) \gtrsim (\varepsilon^\star_{\mathrm{test}}-\varepsilon)\sqrt{n}$.
In particular, if the gram matrix $K_\beta(X,X)$ is nonsingular, then for the min-norm interpolator $\widehat{f}_n \in \lspan_{K_\beta}(X)$, it holds in probability that $\mathfrak{S}(\widehat{f}_n) \gtrsim \varepsilon^\star_{\mathrm{test}} \sqrt{n}$.
\end{restatable}
In particular, the above theorem applies to
\begin{itemize}
\item Two-layer infinite-width neural networks with random Fourier features, the corresponding RKHS is precisely that induced by the Gaussian kernel \cite{rf,rf2}.
\item Certain infinite-width neural networks in RF / NTK regime.
\item etc.
\end{itemize}

An analogous result holds for polynomial kernels with positive degree.

\section{Proof of Theorem \ref{thm:dico}}
\label{subsec:dicoproof}


\dico*
\begin{proof}
One computes the variance of any $f_c \in \lspan_{K_\Phi}(X)$ for random $x \sim \tau_d$, as follows
\begin{eqnarray}
\begin{split}
\mathfrak{S}(f_c)^2 \ge d\cdot \var_{x \sim \tau_d}(f_c(x)) &= d\cdot  \var_{x \sim \tau_d}(c^\top \Phi(X)\Phi(x))\\
&= c^\top \Phi(X)\cov_{x \sim \tau_d}(\sqrt{d}\Phi(x))\Phi(X)^\top c\\
&\ge \lambda_{\min}(\cov_{x \sim \tau_d}(\sqrt{d}\Phi(x))\|\Phi(X)^\top c\|^2\\
&= \lambda_{\min}(C_\Phi)\|f_c\|_{\mathcal H_{K_\Phi}}^2,
\end{split}
\label{eq:manip}
\end{eqnarray}
where the $m \times m$ psd matrix $C_\Phi$ is the covariance matrix of the random vector $\sqrt{d}\Phi(x) \in \mathbb R^m$.

We apply Lemma \ref{lm:gen}. For this, we need to compute the trace of the kernel integral operator $T_{K_\Phi}$, which equals
$$
\trace(T_{K_\Phi}) = \mathbb E_{x \sim \tau_d}[K_\Phi(x,x)] = \mathbb E_{x \sim \tau_d}[\|\Phi(x)\|^2] = \|\Phi\|^2_{L^2(\tau_d)}.
$$
The result then follows upon invoking Lemma \ref{lm:gen} to lower-bound $\|f_c\|_{\mathcal H_{\Phi}}$ and then invoking \eqref{eq:manip} to lower-bound $\mathfrak{S}(f_c)$.
\end{proof}

\section{Neural networks in infinite-width RF and NTK regimes}
\label{sec:infinitewidth}
We now place ourselves in the exact kernel regimes (where $k=\infty$), for two-layer neural networks in RF and NTK regimes.

\subsection{Proof of Theorem \ref{thm:infiniteRFNTKLowerbound} (tightness of lower-bound in Theorem \ref{thm:infiniteRFNTKLaw})}
We recall the following lemma, needed for the proof.
\invertibleK*
\begin{proof}
Follows from Theorems \ref{thm:kernelCmin} and \ref{thm:randVrestated}.
\end{proof}

We now prove Theorem \ref{thm:infiniteRFNTKLowerbound}, namely the tightness of the lower-bound in Theorem \ref{thm:infiniteRFNTKLaw}.
\infiniteRFNTKLowerbound*
\begin{proof}
Let $\mathcal H$ be the RKHS induced by the infinite-width ReLU random features kernel. Note that the coefficients of $\widehat{f}_\rf^\infty$ in the representer subspace $\lspan_{K^\infty_\rf}(X) \subseteq \mathcal H$ are given by $\widehat{c} := K^\infty_\rf(X,X)^{-1} y \in \mathbb R^n$, so that $\widehat{f}_\rf^\infty(x) = \widehat{c}^\top K^\infty_\rf(X,x)=\widehat{c}^\top \phi_1(Xx)$, where $\phi_1(Xx):=(\phi_1(x_1^\top x),\ldots,\phi_1(x_n^\top x)) \in \mathbb R^n$ and $\phi_1$ is the order-1 arc-cosine dot-product function defined in \eqref{eq:arccos}. Thanks to \cite[Lemma 1]{biettiinvariance}, we know that the Lipschitz constant of $\widehat{f}_\rf^\infty$ is upper-bounded by its RKHS norm in $\mathcal H$. Thus, one computes
\begin{eqnarray}
\begin{split}
\Lip(\widehat{f}_{\rf}^\infty)^2 &\le  \|\widehat{f}_\rf^\infty\|_{\mathcal H} = \widehat{c}^\top K_\rf^\infty(X,X) \widehat{c} = y^\top K^\infty_\rf(X,X)^{-1} y\\
&\le \lambda_{\max}(K^\infty_\rf(X,X)^{-1})\|y\|^2 \le \frac{\|y\|^2}{\lambda_{\min}(K^\infty_\rf(X,X))}. 
\end{split}
\end{eqnarray}
By Lemma \ref{lm:invertibleK}, we know that $\lambda_{\min}(K_\rf^\infty(X,X)) \ge \Omega(1)$ w.p $1-o(1)$. 
Also, each label $y_i$ in the dataset $\mathcal D_n$ is $\zeta^2$-sub-Gaussian around $x_i^\top w_0$ with $\|w_0\| \le 1$ and $\|x_i\|=1$, we know that $\|y\|^2 = \mathcal O(\zeta^2 n)$ w.p $1-o(1)$. Putting things together gives the result.
\end{proof}

\section{Finite-width random features regime}

\subsection{Simplifying the matrix $C_\sigma(W)$, the covariance matrix of $\sqrt{d}\sigma(Wx)$ for $x \sim \tau_d$}

Let us restrict our attention to the following class of activation functions $\sigma$. For concreteness, the reader may think of the ReLU of the absolute value activation functions.
\begin{restatable}{cond}{hom}
The activation function $\sigma$ is $1$-Lipschitz and positive-homogeneous of order $1$.
\label{cond:hom}
\end{restatable}    

The following remarkable property of positive-homogeneous functions will be very helpful in the sequel.
\begin{restatable}[Kernel function induced by homogeneous activations
~\cite{Erez2016}]{prop}{}
\label{prop:catalan}
If $h:\mathbb R \to \mathbb R$ is positive-homogeneous of order $p \ge 0$, then for every $a,b \in \sphere$ we have the identity
\begin{equation}
\label{eq:catalan}
    \mathbb E_{x \sim \tau_d}[h(x^\top u)h(x^\top v)]=\dfrac{C_{2,2p}}{C_{d,2p}}\phi_h(u^\top v),\text{ with }\phi_h(t) := \frac{1}{2\pi}\int_{0}^{2\pi}h(\cos u)h(\cos(u-\arccos t))\dif u,
\end{equation}
where $C_{d,p} := 2^{p/2-1}\cdot \dfrac{d\cdot \Gamma((d+p)/2)}{\Gamma((d+2)/2)}$.
\end{restatable}
By the above proposition, the order-$1$ positive-homogeneity of the activation function $\sigma$ implies the existence $\phi_\sigma:[-1,1] \to \mathbb R$ such that if the rows if $u,v \in \sphere$ (i.e $u$ and $v$ are unit-vectors), then
\begin{eqnarray}
\mathbb E_{x \sim \tau_d}[\sigma(x^\top u)\sigma(x^\top v)] = \frac{1}{d}\phi_\sigma(u^\top v).
\label{eq:phifunction}
\end{eqnarray}
For example, if $\sigma$ is the ReLU activation function, then
\begin{eqnarray}
\phi_\relu(t) = \frac{1}{2\pi}(t\arccos(-t) + \sqrt{1-t^2}),\,\forall t \in [-1,1].
\label{eq:reluphi}
\end{eqnarray}

Importantly, the function $\phi_\sigma$ depends on the activation function $\sigma$ alone (and not on problem parameters like $n$, $d$, $k$, etc.). The following lemme is a first step towards a spectrally consistent linearization of the troublesome matrix $C_\sigma(W)$.

\begin{restatable}{lm}{}
Suppose the function $\phi_\sigma$ appearing in \eqref{eq:phifunction} is thrice continuously-differentiable at $0$ and the rows of $W$ are unit-vectors, then
\begin{eqnarray}
C_\sigma(W) = \phi_\sigma(WW^\top) - \phi_\sigma(0)1_k1_k^\top + E,
\end{eqnarray}
for some $k \times k$ matrix with $\|E\|_{op} = \mathcal O(1/d)$.
\label{lm:homrep}
\end{restatable}
\begin{proof}
The $(j,\ell)$ entry of $C_\sigma(W)$ is given by
\begin{eqnarray*}
\begin{split}
c_{j,\ell} &= d\cdot(\mathbb E_{x \sim \tau_d}[\sigma(x^\top w_j)\sigma(x^\top w_\ell)] - \mathbb E_{x \sim \tau_d}[\sigma(x^\top w_j)]\mathbb E_{x \sim \tau_d}[\sigma(x^\top w_\ell)])\\
&= \phi_\sigma(w_j^\top w_\ell) - d\cdot \mathbb E_{x \sim \tau_d}[\sigma(x^\top w_j)]\mathbb E_{x \sim \tau_d}[\sigma(x^\top w_\ell)]).
\end{split}
\end{eqnarray*}
On the other hand, because $w_j$ and $w_\ell$ are unit-vectors and the distribution of $x$ is isotropic, we may write
\begin{eqnarray*}
\begin{split}
\mathbb E_{x \sim \tau_d}[\sigma(x^\top w_j)]\mathbb E_x[\sigma(x^\top w_\ell)] &= \mathbb E_{x \sim \tau_d}[\sigma(x^\top w_j)]^2 = \mathbb E_{w \sim \tau_d}\mathbb E_{x \sim \tau_d}[\sigma(x^\top w_j)]^2\\
&=\mathbb E_{(x,z) \sim \tau_d \otimes \tau_d}\mathbb E_{w}[\sigma(x^\top w)\sigma(z^\top w)]\\
&= \mathbb E_{(x,z) \sim \tau_d \otimes \tau_d}[\phi_\sigma(x^\top z)]=\phi_\sigma(0) + \mathcal O(\frac{1}{d^2}),
\end{split}
\end{eqnarray*}
where the last step is thanks to a taylor expansion of $\phi_\sigma$ around $0$ and the fact that $\mathbb E[x^\top z] = 0$ due to isotropy and independence of $x$ and $z$.
Putting things together then gives
$$
c_{j,\ell} = \phi_\sigma(w_j^\top w_\ell) - \phi_\sigma(0) + \mathcal O(\frac{1}{d^2}),
$$
from whence the result follows.
\end{proof}


\subsection{Spectrally consistent linearizations of $\widetilde{C}_\sigma(W)$ for random $W$}
Let $W \in \mathbb R^{k \times d}$ be a random matrix with independent rows uniformly on the unit-sphere $\sphere$ (i.e according to the uniform distribution $\tau_d$ thereupon), and let $\widetilde{C}_\sigma(W)$ be the $k \times k$ psd matrix defined in \eqref{eq:Ctilde}. If the input dimension $d$ is sufficiently large, then for distinct $j,\ell \in [k]$, it is clear that $w_j^\top w_\ell=\mathcal O(1/d)$ w.h.p. Thus, if we suppose the function $\phi_\sigma$ defined in \eqref{eq:phifunction} is sufficiently smooth in a neighborhood of $t=0$, one can hope to Taylor-expand $\widetilde{C}_\sigma(W)$ entry-wise.
In ~\cite{elkaroui2010,justinterpolate}, such arguments are made more precise and quantitative estimates for the extreme eigenvalues of $\widetilde{C}_\sigma(W)$ are obtained via a linearization trick. 

Now, consider the $k \times k$ matrix $\widetilde{C}_\sigma(W)^{\mathrm{lin}}$ with entries given by
\begin{eqnarray}
\widetilde{C}_\sigma(W)^{\mathrm{lin}}_{j,\ell} := \phi_\sigma(0) + \frac{\phi_\sigma''(0)}{2d} + \phi_\sigma'(0)w_j^\top w_\ell + (\phi_\sigma(1)-\phi_\sigma(0) - \phi_\sigma'(0)) \delta_{j,\ell}.
\end{eqnarray}
We now show that that the curvature coefficients $\beta_0(\sigma)\ge 0$, $\beta_1(\sigma) \ge 0$, and $\beta_\star(\sigma) \in \mathbb R$ defined in \eqref{eq:betas} are precisely the low-order coefficients in the above polynomial.
\begin{restatable}{lm}{}
We have the following identities
\begin{eqnarray}
\begin{split}
\beta_0(\sigma) &= \phi_\sigma(0),\\
\beta_1(\sigma) &= \phi_\sigma'(0),\\
\beta_\star(\sigma) &= \phi_\sigma(1) - \phi_\sigma(0) - \phi_\sigma'(0).
\end{split}
\end{eqnarray}
\end{restatable}
\begin{proof}
By definition, $\phi_\sigma(t) := d\cdot \mathbb E_{x\sim \tau_d}[\sigma(x^\top w_j)\sigma(x^\top w_\ell)]$, where $t:=w_j^\top w_\ell$. If $t=0$, then $w_j$ and $w_\ell$ are orthogonal, and $x^\top w_j$ and $x^\top w_\ell$ are (statistically) independent, with the same distribution, which is approximately $N(0,1/d)$ (the approximation error in Kolmogorov distance is of order $O(1/\sqrt{d})$). We deduce that $\phi_\sigma(0) = \mathbb E_{z \sim N(0,1)}[\sigma(z)]^2+\mathcal O(1/d) = \beta_0(\sigma) + \mathcal O(1/d)$, by definition of $\beta_1(\sigma)$. One can use analogous arguments to obtain $\phi_\sigma'(0) = \beta_1(\sigma)+\mathcal O(1/d)$.

If $t=1$, then $w_j^\top w_\ell = 1$, and so $w_j = w_\ell$. Thus, one computes
\begin{eqnarray*}
\begin{split}
\phi_\sigma(1) &= d\cdot \mathbb E_{x\sim \tau_d}[\sigma(x^\top w_j)\sigma(x^\top w_j)] = d\cdot \mathbb E_{x \sim \tau_d}[\sigma(\sqrt{d}x^\top w_j)^2]\\
&= \mathbb E_z[\sigma(z)^2] + \mathcal O(1/d) =: \beta_1(\sigma)+\mathcal O(1/d),
\end{split}
\end{eqnarray*}
which completes the proof.
\end{proof}

The following lemma which is a direct consequence of a result of ~\cite{justinterpolate} (see also previous work in \cite{elkaroui2010}), establishes that $\widetilde{C}_\sigma(W)^{\mathrm{lin}}$ is a linearization of $\widetilde{C}_\sigma(W)$, which keeps the main spectral information of the former.
\begin{restatable}[Linear approximation of $\widetilde{C}_\sigma(W)$]{lm}{}
For sufficiently large $d$, it holds w.p $1-d^{-\Omega(1)}$ over the choice of $W$ that
$\|\widetilde{C}_\sigma(W)-\widetilde{C}_\sigma(W)^{\mathrm{lin}}\|_{op} = o(1)$.
\label{lm:specC}
\end{restatable}
\begin{proof}
The proof is based on ~\cite[Proposition A.2]{justinterpolate} which is itself a non-asymptotic / quantitative version of ~\cite[Theorem 2.1]{elkaroui2010}.
One may write $w_j = \Sigma_d^{-1/2}z_j$ where $\Sigma_d=I_d$ and $z_j$ is uniformly distributed on the sphere of radius $\sqrt{d}$ in $\mathbb R^d$, as thus is $1$-subGaussian.
thanks to Lemma \ref{lm:homrep}. In \cite[Proposition A.2]{justinterpolate}, noting that $\trace(\Sigma_d) = \trace(\Sigma_d^2)$, and taking $m=\infty$ (i.e $\theta=1/2$) (since the $z_j$'s are $1$-subGaussian isotropic random vectors), we deduce that for $\delta$ sufficiently small and $d$ sufficiently large, it holds w.p $1-\delta-d^{-2}$ that
$
\|\widetilde{C}_\sigma(W)-\widetilde{C}_\sigma(W)^{\mathrm{lin}}\|_{op} \le d^{-1/2}(\delta^{-1/2}+\log^{0.51} d).
$
It then suffices to take $\delta=d^{-c}$ for any $0 < c < 1$ to complete the proof.
\end{proof}

\subsection{Proof of Theorem \ref{thm:finiteRFLaw} (Law of robustness in RF regime with finite width)}
\label{subsec:rf}

For the proof of the theorem, we shall need a specialized corollary to Lemma \ref{lm:specC} to give probabilistic estimates for the extreme eigenvalues of $C_\sigma(W) \in \mathbb R^{k \times k}$, the covariance matrix of $\sqrt{d}\sigma(Wx)$ for $x \sim \tau_d$.
Recall the definition of the curvature coefficients $\beta_0(\sigma)$, $\beta_1(\sigma)$, and $\beta_\star(\sigma)$ from \eqref{eq:betas}.

\begin{restatable}[Extreme eigenvalues of $C_\sigma(W)$]{cor}{}
If Condition \ref{cond:relulike} holds, then for sufficiently large $d$ and $k$ with $k \asymp d$, it holds $1-d^{-\Omega(1)}$ over $W$ that
\begin{eqnarray}
c \le \lambda_{\min}(C_\sigma(W)) \le \lambda_{\max}(C_\sigma(W)) \le C.
\end{eqnarray}
where $c,C>0$ are constants which only depend on the ratio $k/d$ and the activation function $\sigma$.
\label{cor:Cspecrelulike}
\end{restatable}
\begin{proof}
Using Lemma \ref{lm:specC} and the fact that $C_\sigma(W) = \widetilde{C}_\sigma(W) - \phi_\sigma(0)1_k1_k^\top$, we have
$\|C_\sigma(W)-(\beta_1(\sigma)WW^\top + \beta_\star(\sigma)I_d)\|_{op} = o(1)$ w.p $1-d^{-\Omega(1)}$. On the other hand,  standard RMT ~\cite{rmt} guarantees the existence of universal constants $c',C'>0$ such that $c' \le \lambda_{\min}(WW^\top) \le \lambda_{\max}(WW^\top) \le C'$ w.p $1-e^{-\Omega(d)}$. The result then follows upon taking into account Condition \ref{cond:relulike}.
\end{proof}

We are now ready to establish a law of robustness for finite-width neural two-layer neural networks in the random features regime.
We restate the theorem for convenience. As before, let $\varepsilon^\star_{\mathrm{test}}$ be the Bayes-optimal error for the problem and let $\varepsilon$ be any error threshold in the interval $[0,\varepsilon^\star_{\mathrm{test}})$.
\finiteRFLaw*
\begin{proof}
From Corollary \ref{cor:Cspecrelulike}, we know that $\lambda_{\min}(C_\sigma(W)) = \Omega(1)$ w.p $1-d^{-\Omega(1)}$. The result then follows directly upon combining with Theorem \ref{thm:frozenlaw} and the fact that $\|W\|_F = \sqrt{k}$ because the rows of $W$ are on the unit-sphere $\sphere$.
\end{proof}

\subsection{Proof of Theorem \ref{thm:law} (tightness of lower-bound in Theorem \ref{thm:finiteRFLaw})}
We shall now prove that the $\sqrt{n}$ lower-bound in Theorem \ref{thm:law} is tight: it is achieved by the min-norm interpolator.
\law*


We will make use of the following result from ~\cite{Mei2019}.
\begin{restatable}[Theorem 6 of \cite{Mei2019}, specialized to the case of positive-homogeneous activation functions]{prop}{}
\label{prop:montanari}
Assume Condition \ref{cond:purelynonlin}. In the limit when $n,d,k \to \infty$ in the sense of \eqref{eq:proportionate}, the following hold.
\begin{itemize}
    \item {\bfseries Memorization.} There is a constant $L(\gamma_1,\gamma_2,\theta^2,\lambda) \ge 0$ which is increasing in $\lambda$ with $L(\gamma_1,\gamma_2,\theta^2, 0) = 0$,  such that $\mathbb E_{X,W}|\mathrm{MSE}(\widehat{f}_{\rf,\lambda}) - L(\gamma_1,\gamma_2,\theta^2,\lambda)| = o(1)$.
    \item {\bfseries Norm of min-norm interpolator.}
    There is a constant $A(\gamma_1,\gamma_2,\theta^2,\lambda) \in [0,\infty]$ satisfying
    \begin{itemize}
        \item $A(\gamma_1,\gamma_2,\theta^2,\lambda)$ is decreasing in $\lambda$,
        \item $A(\gamma_1,\gamma_2,\theta^2, 0)$ is finite and increasing in $\gamma := \gamma_2/\gamma_1$, for $\gamma \in (0, 1)$,
        \item $\lim_{\gamma \to 1}A(\gamma_1,\gamma_2,\theta^2,0) = \infty$,
        \item $A(\gamma_1,\gamma_2,\theta^2,0)$ is finite and increasing in $\gamma$, for $\gamma \in (1,\infty)$,
        \end{itemize}
        such that $\mathbb E_{X,W}|(\beta_\star^2/d)\|\widehat{v}_{\rf,\lambda}\|^2 - A(\gamma_1,\gamma_2,\theta^2,\lambda)| = o(1)$.
    \end{itemize}
\end{restatable}
\begin{restatable}{rmk}{}
The following remarks are in place.
\begin{itemize}
\item We have restated the result of \cite{Mei2019} for our purposes. In particular, the authors proved a stronger statement in which the labels are not entirely independent of the data (i.e positive SNR). The version stated above corresponds to noise-only regime where the SNR is zero.
\item The $1/d$ factor in $\|\widehat{v}_{\rf,\lambda}\|^2$ in the above proposition accommodates for the fact that we work on the unit-sphere $\sphere$ while the results of \cite{Mei2019} were stated for $\sqrt{d}\sphere$, 
The above version of their result is then obtained via a simple change of activation function $\widetilde{\sigma}(t) := \sigma(t)/\sqrt{d}$ by $1$-homogeneity of $\sigma$, from where we obtain the relations $\beta_\star^2(\widetilde{\sigma}=(\beta_\star^2(\sigma)/d)$ and $\theta^2(\widetilde{\sigma}) = \theta^2(\sigma)$.
    \item It was also observed (empirically) in \cite{Mei2019} that when $\gamma \to \infty$, $A(\gamma_1,\gamma_2,\theta^2,\lambda)$ converges to a positive finite constant which does not depend on any of $\gamma_1$, $\gamma_1$, $\beta_\star^2$, or $\theta^2$.
\end{itemize}
\end{restatable}

\begin{proof}[Proof of Theorem \ref{thm:law}]
The memorization part of the theorem is a direct consequence of Proposition \ref{prop:montanari}. Still by Proposition \ref{prop:montanari}, we know that $(\beta_\star^2/d)\|\widehat{v}_{\rf,\lambda}\|^2 = A(\gamma_1,\gamma_2,\theta^2,\lambda) + o_{\mathbb P}(1)$ for a constant $A(\gamma_1,\gamma_2,\theta^2,\lambda)$ satisfying all the properties in the proposition. Since, $n$ is proportional to $d$ and $\beta_\star^2 > 0$ by hypothesis, we conclude upon invoking Theorem \ref{thm:equiv}, that
\begin{eqnarray}
\mathfrak{S}(\widehat{f}_{\rf,\lambda}) \asymp \|\widehat{v}_{\rf,\lambda}\| = \begin{cases}
\omega_{\mathbb P}(\sqrt{d})=\omega_{\mathbb P}(\sqrt{n}),&\mbox{ if }(\gamma,\lambda)=(1,0),\\
\Theta_{\mathbb P}(\sqrt{d})=\Omega_{\mathbb P}(\sqrt{n}),&\mbox{ else,}
\end{cases}
\label{eq:vnormbound}
\end{eqnarray}
which concludes the proof.
\end{proof}

\section{Finite-width NTK regime}
\subsection{Proof of Theorem \ref{thm:finiteNTKLaw}}
\finiteNTKLaw*
We start with an auxiliary lemma that will be crucial for the proof of the Theorem.
\begin{restatable}{lm}{}
\label{lm:kron}
For $x \sim \tau_d$, the covariance matrix of $\sqrt{d}\Phi_\ntk(x)$ is given by
$$
C_{\Phi_{\ntk}} = \frac{1}{k}(\widetilde{C}_{\sigma'}(W) \otimes I_d) \in \mathbb R^{kd \times kd},
$$
where $\widetilde{C}_{\sigma'}(W)$ is the $k \times k$ psd matrix with entries given by $\widetilde{C}_{\sigma'}(W)_{j,\ell} = \mathbb E_{x \sim \tau_d}[\sigma'(x^\top w_j)\sigma'(x^\top w_\ell)]$.
\label{lm:covkron}
\end{restatable}
\begin{proof}
Let $x \sim \tau_d$ and $z(x) := (1/\sqrt{k})\sigma'(Wx) := ((1/\sqrt{k})\sigma'(x^\top w_1),\ldots,(1/\sqrt{k})\sigma'(x^\top w_k)) \in \mathbb R^k$, and observe $\Phi_\ntk(x) = z(x) \otimes x \in \mathbb R^{kd}$, the Kronecker product of $z(x)$ and $x$. On the other hand, it is clear that $z(x)$ and $x$ are independent\footnote{Because $Wx$ and $x$ are independent, since $W$ and $x$ are.}. Thanks to Lemma \ref{lm:covkron}, we then obtain
$$
C_{\Phi_\ntk} = d\cdot \cov(z(x) \otimes x) =  \mathbb E[z(x)z(x)^\top ]) \otimes \cov(\sqrt{d}x) = \frac{1}{k}\widetilde{C}_{\sigma'}(W) \otimes I_d,
$$
as claimed.
\end{proof}

Note that under Condition \ref{cond:hom}, $\sigma'$ is positive-homogeneous of order $0$, and thus by Proposition \ref{prop:catalan}, there exists a continuous function $\phi_{\sigma'}:[-1,1] \to \mathbb R$ such that
\begin{eqnarray}
\mathbb E_{x \sim \tau_d}[\sigma'(x^\top u)(\sigma'(x^\top v)] = \phi_{\sigma'}(u^\top v).
\label{eq:homder}
\end{eqnarray}
The following Lemma can be easily proved by differentiating through formua \ref{eq:catalan}.
\begin{restatable}{lm}{}
We have the functional identity: $\phi_{\sigma'} = (\phi_\sigma)'$.
\end{restatable}

For example, if $\sigma$ is the ReLU activation function, then
\begin{eqnarray}
\phi_{\relu'}(t) = (\phi_{\relu})'(t) = \frac{\arccos(-t)}{2\pi},
\end{eqnarray}
by differentiating equation \eqref{eq:reluphi}.

\begin{proof}[Proof of Theorem \ref{thm:finiteNTKLaw}]
One may upper-bound the energy of $\Phi_\ntk$ like so
\begin{eqnarray}
\begin{split}
\|\Phi_\ntk\|_{L^2(\tau_d)}^2 = \frac{1}{k}\mathbb E_{x \sim \tau_d}[\|\sigma'(Wx) \otimes x\|^2] &\le \frac{1}{k}\mathbb E[\|\sigma'(Wx)\|^2\|x\|^2]\\
&= \frac{1}{k}\mathbb E_x[\|\sigma'(Wx)\|^2] \le 1,
\end{split}
\label{eq:Phiboundntk}
\end{eqnarray}
where the last step is because $\sigma$ is $1$-Lipschitz. Combining with Corollary \ref{cor:Cspecrelulike} and Lemma \ref{lm:kron}, gives $$
\alpha_{\Phi_\ntk} := \frac{\|\Phi_\ntk\|^2_{L^2(\tau_d)}}{\lambda_{\min}(C_{\Phi_\ntk})} \le  \dfrac{\mathcal O(1)}{\Omega(1/k)} \le \mathcal O(k),
$$
w.p $1-d^{-\Omega(1)}$ over the random matrix $W$.
The result then follows from Theorem \ref{thm:dico}.
\end{proof}

\section{Misc: Arbitrary / nonhomogeneous activation functions}
\label{sec:arbitrary}
We now drop the homogeneity assumption on the activation function $\sigma$. In this scenario, we cannot carry out computations as in section \ref{sec:sob}.
Given a neural network $f=f_{W,v} \in \mathcal F_{d,k}(\sigma)$, analysing the Lipschitz constant $\Lip_{\sphere}(f)$ of a function, or even the lower-bound $\mathfrak{S}(f)$ thereof, is difficult as the parameters $W$ and $v$ enter the definition of $f$ in a rather complex manner (due to the nonlinearity $\sigma$). Fortunately, the \emph{Poincar\'e inequality} is there for the rescue: we can bound the later quantity via the variance of $f$, which leads to quadratic-form in $v$ by means of a kernel matrix generated by $W$ and $\sigma$. As we shall see, this will lead to the emergence of another kernel matrix $C'_\Phi(W):=\cov_{x \sim \tau_d}(\sqrt{d}Wx))$ which will take over the role of $C_\Phi(W)$ introduced in \eqref{eq:analytic}.


\subsection{Poincar\'e inequality on the sphere and the emergence of another kernel matrix}
\label{sec:emergence}
Recall that, for uniform-distribution $\tau_d$ on the unit-sphere $\sphere$ (assumed in the definition of generic data), the Poincar\'e inequality tells us that, for any continuously-differentiable function $f:\mathbb R^d \to \mathbb R$,
\begin{eqnarray}
\mathfrak{S}(f)^2 \ge c\cdot (d-1)\var_{\tau_d}(f) \gtrsim d \cdot \var_{\tau_d}(f),
\label{eq:poincare}
\end{eqnarray}
where $c>0$ is an absolute constant (with a concrete value like $1$ or $2$, independent of the dimension $d$ and the test function $f$), and $\var_{\tau_d}(f):=\|f-\mathbb E_{\tau_d}[f]\|_{L^2(\tau_d)}^2$ is the variance of $f$, with $\mathbb E_{\tau_d}[f] = \mathbb E_{x \sim \tau_d}[f(x)] := \int_{\sphere}f(x)\dif\tau_d(x)$ being the average value of $f$ w.r.t the measure $\tau_d$. The factor $d-1$ in \eqref{eq:poincare} is optimal; it is the (optimal) \emph{Poincar\'e constant} for the uniform distribution $\tau_d$ on the unit-sphere $\sphere$. We refer the reader to standard monographs on the subject, like ~\cite{ledoux,boucheron2013,Gozlan2015}.


Let $\Phi:\sphere \to \mathbb R^m$ be a measurable function $a \in \mathbb R^m$, and consider a general linear model $f:\sphere \to \mathbb R$ given by
\begin{eqnarray}
f(x) = a^\top \Phi(x).
\label{eq:featurized}
\end{eqnarray}
The vector $\Phi(x) \in \mathbb R^m$ are the features of the example $x$. Note that we allow for cases where the feature mapping $\Phi:\sphere \to \mathbb R$ is learnable. This subsumes feed-forward linear neural networks, and in particular, the class $\mathcal F_{d,k}(\sigma)$ of two layer neural networks $f:x \mapsto v^\top\sigma(Wx)$ with activation function $\sigma$, by taking $a = v$ and $\Phi(x)=\sigma(Wx)$. Least squares estimators in general RKHSs are also an instance of \eqref{eq:featurized}.
Let $\mu_\Phi$ and $C_\Phi$ be the mean and the covariance (resp.) w.r.t $x \sim \tau_d$ of the feature vector $\sqrt{d}\Phi(x)$, i.e
\begin{eqnarray}
\label{eq:C}
\begin{split}
\mu_\Phi &:= \mathbb E_x[\Phi(x)] \in \mathbb R^m,\text{ and }\\
C_\Phi &:= \mbox{cov}_x(\sqrt{d}\Phi(x)):= \mathbb E_x[\sqrt{d}\Phi(x)\sqrt{d}\Phi(x)^\top] - \mu_\Phi\mu_\Phi^\top \in \mathbb R^{m \times m}.
\end{split}
\end{eqnarray}
\begin{restatable}{thm}{}
\label{thm:cov}
For any function $f:\sphere \to \mathbb R$ of the form \eqref{eq:featurized}, we have the lower-bound
\begin{eqnarray}
\label{eq:varbound}
\mathfrak{S}(f)^2 \ge a^\top {C_\Phi} a \ge \|a\|^2\lambda_{\min}(C_\Phi).
\end{eqnarray}
\end{restatable}
This result will be heavily used in subsequent sections to analyze robustness analysis of random features and NTK regimes induced by general / non-homoegeneous activation functions.
\begin{proof}[Proof of Theorem \ref{thm:cov}]
Using standard formulae for expectations of quadratic forms, one computes
\begin{eqnarray*}
\begin{split}
d\cdot \mathbb E_x[f(x)^2] &= \mathbb E_x[(a^\top\sqrt{d}\Phi(x))^2] = \mathbb E_x[\sqrt{d}\Phi(x)^\top aa^\top\sqrt{d}\Phi(x)] = a^\top C_\Phi a + (a^\top\mu_\Phi)^2\\
&= a^\top C_\Phi a + d\cdot (\mathbb E_x[f(x)])^2 ) = a^\top C_\Phi a + d\cdot (\mathbb E_x[f(x)])^2,
\end{split}
\end{eqnarray*}
Thus, the variance of $f$ w.r.t to $x \sim \tau_d$ is given by the following quadratic form in $V$
\begin{eqnarray}
d\cdot \var_{x}(f(x)) := d\cdot \mathbb E_x[f(x)^2] - d\cdot (\mathbb E_x[f(x)])^2 = a^\top C_\Phi a.
\label{eq:var}
\end{eqnarray}
Combining with the Poincar\'e inequality \eqref{eq:poincare}, this proves the following template result linking the Lipschitz constant of $f$ with the $L_2$-norm w.r.t the covariance matrix feature $C_\Phi$, of the parameter vector $a \in \mathbb R^{m}$.
\end{proof}

\subsection{Spectral analysis of $C_\Phi$, for embeddings of the form $\Phi(x)=\varphi(Vx)$}
\label{subsec:genCPhi}
Suppose the embedding function $\Phi:\sphere \to \mathbb R^m$ is of the form
\begin{eqnarray}
\Phi(x)=\varphi(Vx):=(\varphi(x^\top v_1),\ldots,(\varphi(x^\top v_m)),
\label{eq:Vmodel}
\end{eqnarray}
for some continuous scalar function $\varphi:[-1,1] \to \mathbb R$ and $m \times d$ matrix $V$ with rows $v_1,\ldots,v_m \in \mathbb R^d$. This is the case of exact two-layer neural networks where $\varphi=\sigma$ (the activation function), $m=k$ (the number of hidden neurons), and $V=W \in \mathbb R^{k \times d}$ (the hidden weights matrix).

In view of applying Theorem \ref{thm:cov} to get lower-bounds on the nonrobustness of the model $f:\sphere \to \mathbb R$, $x\mapsto a^\top \Phi(x) = a^\top \varphi(Vx)$, one must lower-bound the smallest eigenvalue of $C_{\Phi}$, the covariance matrix of the random vector $\sqrt{d}\Phi(x) \in \mathbb R^m$, for $x \sim \tau_d$. This is the purpose of the next theorem.
\begin{restatable}[Lower-bound on $\lambda_{\min}(C_\Phi)$]{thm}{}
Suppose $\varphi$ is thrice continuously-differentiable at zero, with Maclaurin expansion $\varphi(t)=a_0+a_1t + a_2t^2+a_3t^3 + \mathcal O(t^4)$. Then, we have
\begin{eqnarray}
\begin{split}
(C_\Phi)_{j,\ell} &= (\overline{C}_\Phi)_{j,\ell}-\frac{a_2^2}{d} +\mathcal O(\frac{1}{d^2}),
\end{split}
\end{eqnarray}
where  $\overline{C}_\Phi := c_dVV^\top + \widetilde{c}_dVV^\top \circ VV^\top \in \mathbb R^{m \times m}$, and $c_d$ and $\widetilde{c}_d$ are defined by
\begin{eqnarray}
 \begin{split}
 c_d &:= a_1^2 + \frac{6a_1a_3}{d},\;\widetilde{c}_d &:= \frac{2a_2^2}{d}.
 \end{split}
 \end{eqnarray}
\label{thm:lambdamingeneralC}
\end{restatable}
For the proof of Theorem \ref{thm:lambdamingeneralC}, we will need the following lemma.
\begin{restatable}[Correlation functions of coordinates of uniform random vector on sphere]{lm}{prodexp}
Suppose $\varphi$ is thrice continuously-differentiable at zero. If $\varphi(t)=a_0 + a_1 t + a_2 t^2 + a_3 t^3 + \mathcal O(t^4)$ is its Maclaurin expansion, then for every $u,v \in \sphere$, and $x \sim \tau_d$, we have the approximation
\begin{eqnarray}
\begin{split}
\mathbb E_x[\varphi(x^\top u)\varphi(x^\top v)]-\mathbb E_x[\varphi(x^\top u)]\mathbb E_x[\varphi(x^\top v)] &= -\frac{a_2^2}{d^2}+(\dfrac{a_1^2}{d}+\frac{6a_1a_3}{d^2})u^\top v + \dfrac{2a_1^2}{d^2}(u^\top v)^2\\
&\quad\quad\quad\quad\quad\quad\quad\quad\quad\quad\quad\quad\, +\mathcal O(\dfrac{1}{d^3}).
\end{split}
\end{eqnarray}
In particular, if $u$ and $v$ are perpendicular, then
\begin{eqnarray}
\mathbb E_x[\varphi(x^\top u)\varphi(x^\top v)]-\mathbb E_x[\varphi(x^\top u)]\mathbb E_x[\varphi(x^\top v)]=-\frac{a_2^2}{d^2}+\mathcal O(\frac{1}{d^3}).
\end{eqnarray}
\label{lm:prodexp}
\end{restatable}
The proof of the lemma is given in Appendix \ref{sec:techproof}.

\begin{restatable}[Eigenvalues of perturbed matrix]{lm}{}
If $A$ and $E$ are $N \times N$ matrices with $|e_{i,j}| \le \varepsilon$ for all $i,j \in [N]$, then
$$
\sup_{1 \le i \le N}|\tau_i(A+E)-\tau_i(A)| \le N\varepsilon,
$$
where $\tau_1(A) \ge \tau_2(A) \ge \ldots \ge \tau_N(A)$ are the singular-values of $A$ (and similarly for $A+E$).
\label{lm:eigenpert}
\end{restatable}
\begin{proof}
Its is well-known that $\sup_{1 \le i \le N}|\tau_i(A+E)-\tau_i(A)| \le \|E\|$. It then suffices to observe that $\|E\|_{op} \le \|E\|_{F} \le \sqrt{N^2 \varepsilon^2} = N\varepsilon$.
\end{proof}
\begin{proof}[Proof of Theorem \ref{thm:lambdamingeneralC}]
From Lemma \ref{lm:prodexp}, we know that
\begin{eqnarray*}
\begin{split}
(C_{\Phi})_{j,\ell} &= d\cdot(\mathbb E_x[\varphi(x^\top v_j)\varphi(x^\top v_\ell)]-\mathbb E_x[\varphi(x^\top v_j)\mathbb E_x[\varphi(x^\top v_\ell)]) \\
&=d\cdot(-\frac{a_2^2}{d^2} + \dfrac{c_d}{d}v_j^\top v_\ell+ \dfrac{\widetilde{c}_d}{d}(v_i^\top v_\ell)^2 + \mathcal O(\frac{1}{d^3})) = -\frac{a_2^2}{d} +(\overline{C}_\Phi)_{j,\ell}+E_{j,\ell},
\end{split}
\end{eqnarray*}
where $E:=C_\Phi-\overline{C}_\Phi-(a_2^2/d)1_m1_m^\top$ with $\|E\|_{op} = d \cdot \mathcal O(1/d^2)=\mathcal O(1/d)$, thanks to the above display and Lemma \ref{lm:eigenpert}. One then derives that
\begin{eqnarray*}
\begin{split}
\lambda_{\min}(C_{\Phi}) &= \lambda_{\min}(\overline{C}_\Phi+E-\frac{a_2^2}{d}1_m1_m^\top) \ge \lambda_{\min}(\overline{C}_\Phi-a_2^2 I_m)-\|E\|_{op}\\
&\ge \lambda_{\min}(\overline{C}_\Phi-a_2^2I_m)-\mathcal O(\frac{1}{d})\ge \lambda_{\min}(a_1^2VV^\top - a_2^2 I_m)- \mathcal O(\frac{1}{d}),
\end{split}
\end{eqnarray*}
where the first inequality  is thanks to \emph{Cauchy-Weyl interlacing inequality} to compare the eigenvalues of psd matrices $C_\Phi+(a_2^2/d)1_m1_m^\top$ and $C_\Phi=\overline{C}_\Phi+E$, the third is by definition of $\|E\|_{op}$, and the last inequality uses the fact that  $VV^\top \circ VV^\top$ is psd (thanks to the \emph{Shur product theorem}). This proves part (A) of the theorem.
Part (B) is a direct consequence of Theorem \ref{thm:cov}.
\end{proof}
The following Corollary to Theorem \ref{thm:lambdamingeneralC} will be crucial for our analysis of finite-width RF models and infinite-width RF / NTK models.

\begin{restatable}[Theorem \ref{thm:randVrestated} restated]{cor}{}
Suppose $\varphi$ is thrice continuously-differentiable at zero.
Suppose the rows of $V$ are drawn iid from an isotropic $1$-subGaussian distribution in $\mathbb R^d$. If $m,d \to \infty$ such that $m/d \to  \gamma_1 \in [0,1)$, then $\lambda_{\min}(C_\Phi) \to \varphi'(0)^2(1-\sqrt{\gamma_1})^2$ almost-surely.
\label{cor:randV}
\end{restatable}
\begin{proof}
Thanks to Bai-Yin, we know that $\lambda_{\min}(VV^\top) \to (1-\sqrt{\gamma_1})^2$ a.s. Invoking Theorem \ref{thm:lambdamingeneralC} and the fact that finite-rank perturbations do not affect the limiting spectral distribution of random matrices, we deduce that $\lambda_{\min}(C_\Phi) \to \varphi'(0)^2(1-\sqrt{\gamma_1})^2$ a.s as claimed.
\end{proof}



\label{sec:sob}
\subsection{An analytic formula for $\mathfrak{S}(f_{W,v})$}

Suppose the activation function $\sigma$ is positively-homogeneous of order 1. As an example, the reader may think of the ReLU or the absolute-value activation function. For any two-layer neural network $f=f_{W,v} \in \mathcal F_{d,k}(\sigma)$, may compute the squared Sobolev-seminorm of $f$ as follows
\begin{eqnarray}
\begin{split}
\mathfrak{S}(f)^2 := \mathbb E_x \|\nabla_{\sphere} f(x)\|^2 &= \mathbb E_x\|\nabla f(x)\|^2 - \mathbb E_x (x^\top \nabla f(x))^2\\
 &= \mathbb E_x \|\nabla f(x)\|^2 - \mathbb E_x [|f(x)|^2],
\end{split}
\end{eqnarray}
where we have used the order-$1$ positive-homogeneity of the activation function $\sigma$ in the last step (\emph{Euler's Theorem}). We now compute each term of the rightmost side separately.

\emph{The first term.} Now, $f(x) := v^\top\sigma(Wx) = \sum_{j=1}^k v_j\sigma(x^\top w_j)$ and so
\begin{eqnarray*}
\begin{split}
\mathbb E_x \|\nabla f(x)\|^2 &= \mathbb E_x \|\sum_{j=1}^kv_j\sigma'(x^\top w_j)w_j\|^2 = \mathbb E_x\sum_{j=1}^k\sum_{\ell=1}^kv_jv_\ell\sigma'(x^\top w_j)\sigma'(x^\top w_\ell)w_j^\top w_\ell\\
&=\sum_{j=1}^k\sum_{\ell=1}^kv_jv_\ell \mathbb E_x[\sigma'(x^\top w_j)\sigma'(x^\top w_\ell)]w_j^\top w_\ell = \|v\|^2_{ \widetilde{C}_{\sigma'}(W,W) \circ WW^\top},
\end{split}
\end{eqnarray*}
where $\widetilde{C}_{\sigma'}(W)$ is the $k \times k$ matrix with entries given by
\begin{eqnarray} (\widetilde{C}_{\sigma'}(W))_{j,\ell} := \mathbb E_{x \sim \tau_d}[\sigma'(x^\top w_j)\sigma'(x^\top w_\ell)] = \phi_{\sigma'}(w_j^\top w_\ell)=\phi_\sigma'(w_j^\top w_\ell),
\end{eqnarray}
where $\phi_\sigma:[-1,1] \to \mathbb R$ is the continuous function whose existence is guaranteed by Proposition \ref{prop:catalan}.

\emph{The second term.} One computes $|f(x)|^2 = \sum_{j=1}^k \sum_{\ell=1}^k v_jv_\ell \sigma(x^\top w_j)\sigma(x^\top w_\ell)$, and so
$$
\mathbb E_x [f(x)^2] = \sum_{\ell=1}^k v_jv_\ell\mathbb E_x[\sigma(x^\top w_j)\sigma(x^\top w_\ell)] = \|v\|^2_{\widetilde{C}_{\sigma}(W)},
$$
where $\widetilde{C}_\sigma(W)$ is the $k \times k$ psd matrix with entries given by 
\begin{eqnarray}
(\widetilde{C}_\sigma(W))_{j,\ell} := \mathbb E_{x \sim \tau_d}[\sigma(x^\top w_j)\sigma(x^\top w_\ell)].
\end{eqnarray}

Let $G_\sigma(W)$ be the $k \times k$ psd matrix with entries given for all $j,\ell \in [k]$ by
\begin{eqnarray}
\label{eq:analytic}
(G_\sigma(W))_{j,\ell} :=  (WW^\top \circ \widetilde{C}_{\sigma'}(W,W))_{j,\ell}-(\widetilde{C}_{\sigma'}(W,W))_{j,\ell}=\widetilde{\phi}_\sigma(w_j^\top w_j),
\end{eqnarray}
where $\widetilde{\phi}_\sigma:[-1,1] \to \mathbb R$ is defined by $\widetilde{\phi}_\sigma(t) := t\phi_\sigma'(t)-\phi_\sigma(t)$.
Putting things together, we obtain the following result which gives an analytic formula for the $\mathfrak{S}(f_{W,v})$ as a quadratic form in $v$, with coefficient matrix $G_\sigma(W)$. Thanks to \cite[Table 1]{louart2018}, we obtain Table \ref{tab:maclaurin} below which summaries the Maclaurin expansion of $\widetilde{\phi}_\sigma$ for a certain number of common activation functions.
\begin{table}[h!]
  \begin{center}
       {\tabulinesep=1.2mm
    \begin{tabu}{|c|c|c|}
    \hline
       $\sigma$ & $\widetilde{\phi}_\sigma(t):=t\phi_{\sigma'}(t)-\phi_\sigma(t)$ & Maclaurin expansion of $\widetilde{\phi}_\sigma$ \\
      \hline
        ReLU & $\dfrac{t\arccos(-t)}{2\pi}-\dfrac{t\arccos(-t)+\sqrt{1-t^2}}{2\pi d}$ &
        $\dfrac{1}{4}t+\dfrac{1}{2\pi}t^2+\mathcal O(t^3,\dfrac{1}{d})$\\
      \hline
        abs & $\dfrac{2t\arcsin(t)}{\pi}-\dfrac{2t\arcsin(t)+2\sqrt{1-t^2}}{\pi d}$ & $\dfrac{2}{\pi}t^2+\mathcal O(t^3,\dfrac{1}{d})$\\
      \hline
    erf & $\dfrac{4t}{\pi \sqrt{9-4t^2}}-\dfrac{2}{\pi d}\arcsin(2t/3)$ & $\dfrac{4}{3\pi}t+\mathcal O(t^3,\dfrac{1}{d})$\\
      \hline
    \end{tabu}
    }
  \end{center}
    \caption{Table of the function $\widetilde{\phi}_\sigma$ defined in \eqref{eq:analytic}, for the ReLU and absolute-value activation functions.}
\label{tab:maclaurin}
\end{table}
\begin{restatable}[Analytic formula for Sobolev norm of two-layer neural network]{thm}{}
For any $f=f_{W,v} \in \mathcal F_{d,k}(\sigma)$, we have the identity
$\mathfrak{S}(f) = v^\top G_\sigma(W)v$.
\label{thm:analytic}
\end{restatable}

Thanks to the definition of extreme singular-values of matrices, we know that
\begin{eqnarray}
 \label{eq:sandwitch}
 \lambda_{\min}(G_\sigma(W))\|v\|^2 \le v^\top G_\sigma(W) v \le \lambda_{\max}(G_\sigma(W))\|v\|^2.
\end{eqnarray}
 Thus, in virtue of Theorem \ref{thm:analytic}, to get lower- and upper-bounds for $\mathfrak{S}(f_{W,v})$, it suffices to
 \begin{itemize}
 \item control the extreme singular-values of $G_\sigma(W)$, and
 \item control the $L_2$-norm of the output weights vector $\|v\|$.
 \end{itemize}

\begin{restatable}{thm}{}
Let the activation function $\sigma$ be the ReLU and let the rows of the hidden weights matrix $W \in \mathbb R^{k \times d}$ be drawn iid from $\tau_d$. For sufficiently large $k \asymp d$, it holds w.p $1-d^{-\Omega(1)}$ that
\begin{eqnarray}
\|G_\sigma(W)  - \frac{1}{4}(WW^\top + I_k)\|_{op} = o(1).
\end{eqnarray}
In particular, there exist constants $C\ge c \ge 1/4$ (only depending on the ratio $k/d$) such that
$c \le \lambda_{\min}(G_\sigma(W)) \le \lambda_{\max}(G_\sigma(W)) \le C$ w.p $1-d^{-\Omega(1)}$
\label{thm:specReLU}
\end{restatable}
\begin{proof}
The first part is completely analogous to the proof of Lemma \ref{lm:specC}, with $\phi_\sigma$ replaced with $\widetilde{\phi}_\sigma$. We also make use of Table \ref{tab:maclaurin} for the computations for extracting the Maclaurin coefficients of $\widetilde{\phi}_\sigma$. The second part follows from standard RMT ~\cite{rmt}.
\end{proof}

\begin{restatable}{thm}{}
Let the activation function  $\sigma$ be the ReLU, absolute-value, gaussian rf, or tanh, and let the rows of the hidden weights matrix $W \in \mathbb R^{k \times d}$ be drawn iid from $\tau_d$. Then, there exist constants $c,C>0$ (only depending on the ratio $k/d$) such that $c \le \lambda_{\min}(G_\sigma(W)) \le \lambda_{\max}(G_\sigma(W)) \le C$ w.p $1-d^{-\Omega(1)}$.
\label{thm:specReLULike}
\end{restatable}

\begin{restatable}{thm}{equiv}
Let $\sigma$ and $W$ be as in Theorem \ref{thm:specReLULike}. Then, ffor sufficiently large $k \asymp d$ it holds w.p $1-d^{-\Omega(1)}$ over $W$ that $\mathfrak{S}(f_{W,v}) \asymp \|v\|$, for every $f_{W,v} \in \mathcal F_W(\sigma)$.
\label{thm:equiv}
\end{restatable}
Thus, any lower / upper-bound on the output weights of a neural network in $\mathcal F_W$ immediate translate to a comparable lower / upper-bound on $\mathfrak{S}(f)$. 

\begin{proof}
Proof follows from combining Theorems \ref{thm:analytic} and \ref{thm:specReLULike}.
\end{proof}

\section{Technical proofs}
\label{sec:techproof}

\subsection{Proof of Lemma \ref{lm:prodexp}}
We will need the following auxiliary lemma proved further below.
\begin{restatable}{lm}{pq}
Let $u$ and $v$ be fixed and $x$ be uniformly random on the unit-sphere $\sphere$. Let $p$ and $q$ be nonnegative integers and define $c_{p,q}(u,v):=\mathbb E_x[(x^\top u)^p(x^\top v)^q]$. If $p$ and $q$ have different parities, then $c_{p,q}(u,v)=0$. Otherwise, we have the formula
\begin{eqnarray}
c_{p,q}(u,v) = \dfrac{p!q!\Gamma(\frac{d}{2})}{2^{p+q}\Gamma(\frac{d+p+q}{2})} \sum_t \dfrac{2^t}{t!(\frac{p-t}{2})!(\frac{q-t}{2})!}(u^\top v)^t,
\end{eqnarray}
where the sum is over all $t$ between $0$ and $p \land q$ inclusive, that have the same parity as $p$ and $q$. The formula is simplified in the table below for special values of $p$ and $q$.
\label{lm:pq}
\end{restatable}

\begin{table}[h!]
    \label{Tab:tablepq}
  \begin{center}
       {\tabulinesep=1.2mm
    \begin{tabu}{|c|c|c|c|}
    \hline
      $p$ & $q$ & $c_{p,q}(u,v)$ & Comment \\
      \hline
      $2m$ & 0 & $\dfrac{C_m}{d^{m}}$ & $C_m>0$ only depends on $m$\\
       \hline
      odd & even & $0$ & Opposite parity\\
      \hline
      $1$ & $1$ & $\dfrac{u^\top v}{d}$ & \\
      \hline
      $2$ & $2$ & $\dfrac{(u^\top v)^2}{d(d+2)}$ &\\
      \hline
      $1$ & $3$ & $\dfrac{3u^\top v}{d(d+2)}$ &\\
      \hline
    \end{tabu}
    }
  \end{center}
    \caption{Table of formulae for $c_{p,q}(u,v) := \mathbb E_x[(x^\top u)^p(x^\top v)]$, where $u,v \in \sphere$ (fixed) and $x$ is uniformly random over the unit-sphere $\sphere$. This table is a direct application of Lemma \ref{lm:pq}. Note that $c_{p,q}(u,v)$ is symmetric in $(p,q)$ and in $(u,v)$.}
\end{table}

\begin{proof}[Proof of Lemma \ref{lm:prodexp}]
WLOG, assume $a_0=0$. Thanks to Lemma \ref{lm:pq}, one may compute
$$
(\mathbb E_x[h(x^\top u)])^2 = (\frac{a_2}{d}+\mathcal O(\frac{1}{d^2}))^2 = \frac{a_2^2}{d^2} + \mathcal O(\frac{1}{d^3}),
$$
and similarly
\begin{eqnarray*}
\begin{split}
    \mathbb E_x[h(x^\top u)h(x^\top v)] &= \sum_{p=0}^3\sum_{q=0}^3 a_p a_q\mathbb E_x[(x^\top u)^p(x^\top v)^q]+\mathcal O(\dfrac{1}{d^3})\\
    &= \dfrac{a_1^2}{d}u^\top v + \dfrac{2a_2^2}{d(d+2)}(u^\top v)^2 + \dfrac{6 a_1a_3}{d(d+2)}u^\top v+\mathcal O(\dfrac{1}{d^3})\\
    &= (\dfrac{a_1^2}{d}+\frac{6a_1a_3}{d^2})u^\top v + \dfrac{2a_2^2}{d^2}(u^\top v)^2 + +\mathcal O(\dfrac{1}{d^3}),
\end{split}
\end{eqnarray*}
and the claim follows after subtracting the previous display.
\end{proof}

\begin{proof}[Proof of Lemma \ref{lm:pq}]
Let $z \sim \mathcal N(0,I_d)$ and $x \sim \tau_d$. Then by using (hyper)spherical coordinates
\begin{eqnarray*}
\begin{split}
\mathbb E_z[(z^\top u)^p(z^\top v)^q] &= (2\pi)^{-\frac{d}{2}}\int_{0}^{\infty}\dif r\ {\rm vol}(S^{d-1}) \ r^{d-1+p+q}\ e^{-\frac{r^2}{2}}
\mathbb E[(x^\top u)^p(x^\top v)^q]\\
&=\frac{2^{\frac{p+q}{2}}\Gamma\left(\frac{p+q+d}{2}\right)}{\Gamma\left(
\frac{d}{2}\right)}\mathbb E_x[(x^\top u)^p(x^\top v)^q]
\end{split}
\end{eqnarray*}
We need $p+q$ even for a nonzero result, so we assume that. The Gaussian expectation
$\mathbb E_x[(z^\top u)^p(z^\top v)^q]$ can be computed with the Isserlis-Wick Theorem. It amounts to a sum over complete matchings of a set with $p+q$ elements subdivided into two compartments, one of size $p$ and one of size $q$. Let me organize the count according to $k$, the number of matched pairs joining the two compartments. We then get
\begin{eqnarray*}
\begin{split}
\mathbb E_z[(z^\top u)^p(z^\top v)^q] &=
\sum_{t}\binom{p}{t}\binom{q}{t}
 t!\frac{(p-t)!}{2^{\frac{p-t}{2}}\left(\frac{p-t}{2}\right)!}
\frac{(q-t)!}{2^{\frac{q-t}{2}}\left(\frac{q-t}{2}\right)!}
 (u^\top  v)^{t}(u^\top  u)^{p-t}(v^\top  v)^{q-t}
 \\
 &= \sum_{t}\binom{p}{t}\binom{q}{t}
 t!\frac{(p-t)!}{2^{\frac{p-t}{2}}\left(\frac{p-t}{2}\right)!}
\frac{(q-t)!}{2^{\frac{q-t}{2}}\left(\frac{q-t}{2}\right)!}
 (u^\top  v)^{t}
\end{split}
\end{eqnarray*}
where the sum is over $0\le t\le\min(p,q)$ of same parity as $p$ and $q$.

Finally, after some cleanup,
\begin{eqnarray*}
\begin{split}
\mathbb E_x[(x^\top u)^p(x^\top v)^q] &=
\frac{p!q!\ \Gamma\left(
\frac{d}{2}\right)}{2^{p+q} \Gamma\left(
\frac{p+q+d}{2}\right)} \sum_t
\frac{2^t}{t!\left(\frac{p-t}{2}\right)!\left(\frac{q-t}{2}\right)!}
(u^\top  v)^{t}
\end{split}
\end{eqnarray*}
with the same range of summation for $t$.
\end{proof}

\subsection{Covariance matrix of outer product of independent random vectors}
\begin{restatable}[Covariance of outer product of independent random vectors]{lm}{}
If $z$ and $x$ are independent random vectors, at least one of which has zero mean, then $\cov(\mathrm{vec}(z \otimes x)) = \mathbb E[zz^\top] \otimes \mathbb E[xx^\top]$.
\end{restatable}
\begin{proof}
Let $r$ b the dimensionality of $z$ and $s$ be the dimensionality of $x$. Every index $I \in [rs]$ can be identified with a pair $(i,j) \in [r] \times [s]$ of indices in an obvious way so that $(zx^\top)_I = z_ix_j$. For $I,I' \in [rs]$, on compute the $(I,I')$th entry of the covariance matrix of $z \otimes x \in \mathbb R^{rs}$ as 
\begin{eqnarray*}
\begin{split}
\mathbb E[(zx^\top)_I(zx^\top)_{I'}] - \mathbb E[(zx^\top)_I]\mathbb E[(zx^\top)_{I'}] &= \mathbb E[z_iz_{i'}]\mathbb E[x_jx_{j'}] - \mathbb E[z_i]\mathbb E[z_{i'}]\mathbb E[x_j]\mathbb E[x_{j'}]\\
&= \mathbb E[z_iz_{i'}]\mathbb E[x_jx_{j'}]=(\mathbb E[zz^\top] \otimes \mathbb E[xx^\top])_{I,I'},
\end{split}
\end{eqnarray*}
where in the last but one step, we have used the fact that on of $\mathbb E[z_i]$ and $\mathbb E[x_j]$ equals zero. We conclude that $\cov(z \otimes x) = \mathbb E[zz^\top] \otimes \mathbb E[xx^\top]$ as claimed.
\end{proof}

\section{Alternative proof of Theorem \ref{thm:generalkernelbound} (removing the hidden log-factors)}
Let $K:\sphere \times \sphere \to \mathbb R$ be a Mercer kernel and $\mathcal H_K$ be the induced RKHS. We are interested in lower-bounding the RKHS norm of functions in $\mathcal H_K$, which memorize the generic dataset $\mathcal D_n$. To this end, define the random variable $\eta_K(n,r) \ge 0$ by
\begin{eqnarray}
\eta_K(n,r) := \inf_{f \in B_K(r)}\frac{1}{n}\|f(X)-y\|^2.
\end{eqnarray}
 By the generalized representer theorem (see Proposition \ref{prop:representer}), every minimizer in the above problem is an element of the representer subspace $\lspan_K(X) := \{f_c:x\mapsto \sum_{i=1}^n c_iK(x,\cdot) \mid c \in \mathbb R^n\} \subseteq \mathcal H_K$. Recall that the RKHS norm of every $f_c \in \lspan_K(X)$ writes $\|f_c\|_{\mathcal H_K} =\sqrt{c^\top G c} \le \|G\|_{op}\|c\|^2$, where $G:=K(X,X) \in \mathbb R^{n \times n}$ is the kernel gram matrix. Let $\widetilde{\zeta}:=\sqrt{\|w_0\|^2/d + \zeta^2} \ge \zeta$, where $w_0$ and $\zeta$ are as in the noisy linear data generating process \eqref{eq:noisylinear}. One computes
\begin{eqnarray*}
\begin{split}
\inf_{\|c\| \le r}\frac{1}{n}\|f_c(X)-y\|^2 &=  \inf_{\|c\| \le r}\frac{1}{n}\|Gc-y\|^2 = \inf_{\|c\| \le r}\frac{1}{n}\|Gc-y\|^2\\
&\overset{(i)}{\ge}  
\left(\frac{\|y\|}{\sqrt{n}}-\frac{r\|G\|_{op}}{\sqrt{n}}\right)_+^2
\overset{(ii)}{\to} \left(\widetilde{\zeta}-\frac{r\|G\|_{op}}{\sqrt{n}}\right)_+^2,
\end{split}
\end{eqnarray*}
where \emph{(i)} is an application of Lemma \ref{lm:quadgordon} and \emph{(ii)} is thanks to the \emph{Law of Large Numbers} (and the convergence is in probability). 
Thus, if $f_c \in \lspan_K(X)$ memorizes $\mathcal D_n$ then \begin{eqnarray}
\|c\| \ge \Omega(\dfrac{\sqrt{n}}{\|G\|_{op}}).
\label{eq:cbound}
\end{eqnarray}

Thanks to \eqref{eq:rabbithole}, if we can control $\|G\|_{op}$ in \eqref{eq:cbound}, then we'd immediately get lower-bound on the Sobolev-seminorm $\mathfrak{S}(f_c)$ of any memorizer $f_c \in \lspan_K(X)$. The name of the game is then to upper-bound $\|G\|_{op}$, the operator norm of the kernel gram matrix $G$. Below, we sketch a number of examples where this can be done without difficulty.

Thanks to \eqref{eq:cbound}, the name of the game is then to upper-bound the operator norm of the kernel gram matrix $G$. We sketch a number of examples where this can be done without difficulty.
\paragraph{Infinite-width RF and NTK.}
As an example, in the case of infinite-width RF or NTK with $d \asymp n \to \infty$, we know that $\|G\|_{op} = \mathcal O(1)$ w.p $1-d^{-\Omega(1)}$. Thus, w.p $1-d^{-\Omega(1)}$, every $f=f_c \in \lspan_K(X)$ which memorizes $\mathcal D_n$ must verify $\|f\|_{\mathcal H_K}, \|c\| \ge \Omega(\sqrt{n})$. Accordingly, this would remove all log-factors from the lower-bound in Theorem \ref{thm:infiniteRFNTKLaw}.

\paragraph{Ordinary linear models.}
Here, the gram matrix is $G = XX^\top$ and for $n \asymp d \to \infty$, one has $\|G\|_{op} = \mathcal O(1)$. Accordingly, this would remove all the log-factors from the lower-bound in Theorem \ref{thm:linear}.

\paragraph{Finite-width RF with proportionate scaling \eqref{eq:proportionate}.}
It is a classical result (e.g see \cite{pennington17}) that $\|G\|_{op} = \mathcal O_{\mathbb P}(1)$ in this scenario. Accordingly, this would remove all the log-factors from the lower-bound in Theorem \ref{thm:finiteRFLaw}.


\subsection{A useful lemma}
\begin{restatable}{lm}{}
Let $A:\mathcal H_1 \to \mathcal H_2$ be a compact operator between Hilbert spaces and let $b \in \mathcal H_2$. We have the following inequalities
\begin{eqnarray}
(\|b\|-\|A\|_{op})_+ \le \inf_{v \in \mathcal H_1,\, \|u\| \le 1}\|Au-b\| \le (\|b\|-\|A^\top\|_{\min})_+ \le \|b\|,
\end{eqnarray}
where $\|A^\top\|_{\min}$ is the infinimum of the singular-values of the adjoint operator $A^\top$.
\label{lm:quadgordon}
\end{restatable}
\begin{proof}
Let $B_j$ be the unit-ball of $\mathcal H_j$. By duality of norms, one has
\begin{eqnarray*}
\begin{split}
\inf_{u \in B_1}\|Au-b\| &= \inf_{u \in B_1}\sup_{v \in B_2}\langle v,b-Au\rangle = \sup_{v \in B_2} \langle v,b\rangle - \inf_{u \in B_1}\langle u,A^\top v\rangle\\
&= \sup_{v \in B_2}\langle v,b\rangle - \|A^\top v\|
\end{split}
\end{eqnarray*}
The result follows by noting that
\begin{itemize}
\item $\|A^\top\|_{\min}\|u\| \le \|A^\top v\| \le \|A\|_{op}\|u\|$, and
    \item $\sup_{v \in B_2} \langle v,b\rangle - r\|v\| = (\|b\|-r)_+$ for any $r \in \mathbb R$. To see this, note that the optimal $v \in B_2$ must must point in the same direction as $b$. Now, set $v = (R/\|b\|)b$ and optimize over $R \in [0,1]$.
\end{itemize}
\end{proof}

\end{document}